\documentclass[pdflatex,sn-mathphys-num]{sn-jnl}% Math and Physical Sciences Numbered Reference Style

\usepackage{graphicx}%
\usepackage{multirow}%
\usepackage{amsmath,amssymb,amsfonts}%
\usepackage{amsthm}%
\usepackage{mathrsfs}%
\usepackage[most]{tcolorbox}
\usepackage[title]{appendix}%
\usepackage{xcolor}%MetaFine
\usepackage{textcomp}%
\usepackage{manyfoot}%
\usepackage{booktabs}%
\usepackage{algorithm}%
\usepackage{algorithmicx}%
\usepackage{algpseudocode}%
\usepackage{url}
\usepackage{listings}%
\usepackage{subcaption}
\usepackage{siunitx} % for \round command
\usepackage{setspace}
\usepackage[section]{placeins}
\usepackage{float}
\newcommand{\eg}{\textit{e.g.}}
\newtcolorbox{finding}[1][]{
  colback=black!5!white,    
  colframe=black!60!white,  
  fonttitle=\bfseries,     
  title=Finding,          
  arc=4pt,                 
  outer arc=4pt,
  #1                       
}

\theoremstyle{thmstyleone}%
\theoremstyle{thmstyletwo}%

\theoremstyle{thmstylethree}%

\makeatletter
\def\titraggedcenter{\leftskip=0pt plus 0.5fil\rightskip=0pt plus 0.5fil%
\parfillskip=0pt\let\hb=\break}
\def\keywordfont{\reset@font\fontsize{8bp}{9.5bp}\selectfont\leftskip=0pt\rightskip=0pt}
\makeatother
\begin{document}

%%%%hxs-version%%%

\title[Article Title]{Beyond Binary Success: A Diagnostic Meta-Evaluation Framework for Fine-Grained Manipulation}
%\title[Article Title]{MetaFine: Disentangling Understanding, Perception, and Behavior in Fine-Grained Manipulation}

\author[1]{\fnm{He-Yang} \sur{Xu}}
\author[1]{\fnm{Pengyuan} \sur{Zhang}}
\author[2]{\fnm{Zongyuan} \sur{Ge}}
\author[3]{\fnm{Xiaoshuai} \sur{Hao}}
\author[4]{\fnm{Serge} \sur{Belongie}}
\author*[1]{\fnm{Xin} \sur{Geng}}
\author*[5]{\fnm{Yuxin} \sur{Peng}}
\author*[1]{\fnm{Xiu-Shen} \sur{Wei}}\email{weixs@seu.edu.cn}

\affil[1]{\orgname{Southeast University}, \country{China}}
\affil[2]{\orgname{Monash University}, \country{Australia}}
\affil[3]{\orgname{Xiaomi EV}, \country{China}}
\affil[4]{\orgname{University of Copenhagen}, \country{Denmark}}
\affil[5]{\orgname{Peking University}, \country{China}}


\abstract{
Fine-grained manipulation marks a regime where global scene context no longer suffices, and success hinges on the tight coupling of local attribute grounding, high-fidelity spatial perception, and constraint-respecting motor execution. However, current embodied AI benchmarks collapse these capacities into binary success rates, systematically inflating reported capabilities by up to 70\% and masking the architectural bottlenecks that impede real-world deployment. We introduce \textit{MetaFine}, a diagnostic meta-evaluation framework that disentangles manipulation competency along three axes: understanding, perception, and controlled behavior. Built on a compositional task graph, MetaFine absorbs heterogeneous external benchmarks and reconstructs them into diagnostic scenarios of varying complexity under a unified protocol. Evaluating state-of-the-art vision-language-action (VLA) models through this lens exposes severe dimension-specific failures invisible to conventional metrics. Through targeted causal intervention, we identify the visual encoder's ability to preserve local spatial structure as a key bottleneck for fine-grained precision: improving it directly unlocks previously inaccessible manipulation capabilities without modifying downstream policies. MetaFine further supports hybrid real--sim validation, using limited paired real-world rollouts to calibrate scalable simulation-based estimates for more stable physical benchmarking. By shifting evaluation from ranking to diagnosis, MetaFine turns benchmarking into an actionable compass for repairing the layered capacities underlying genuine physical dexterity. The MetaFine framework, benchmarks, and supporting resources will be publicly released at our project page: \url{https://metafine.github.io/}.
}
% \keywords{Fine-Grained Manipulation, Embodied AI, Diagnostic Evaluation, \\Vision-Language-Action Models, Robotic Dexterity, Meta-Evaluation Framework}
\keywords{\raggedright
Fine-Grained Manipulation, Embodied AI, Diagnostic Evaluation, Vision-Language-\\Action Models, Robotic Dexterity, Meta-Evaluation Framework}
\maketitle

\section{Introduction}\label{sec1}

For humans, dexterity is seamless. For embodied artificial intelligence, however, the gap between coarse movement---such as nudging a box across a table---and fine-grained manipulation---such as threading a needle, aligning a precision gear, or inserting a shaped block into a matching slot---marks a fundamental increase in structural complexity. Conventional robotic manipulation can often succeed by relying on global scene context and approximate object-level transport. Fine-grained manipulation, by contrast, is a regime where global context no longer suffices: success depends on the tight coupling of local attribute grounding, high-fidelity spatial perception, and constraint-respecting control. This distinction is structural rather than merely metric: the difficulty does not lie only in smaller tolerances, but in the fact that semantic, perceptual, and motor requirements become tightly coupled. Failure in any one of these capacities can invalidate the entire task, even when the remaining components otherwise appear competent.

\begin{figure}
\centering
\includegraphics[width=0.98\textwidth]{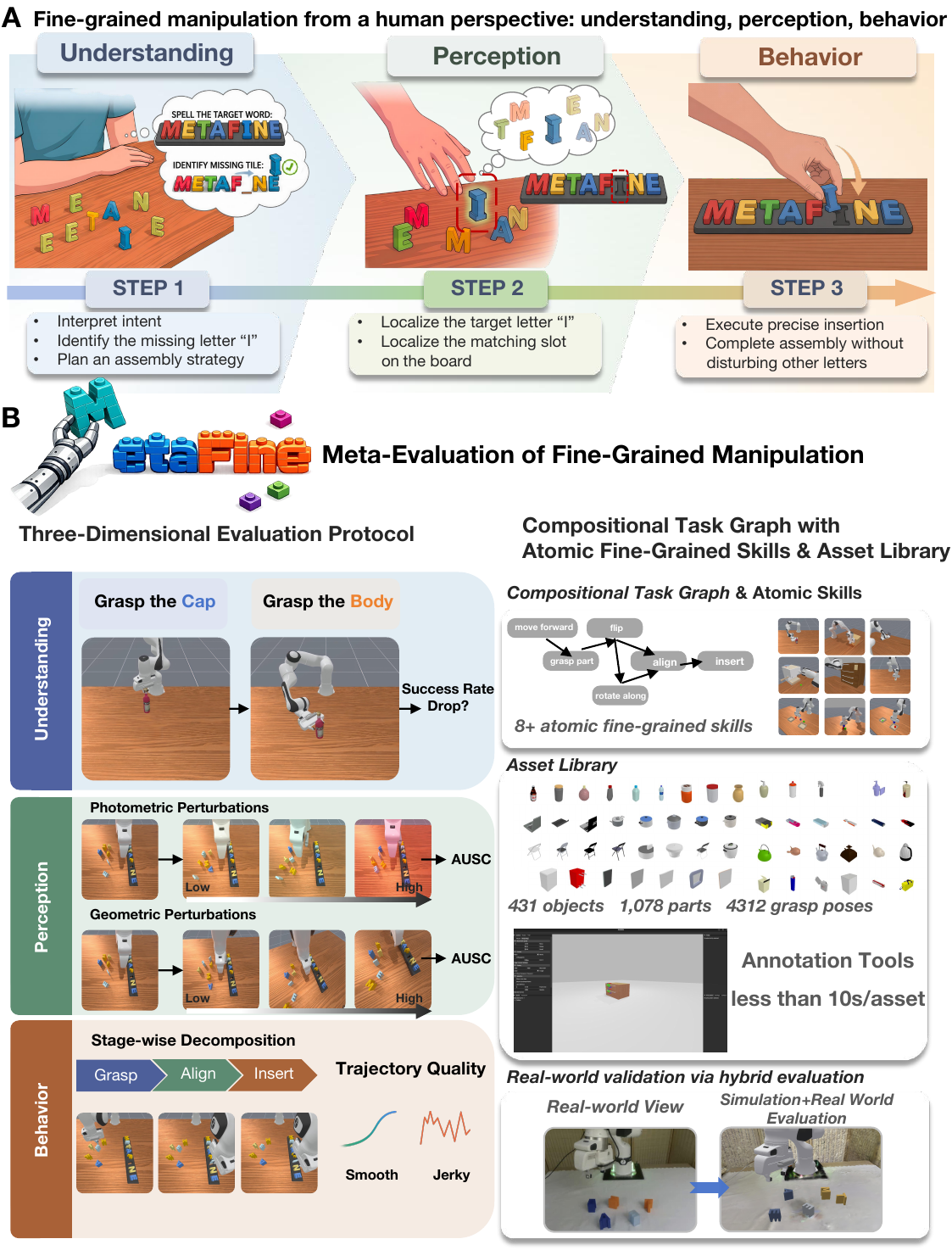}
\caption{\textbf{MetaFine: a meta-evaluation framework for fine-grained manipulation.} \textbf{A,} Letter-block assembly illustrates how understanding, perception, and behavior must jointly succeed for fine-grained manipulation. \textbf{B,} MetaFine provides a three-dimensional diagnostic protocol, a compositional task graph with an asset library, and a hybrid framework for real-world evaluation.}
\label{fig1}
\end{figure}

As illustrated in Fig.~\ref{fig1}A, assembling the word ``\texttt{METAFINE}'' by inserting the missing letter into its matching slot provides a minimal scenario in which all three competencies must succeed jointly. The agent must \emph{understand} the symbolic goal and infer that the missing target is ``\texttt{I}'', \emph{perceive} the correct block and slot among visually similar alternatives, and \emph{execute} a stable insertion without disturbing neighboring letters. The example is therefore not an extreme precision showcase, but a structural exemplar: semantic grounding~\cite{nmi2,yu2025benchmarking}, high-fidelity spatial perception~\cite{wei2021fine,fei2025libero}, and controlled execution~\cite{billard2019trends} must be jointly satisfied.

Prevailing benchmarks, however, typically flatten this structured competency profile into a single pass-or-fail outcome~\cite{li2026matters,zeng2021transporter}. Such scalar evaluation cannot reveal whether failure arises from misunderstanding the task, mislocalizing the target, or losing control during execution. Consequently, the impressive headline success rates reported by recent VLA models~\cite{kim2024openvla,black2024pi0,ze2024dp3} risk creating a premature illusion of near-human dexterity. We show that this mismatch has substantial empirical consequences: existing evaluations can overestimate fine-grained capabilities by up to 70\%, masking the bottlenecks that separate apparent task completion from genuine physical skill.

To move beyond this illusion, we introduce MetaFine, a diagnostic meta-evaluation framework that decouples manipulation competency along three axes (Fig.~\ref{fig1}B). To probe \emph{understanding}, MetaFine applies controlled semantic interventions, such as asking the robot to grasp a different part of the same object while keeping the scene fixed. To stress-test \emph{perception}, it introduces graded geometric~\cite{geometry,hendrycks2019benchmarking} and photometric~\cite{ma2025comprehensive} perturbations. To characterize \emph{behavior}, it decomposes long-horizon tasks into atomic skill phases and measures stage-wise success and trajectory smoothness. The protocol is anchored by a compositional task graph whose nodes encode atomic skills and whose paths define tasks of varying complexity, enabling systematic diagnosis and structured absorption of external task suites. This design is important because diagnostic evaluation must be tied to task construction: only tasks that explicitly separate local attributes, spatial relations, and constrained motion can reveal which capacity is responsible for failure.

Evaluating state-of-the-art VLAs under this three-dimensional lens uncovers failures hidden by conventional metrics. In \emph{understanding}, apparent instruction following collapses under attribute-level intervention, and compound commands require unified language--action co-optimization to avoid behavioral arrest after the first sub-task. In \emph{perception}, the visual encoder's preservation of local spatial structure sets a practical precision ceiling: information lost at the visual frontend cannot be recovered downstream. In \emph{behavior}, deterministic regression yields stable but rigid trajectories, whereas stochastic generative policies provide richer corrective diversity but can accumulate spatial drift under ambiguous visual inputs~\cite{smoothness}. By exposing these bottlenecks separately, MetaFine converts aggregate success rates into actionable design insights.

Beyond diagnosis, MetaFine is designed as extensible evaluation infrastructure. We pursue fine-grainedness in two coupled senses: at the task level, tasks resolve manipulation into separable local demands; at the evaluation level, the protocol resolves competency into separable diagnoses. The latter is meaningful only in the presence of the former. MetaFine operationalizes this link through a task-graph formalism that absorbs tasks from RoboTwin~\cite{robotwin}, CALVIN~\cite{calvin}, and LIBERO~\cite{libero} into a shared diagnostic space, scaling to 431 objects and 4,312 grasp poses. It also supports hybrid real--sim evaluation by combining 3D Gaussian Splatting-based real-to-sim transfer with limited real-world calibration. Thus, MetaFine is not simply another task suite with new objects or scenes; it is intended as an evaluation substrate that turns heterogeneous manipulation tasks into comparable diagnostic evidence. MetaFine therefore shifts evaluation from ranking to diagnosis, providing infrastructure for measuring, comparing, and repairing the layered capacities underlying physical dexterity.

\section{Results}\label{Results}

We first expose how conventional evaluation inflates apparent competency, and then use MetaFine’s tripartite protocol to diagnose failures in understanding, perception, and behavior.

This separation also determines how we interpret performance. A high nominal success rate is not treated as evidence of general dexterity unless it remains stable under semantic substitution, perceptual perturbation, and behavioral decomposition. Conversely, a low end-to-end score is not treated as an undifferentiated failure: MetaFine asks whether the policy fails because it cannot identify the intended part, cannot preserve the spatial relation needed for contact, or cannot generate the corrective motions needed to satisfy the constraint. This diagnostic stance motivates the structure of the following results.

\subsection{Evaluation Protocol}

MetaFine evaluates fine-grained manipulation beyond binary success. In understanding, controlled semantic interventions modify attribute-level instructions while holding the scene fixed. In perception, geometric and photometric perturbations are applied at three severity levels (L1--L3), and robustness is summarized by success rates and area under the success curve (AUSC). In behavior, long-horizon tasks are decomposed into task-graph stages, with both stage-wise success and trajectory smoothness reported. These measurements are intentionally complementary: semantic interventions test whether language changes can redirect behavior, perturbation curves test whether visual representations remain reliable away from the nominal setting, and stage-wise behavior identifies where execution collapses in a long-horizon task. Detailed task specifications and acceptance criteria are provided in Supplementary Note~\ref{supp:sec1}.

\begin{figure}[!t]
\centering
\includegraphics[width=1.0\textwidth]{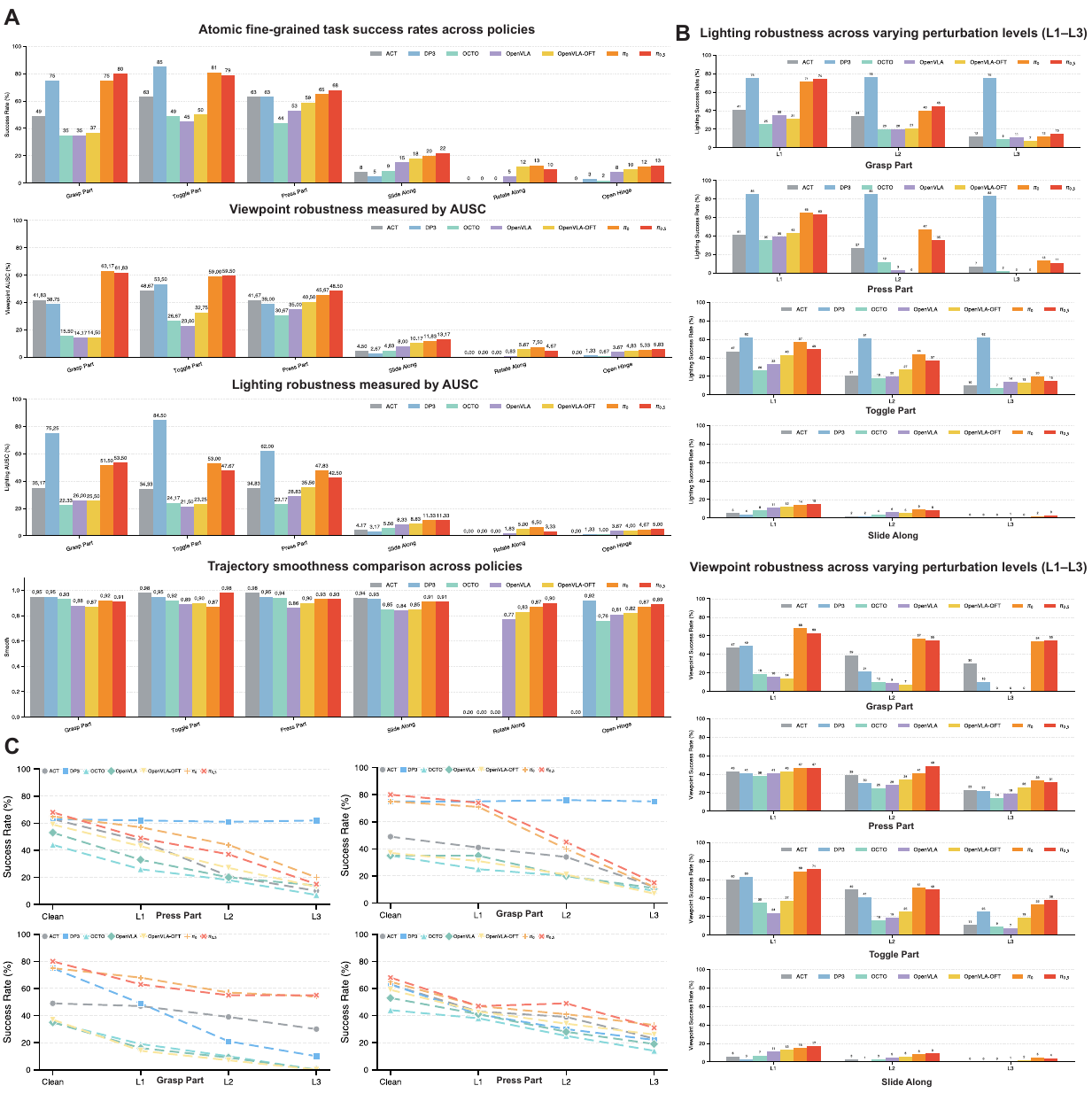}
\caption{\textbf{MetaFine disentangles manipulation competency.} \textbf{A,} Atomic task analysis reveals policy differences in success, robustness, and trajectory smoothness. \textbf{B,} Performance degrades under perturbations. \textbf{C,} Mean trends show that clean performance overestimates generalization.}
\vspace{-1.2em}
\label{fig2}
\end{figure}

All experiments are executed within the same evaluation infrastructure. We instantiate tasks ranging from atomic skills to long-horizon compositions and fine-tune every policy with 100 demonstrations per task under identical conditions. This standardization is critical because the goal is not to tune each model to its preferred data regime, but to expose how architectural choices behave under matched supervision, scenes, and acceptance criteria. We benchmark seven representative policies spanning four architectural families: ACT~\cite{zhao2023act}, DP3~\cite{ze2024dp3}, Octo~\cite{team2024octo}, OpenVLA~\cite{kim2024openvla}, OpenVLA-OFT~\cite{kim2025openVLAoft}, $\pi_0$~\cite{black2024pi0}, and $\pi_{0.5}$~\cite{black2025pi05}. Together, they cover major design axes in visual encoders, action generation, and language--action coupling, allowing observed failures to be interpreted as architecture-linked bottlenecks rather than idiosyncrasies of a single policy.

The task suite is deliberately heterogeneous. Atomic tasks such as Grasp Part, Press Part, Toggle Part, Rotate Along, and Slide Along isolate individual fine-grained skills, whereas multi-skill tasks such as peg-in-hole, plug-in-charger, and stack-pyramid require the ordered composition of multiple skills. This breadth allows MetaFine to ask two complementary questions: whether a policy can solve isolated local constraints, and whether the same policy can maintain those constraints as they compound over a longer horizon.

\subsection{The Illusion of Competency in Conventional Evaluation}

Standard manipulation benchmarks often report object-level grasping success above 95\%, but rarely enforce whether the correct part is used or whether the intended physical constraint is satisfied. When MetaFine enforces part-level constraints under fixed environmental conditions, this apparent competency fractures (Fig.~\ref{fig2}A). The best policy reaches only 80\% on Grasp Part. As constraints tighten, performance drops further: Toggle Part peaks at 85\%, Press Part at 68\%, and Rotate Along collapses to 12\%. Thus, conventional metrics often credit coarse approximations that do not constitute genuine physical mastery. This gap is especially problematic for fine-grained manipulation because a trial may satisfy a coarse object displacement criterion while violating the very part-level contact, direction, or ordering constraint defining the task.

Binary metrics also conceal distinct failure mechanisms, with apparent capability dropping by over 70\% between nominal and perturbed conditions. Under severe lighting perturbation (Fig.~\ref{fig2}B), the top policy on Grasp Part drops from 80\% to 15\%, whereas under viewpoint perturbation it retains 55\%. On Toggle Part, two models with similar nominal success (85\% and 79\%) diverge sharply under L3 lighting, with one retaining 83\% and the other dropping to 11\%. Trajectory smoothness reveals a further dissociation: one policy achieves a smoothness score of 0.90 on Rotate Along but only 10\% success, indicating fluid motion without correct spatial grounding. These examples show that binary success enforces a false equivalence between policies by collapsing perceptual robustness, spatial grounding, and kinematic stability into a single scalar. In practice, this means that two policies with the same headline number may require entirely different architectural repairs: one may need a more robust visual frontend, another better semantic grounding, and a third a more adaptive action head.

\begin{finding}[title=Finding 1]
Conventional binary metrics create an illusion of competency: enforcing fine-grained constraints and perturbations exposes up to 70\% capability inflation and reveals failures invisible to aggregate success rates.
\end{finding}

\begin{figure}[!t]
\centering
\includegraphics[width=0.92\textwidth]{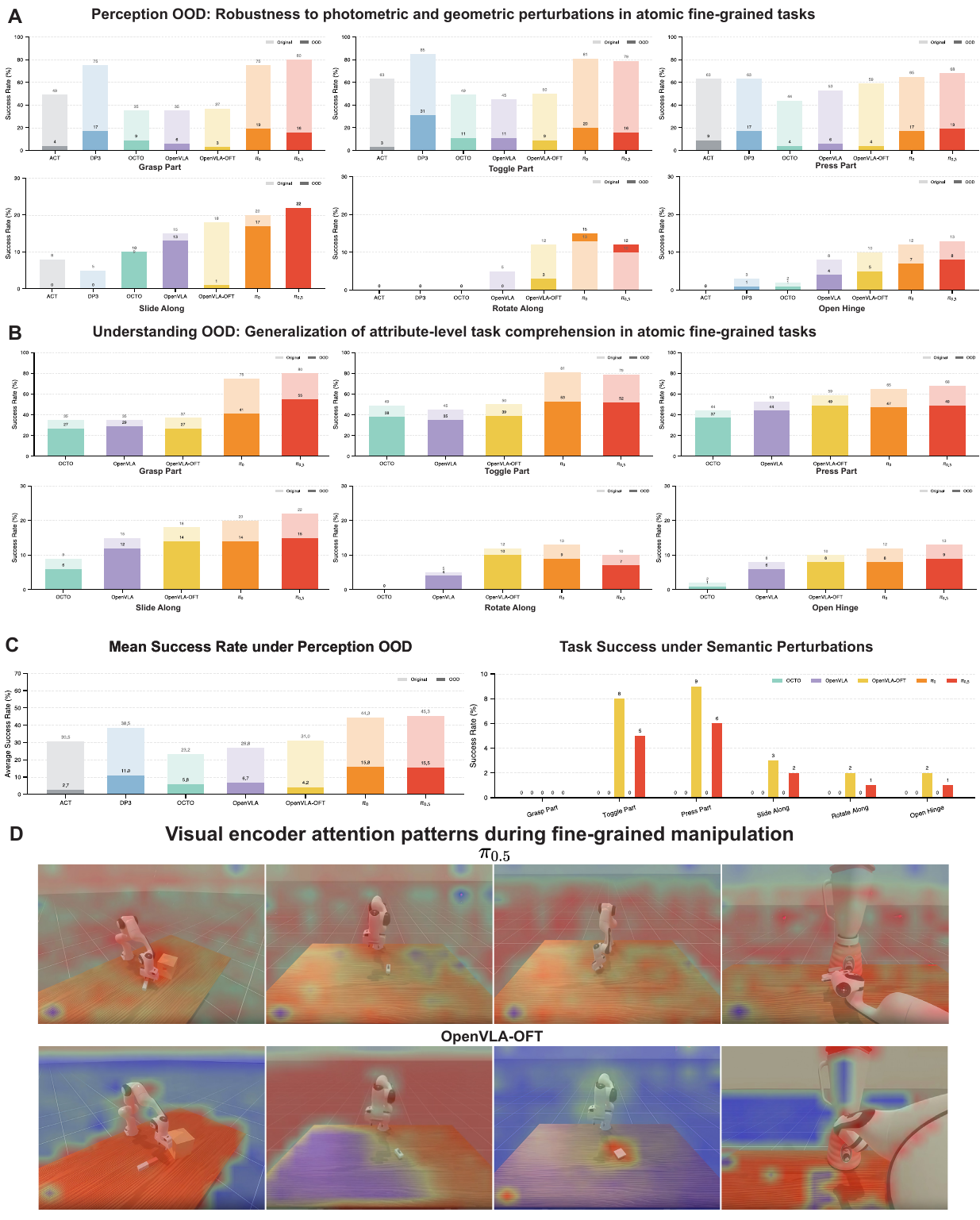}
\caption{\textbf{Out-of-distribution (OOD) evaluation across understanding and perception.} \textbf{A,} Robustness to photometric and geometric perturbations. \textbf{B,} Generalization under semantic interventions. \textbf{C,} Aggregate OOD statistics. \textbf{D,} Encoder attention patterns during fine-grained manipulation.}
\vspace{-1.2em}
\label{fig3_OOD}
\end{figure}

\subsection{Understanding: The Deficit in Semantic Grounding}

We test whether policies genuinely ground attribute-level language by modifying the instruction while keeping the visual scene unchanged, for example changing ``grasp the cap of the bottle'' to ``grasp the body of the bottle.'' A grounded policy should redirect its behavior to the newly specified region. None of the five evaluated VLAs succeeds on the modified instruction, universally scoring 0\% across substitutions (Fig.~\ref{fig3_OOD}C). The substitution still affects original-task behavior unevenly: $\pi_{0.5}$ and $\pi_0$ drop by 31.2\% and 34\%, whereas OpenVLA-OFT, OpenVLA, and Octo drop by 10\%, 6\%, and 8\%. This pattern indicates disrupted instruction--scene association rather than genuine semantic re-grounding. The policies are sensitive to the changed language, but this sensitivity does not manifest as a correct spatial redirection. Language therefore acts more like a correlated cue attached to a familiar scene than a compositional constraint that can be recombined at test time.

We next evaluate compound commands requiring sequential transitions between sub-tasks. In the task ``put the green cube in the left box, and put the other cubes in the right box,'' $\pi_{0.5}$ and $\pi_0$ achieve non-zero success across all four stages, including the later stages that require switching to the remaining cubes. By contrast, OpenVLA-OFT succeeds moderately on the first two stages but collapses to 0\% on stages three and four; OpenVLA and Octo show the same behavioral arrest. This split suggests that unified language--action co-optimization preserves responsiveness to multiple sub-instructions, whereas decoupled pathways often fail beyond the initial sub-task. The result also clarifies why single-step instruction following is insufficient evidence of semantic control: a model may respond to the first clause of a command yet fail to maintain or update the remaining linguistic structure once execution begins. The failure is not merely lower success or noisier motion: in the later stages, the decoupled policies frequently cease to initiate meaningful actions toward the second goal, indicating that the sequential structure of the instruction has not been preserved in the policy's action-generating state.

\begin{finding}[title=Finding 2]
Current VLAs rely heavily on instruction--scene correlations: attribute-level interventions collapse to 0\% success, and compound commands require unified language--action co-optimization to avoid behavioral arrest.
\end{finding}

Taken together, the semantic intervention and compound-command studies show that language failures occur at two levels. At the local level, policies do not reliably bind attribute words to the correct region in a fixed scene. At the sequential level, several architectures fail to preserve unresolved sub-instructions after the first action segment. Both failures are hidden when evaluation only checks whether a familiar demonstration-like behavior happens to complete the original task.

\subsection{Perception: Spatial Fidelity as the Precision Bottleneck}

Fine-grained manipulation requires visual representations that preserve local spatial structure. The peg-in-hole task provides a natural diagnostic because grasp, alignment, and insertion impose progressively tighter spatial demands. Although all five VLAs achieve near-zero overall success (0--3\%), stage-wise evaluation reveals different failure depths. OpenVLA-OFT grasps the peg in 47\% of trials and aligns it in 19\%, whereas $\pi_{0.5}$ grasps in 39\% but never aligns (0\%). CAM visualizations show a corresponding encoder-level difference: OpenVLA-OFT's DINOv2+SigLIP dual encoder concentrates activation on the peg and hole, while $\pi_{0.5}$'s SigLIP-only encoder spreads attention diffusely across the scene (Fig.~\ref{fig3_OOD}D). The precision gap therefore originates at the visual frontend, before downstream action generation is invoked. Importantly, this does not imply that the action head is irrelevant; rather, it indicates that action generation can only operate on the spatial information preserved by the visual encoder. Once the target relation is blurred or displaced in the representation, later modules have no reliable signal from which to recover the missing local geometry. This stage-wise view is important: a binary end-to-end score would label both models as failed, whereas MetaFine distinguishes an agent that fails during alignment from one that cannot move beyond grasping.

\begin{figure}[!t]
\centering
\includegraphics[width=0.95\textwidth]{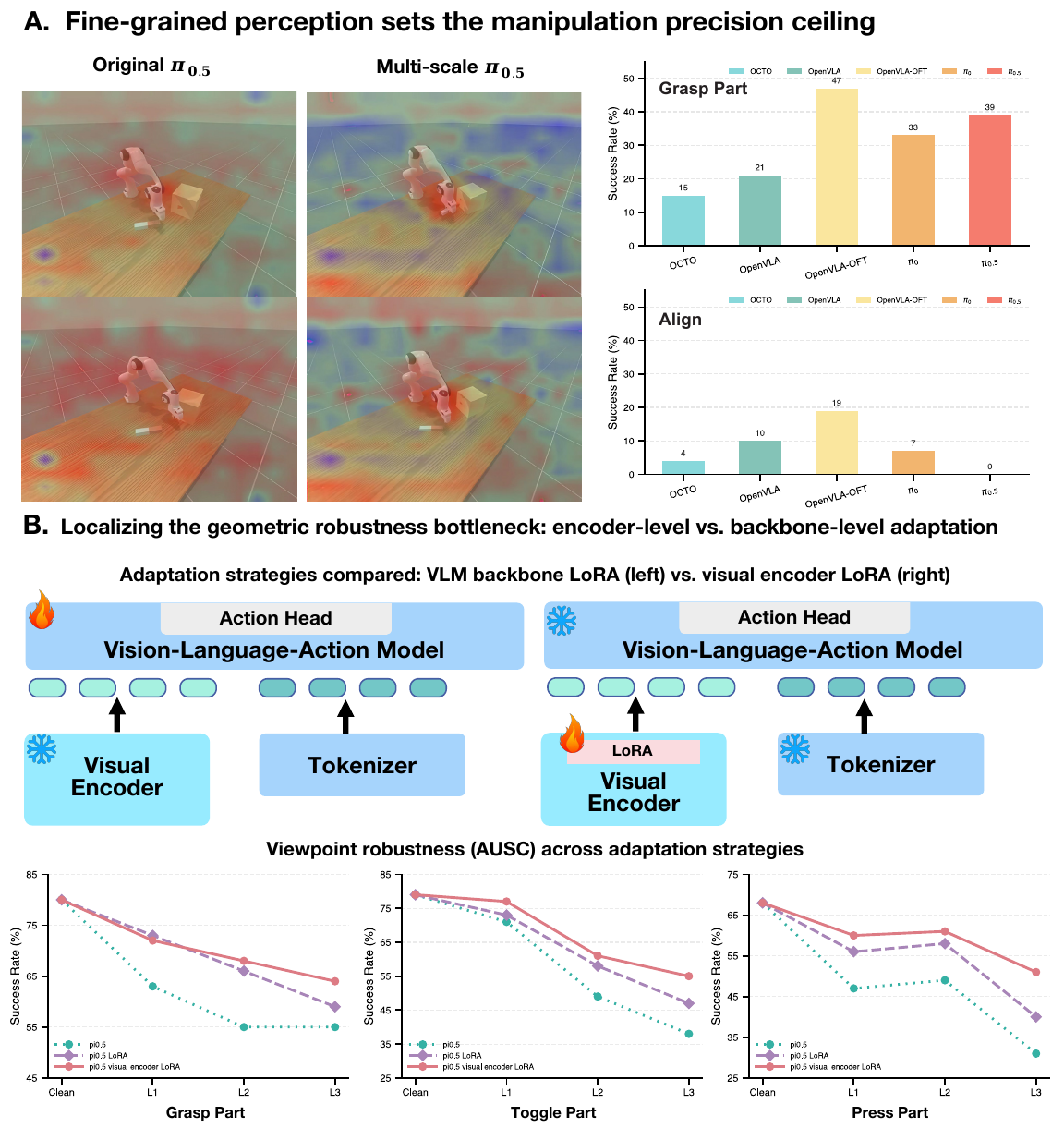}
\caption{\textbf{Perception as a precision and robustness bottleneck.} \textbf{A,} A multi-scale encoder intervention concentrates visual attention and improves stage-wise performance on peg-in-hole. \textbf{B,} Under viewpoint perturbation, encoder-level adaptation matches the gains of full-backbone adaptation.}
\vspace{-1.2em}
\label{fig5}
\end{figure}

We further test whether geometric robustness is localized to the encoder. Freezing downstream components of $\pi_{0.5}$, we compare one-shot adaptation of the visual encoder output with LoRA fine-tuning of the full VLM backbone. Both preserve the nominal 80\% Grasp Part success rate, but under viewpoint perturbations encoder-only adaptation reaches 71.00\% AUSC, matching or exceeding full-backbone LoRA (69.25\%) despite modifying far fewer parameters (Fig.~\ref{fig5}B). This shows that viewpoint degradation primarily corrupts spatial representations before language--action processing. It also suggests that adapting a large VLM backbone is not always the most direct remedy: when the failure is localized to spatial representation, lightweight encoder-level correction can recover robustness more efficiently.

Photometric robustness follows a different pattern. Under viewpoint shifts, the $\pi_0$ family achieves the strongest AUSC across tasks; under lighting perturbations, DP3 dominates, reaching 75.25\%, 84.50\%, and 62.00\% AUSC. This dissociation indicates that geometric and photometric robustness probe different mechanisms: the former depends strongly on spatial representations, whereas the latter benefits from modality-level invariance and pretraining diversity. Treating perceptual robustness as a single aggregate property would therefore obscure which part of the visual pipeline should be improved.

Finally, we perform a causal intervention by replacing $\pi_{0.5}$'s single-scale SigLIP encoder with a multi-scale cross-attention encoder while freezing the VLM backbone and action head. Grasp success rises from 39\% to 67\%, and alignment is unlocked from 0\% to 32\% (Fig.~\ref{fig5}A). Because all downstream components remain unchanged, the improvement is attributable to enhanced spatial representation rather than altered control, data scale, or policy optimization. This intervention turns a correlational observation across architectures into causal evidence that the visual frontend can act as a manipulable precision bottleneck. It also rules out a common alternative explanation: the improvement cannot be attributed to larger policy capacity or additional action training, because the downstream VLM and action head are held fixed.

\begin{finding}[title=Finding 3]
The visual encoder's preservation of local spatial structure is a key causal bottleneck for fine-grained precision; improving it alone unlocks capabilities that downstream policy components could not recover.
\end{finding}

These observations also constrain how future perception modules should be evaluated. A visual encoder that performs well on object recognition may still fail to preserve the local correspondences required for insertion, alignment, or constrained contact. MetaFine therefore evaluates perception at the level of task-relevant spatial relations rather than image-level recognition alone, linking representational fidelity directly to manipulation consequences.

\subsection{Behavior: The Stability--Expressiveness Trade-off}

\begin{figure}
\centering
\includegraphics[width=1.0\textwidth]{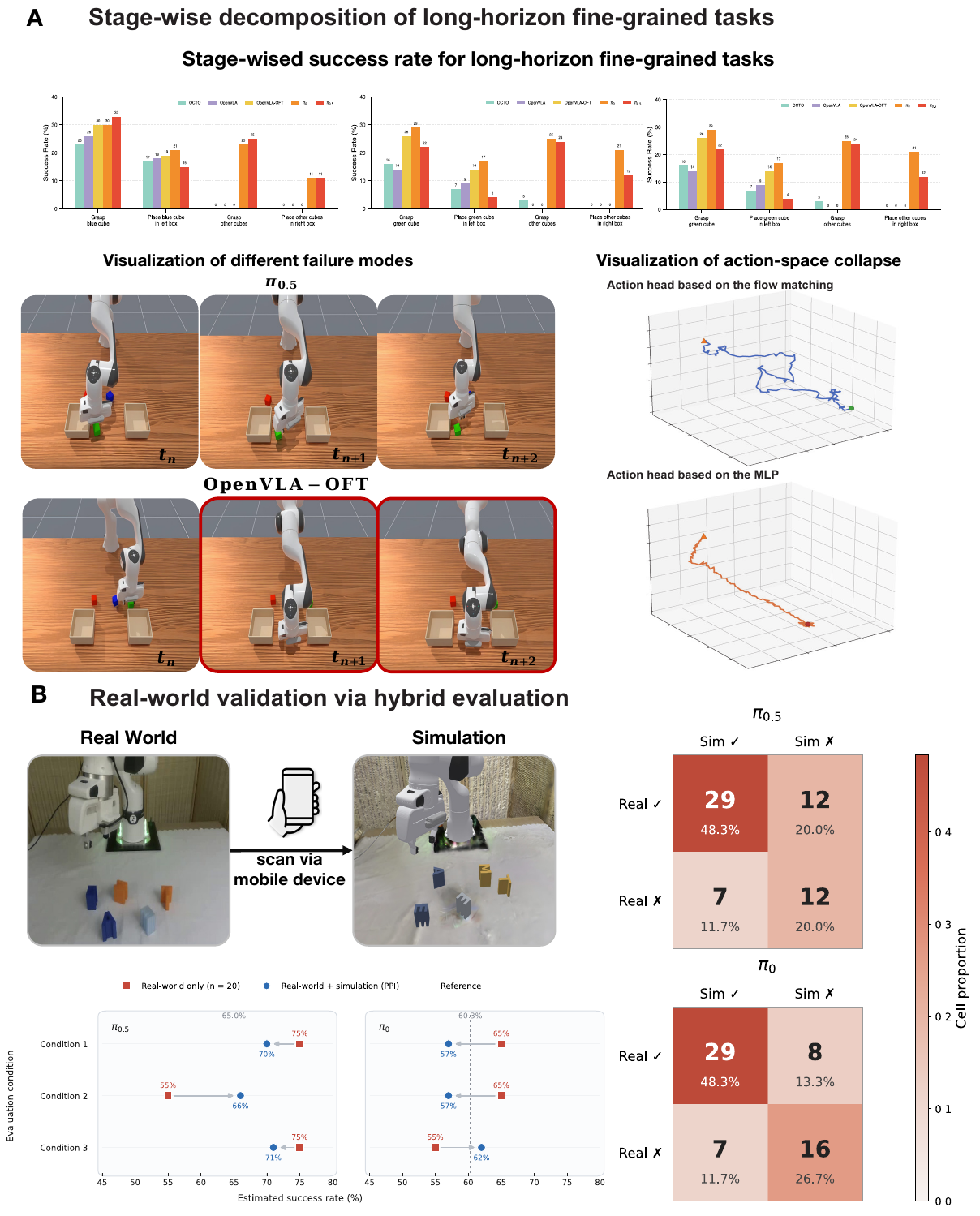}
\caption{\textbf{Stage-wise behavior and hybrid real-world validation.} \textbf{A,} Stage-wise analysis reveals spatial drift and repetitive behavior. \textbf{B,} Paired real--sim rollouts enable PPI-calibrated physical evaluation.}
\label{fig3}
\end{figure}

Continuous action policies differ in how they propagate perceptual uncertainty. OpenVLA-OFT uses deterministic $\ell_1$ regression through an MLP head, whereas $\pi_{0.5}$ uses a flow-matching action expert that samples from a learned conditional distribution. On peg-in-hole, OpenVLA-OFT produces coherent approach trajectories despite systematic perceptual bias, while $\pi_{0.5}$ and $\pi_0$ exhibit spatial drift with directionally inconsistent steps (Fig.~\ref{fig3}A). Determinism can therefore stabilize motion, but this stability becomes rigid in the constrained alignment and insertion phases: OpenVLA-OFT repeatedly executes similar ineffective approaches and lacks the corrective diversity needed for precision insertion. The failure mode is coherent rather than chaotic, which makes it easy to mistake stable motion for competent control unless the task is decomposed into stages.

Stochastic generation offers the opposite trade-off. Flow matching and diffusion can represent multiple feasible corrections, but when conditioned on ambiguous spatial features, successive samples can point in inconsistent directions. Over long horizons, these inconsistencies accumulate into drift. Octo's diffusion head reaches only 4\% alignment success, confirming that stochasticity alone cannot compensate for poorly grounded perception. The key design problem is therefore not whether an action head is deterministic or stochastic in isolation, but how its uncertainty is coupled to the reliability of the perceptual state on which it conditions. Fine-grained behavior likely requires stage-adaptive mechanisms: stable convergence when the spatial target is reliable, and controlled exploration when small corrective motions are needed.

\begin{finding}[title=Finding 4]
Action generation faces a stability--expressiveness dilemma under perceptual uncertainty: deterministic regression yields stable but rigid failures, whereas stochastic generation offers diversity but can compound spatial drift.
\end{finding}

\subsection{Real-World Validation via Hybrid Evaluation}
\label{sec:res-validation}

Beyond simulation-based diagnosis, MetaFine supports hybrid real--sim evaluation by combining 3D Gaussian Splatting-based real-to-sim transfer with prediction-powered inference (PPI)~\cite{ppi}. Real rollouts are authoritative but costly and statistically unstable at small sample sizes; simulation is scalable but imperfect. MetaFine pairs a small set of real rollouts with large-scale simulation under a platform-defined evaluation distribution, producing calibrated estimates anchored to physical ground truth. The shared distribution is important because it prevents calibration from becoming a post-hoc adjustment on mismatched conditions; real and simulated trials are paired to probe the same task, object, and environmental configuration whenever possible.

On a fine-grained grasping task, as shown in Fig.~\ref{fig3}B, we evaluate $\pi_0$ and $\pi_{0.5}$ using large-scale simulation, large-sample real-world references, and three paired real--sim calibration sets. Across paired sets, simulation remains sufficiently faithful to support calibration. Hardware-only estimates from $n=20$ trials vary widely, from 55\% to 75\% for $\pi_{0.5}$ and from 55\% to 65\% for $\pi_0$. PPI calibration substantially reduces this instability, lowering the standard deviation from 11.5\% to 2.6\% for $\pi_{0.5}$ and from 5.8\% to 2.9\% for $\pi_0$, while keeping all calibrated estimates close to the large-sample real reference. In particular, a 55\% hardware-only estimate for $\pi_{0.5}$ is corrected to 66\%, closely recovering the 65.0\% large-sample reference that 20 real rollouts alone fail to resolve. Thus, MetaFine turns limited real-world budgets into more stable and comparable estimates of physical performance. Importantly, the protocol does not replace real-world evaluation with simulation; instead, it uses real rollouts to anchor simulation statistically, making physical benchmarking more reproducible under realistic hardware constraints.

\section{Discussion}

MetaFine challenges the reliance on binary success rates by showing that fine-grained manipulation has a structured failure geometry. In coarse manipulation, wide margins may allow weaknesses in language grounding, perception, or control to compensate for one another. In fine-grained tasks, each dimension can become independently fatal. This is why task-level fine-grainedness and evaluation-level fine-grainedness must be coupled: only tasks that stress local attributes, spatial relations, and constrained execution can produce failure modes that diagnostic evaluation can meaningfully and systematically disentangle.

\begin{figure}[!htbp]
	\centering
	\includegraphics[width=0.94\textwidth]{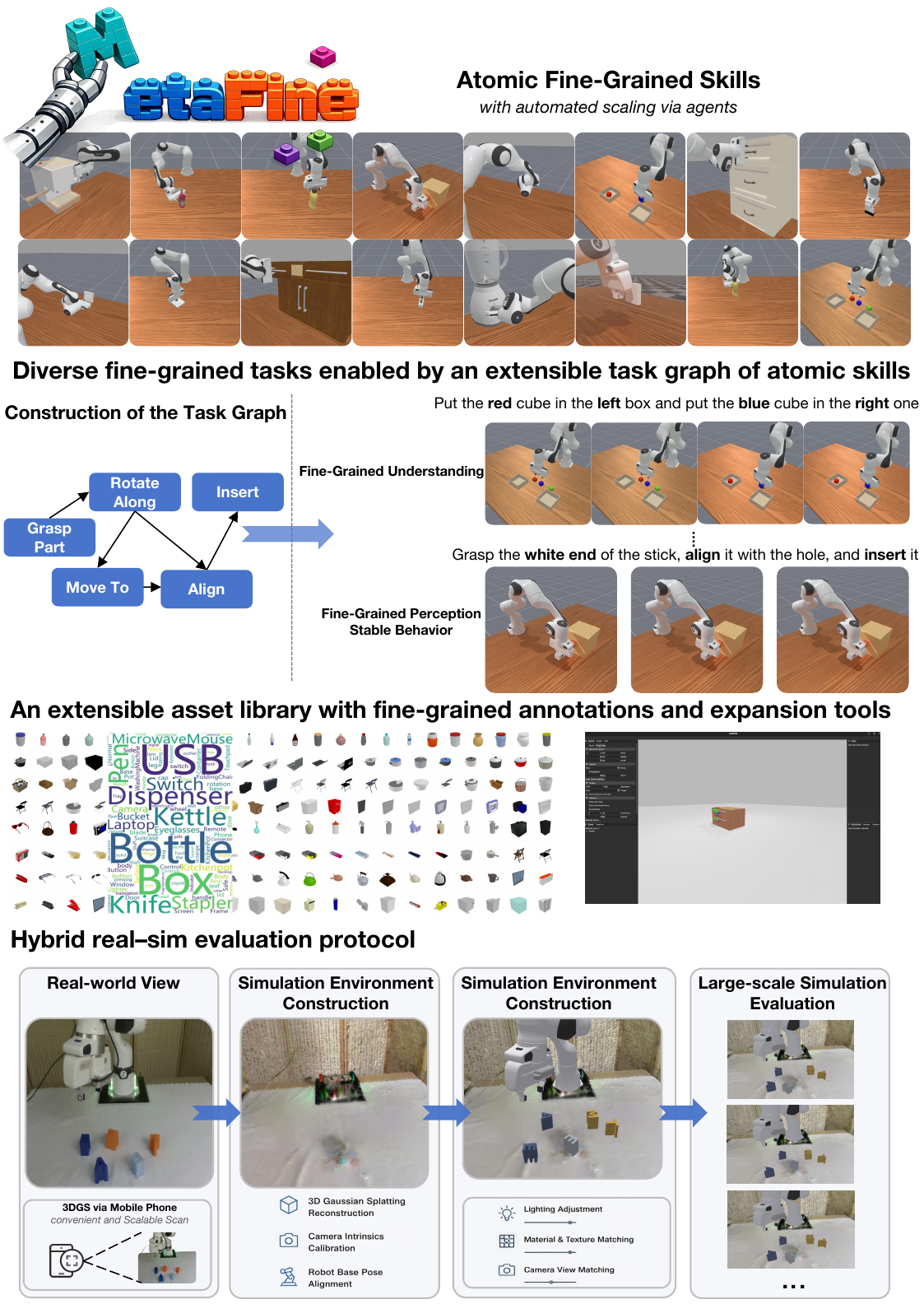} 
	\caption{\textbf{MetaFine overview.} Top: scalable atomic fine-grained skills. Middle: compositional task graphs probing understanding, perception, and control. Bottom: hybrid real--sim evaluation pipeline, where real scenes are transferred to simulation via mobile-phone 3DGS reconstruction and used for large-scale simulation with paired real rollouts for calibrated evaluation.}
	\label{fig6}
    % \vspace{-0.3cm}
\end{figure}

Our findings suggest a concrete agenda for VLA model design. Language should function as a compositional spatial constraint rather than a task-level cue. Visual frontends should be optimized not only for semantic recognition, but also for preserving local spatial structure under viewpoint and appearance variation. Action generation should be stage-adaptive, combining stability during constrained execution with enough expressiveness for local correction under uncertainty. These requirements cannot be solved by independently scaling language models, visual encoders, or action heads; they require co-design across the sensorimotor pipeline. MetaFine provides a practical way to guide such co-design because it identifies whether a failure originates in language grounding, spatial representation, or action generation, rather than merely reporting that the final task failed.

MetaFine also provides community infrastructure rather than a closed benchmark. Its task graph and asset library allow external benchmarks to be reconstructed within a shared diagnostic protocol, enabling results from heterogeneous task suites to be compared in a common capability space. This matters because discrepancies across benchmarks often remain difficult to interpret: they may reflect different object distributions, task definitions, sensory conditions, or acceptance criteria. By mapping tasks into a shared representation, MetaFine makes such differences explicit and connects them to the competencies being tested. This is the main role of MetaFine as a meta-evaluation framework: it does not erase differences between benchmarks, but expresses them in a common language so that differences become analyzable rather than anecdotal. Its hybrid real--sim pathway further offers a practical route toward reproducible real-world evaluation, where simulation supplies statistical power and paired real rollouts provide physical grounding. By releasing MetaFine as an open ecosystem, we aim to shift embodied AI evaluation from superficial ranking toward systematic diagnosis of the layered capacities required for genuine dexterity.

At the same time, MetaFine should be viewed as a diagnostic platform rather than a final claim of solved evaluation. Its value depends on continued expansion of atomic skills, assets, perturbations, and real-world registrations. This extensibility is intentional: fine-grained manipulation is an open-ended domain, and evaluation infrastructure must evolve with the capabilities and failure modes of new policies.

\section{Methodology}\label{methodology}

MetaFine is a meta-evaluation framework for fine-grained manipulation, not a fixed benchmark. It unifies compositional task construction, cross-benchmark evaluation, and hybrid real-to-sim performance estimation (Fig.~\ref{fig6}).

\textbf{Compositional task graph.} MetaFine represents manipulation tasks as compositions of atomic skills, such as grasp-part, align, insert, and constrained-move. Binding these skills to objects, parts, and parameters generates tasks of varying complexity under a unified representation. The same task graph also defines evaluation stages, so that task construction and diagnostic decomposition remain aligned.

\textbf{Asset library and benchmark absorption.} The task graph operates over an extensible asset library comprising 431 part-annotated objects, 1{,}078 labeled parts, and 4{,}312 candidate grasp poses. The same representation also absorbs tasks from existing benchmarks, including RoboTwin, CALVIN, and LIBERO, by mapping them into a common skill-based form with unified task structure and evaluation criteria, thereby enabling consistent comparison across heterogeneous suites.

\textbf{Three-dimensional diagnostic protocol.} MetaFine decomposes competency into understanding, perception, and controlled behavior. Understanding is tested through semantic interventions, perception through geometric and photometric perturbations, and behavior through stage-wise success and trajectory stability.

\textbf{Hybrid real-to-sim evaluation.} MetaFine fixes a shared evaluation distribution, executes most trials in simulation, and calibrates the estimate with paired real rollouts through PPI, yielding statistically grounded estimates comparable across laboratories.

\section*{Declarations}

\subsection*{Data availability}
The fine-grained annotation assets and task data are available on Hugging Face (\url{https://huggingface.co/datasets/hiangx/MetaFine}) and on ModelScope (\url{https://www.modelscope.cn/datasets/hiangx/MetaFine}).

\subsection*{Code availability}
The code for agent training and simulation experiments is available on GitHub (\url{https://github.com/Hiangx-robotics/MetaFine}).

\subsection*{Competing interests}
The authors declare no competing interests.

\section*{Contributions and Acknowledgments}
\label{sec:contributions}
\textbf{Authors} He-Yang Xu, Pengyuan Zhang, Zongyuan Ge, Xiaoshuai Hao, Serge Belongie, \textbf{Xin Geng} (\textbf{Corresponding author}), \textbf{Yuxin Peng} (\textbf{Corresponding author}), \textbf{Xiu-Shen Wei} (\textbf{Corresponding author}).

\noindent \textbf{Acknowledgments} We would like to sincerely thank for the tremendous support from the team, including those not listed above: Qiyuan Zhuang, Meipo Dai, Yingjie Shuai, Hongxiang Gao, Yu-Hao Wei, Hong-Tao Yu, Xin-Yang Zhao, Zi-Yao Lin.

\section*{Author Contributions Statement}

X.-S. Wei conceived and initiated the project, proposed the basic idea, distilled the core contributions of MetaFine, contributed to manuscript writing and presentation of the key findings, and provided equipment and computational resources. H.-Y. Xu designed the MetaFine framework and three-dimensional evaluation pipeline, and contributed to atomic-skill development, model transfer and training, and hybrid evaluation, including camera calibration, real-scene reconstruction, and prediction-powered inference experiments. P. Zhang contributed to atomic-skill development, model training and evaluation, real-to-sim scene transfer, and the design, validation, and analysis of fine-grained tasks. Z. Ge helped define the core contributions and advised on manuscript writing and presentation. X. Hao helped organize the findings and contributed to manuscript writing. S. Belongie participated in the early framework design and discussions leading to the main conclusions. X. Geng discussed the findings and reviewed, commented on, and edited the final manuscript. Y. Peng co-initiated the project, discussed the findings, and reviewed, commented on, and edited the final manuscript. All authors discussed the results and contributed to the final manuscript.

%All core contributors read and approved the final version.

\newpage
\bibliography{sn-bibliography1}% common bib file


\begin{thebibliography}{42}
\ifx \bisbn   \undefined \def \bisbn  #1{ISBN #1}\fi
\ifx \binits  \undefined \def \binits#1{#1}\fi
\ifx \bauthor  \undefined \def \bauthor#1{#1}\fi
\ifx \batitle  \undefined \def \batitle#1{#1}\fi
\ifx \bjtitle  \undefined \def \bjtitle#1{#1}\fi
\ifx \bvolume  \undefined \def \bvolume#1{\textbf{#1}}\fi
\ifx \byear  \undefined \def \byear#1{#1}\fi
\ifx \bissue  \undefined \def \bissue#1{#1}\fi
\ifx \bfpage  \undefined \def \bfpage#1{#1}\fi
\ifx \blpage  \undefined \def \blpage #1{#1}\fi
\ifx \burl  \undefined \def \burl#1{\textsf{#1}}\fi
\ifx \doiurl  \undefined \def \doiurl#1{\url{https://doi.org/#1}}\fi
\ifx \betal  \undefined \def \betal{\textit{et al.}}\fi
\ifx \binstitute  \undefined \def \binstitute#1{#1}\fi
\ifx \binstitutionaled  \undefined \def \binstitutionaled#1{#1}\fi
\ifx \bctitle  \undefined \def \bctitle#1{#1}\fi
\ifx \beditor  \undefined \def \beditor#1{#1}\fi
\ifx \bpublisher  \undefined \def \bpublisher#1{#1}\fi
\ifx \bbtitle  \undefined \def \bbtitle#1{#1}\fi
\ifx \bedition  \undefined \def \bedition#1{#1}\fi
\ifx \bseriesno  \undefined \def \bseriesno#1{#1}\fi
\ifx \blocation  \undefined \def \blocation#1{#1}\fi
\ifx \bsertitle  \undefined \def \bsertitle#1{#1}\fi
\ifx \bsnm \undefined \def \bsnm#1{#1}\fi
\ifx \bsuffix \undefined \def \bsuffix#1{#1}\fi
\ifx \bparticle \undefined \def \bparticle#1{#1}\fi
\ifx \barticle \undefined \def \barticle#1{#1}\fi
\bibcommenthead
\ifx \bconfdate \undefined \def \bconfdate #1{#1}\fi
\ifx \botherref \undefined \def \botherref #1{#1}\fi
\ifx \url \undefined \def \url#1{\textsf{#1}}\fi
\ifx \bchapter \undefined \def \bchapter#1{#1}\fi
\ifx \bbook \undefined \def \bbook#1{#1}\fi
\ifx \bcomment \undefined \def \bcomment#1{#1}\fi
\ifx \oauthor \undefined \def \oauthor#1{#1}\fi
\ifx \citeauthoryear \undefined \def \citeauthoryear#1{#1}\fi
\ifx \endbibitem  \undefined \def \endbibitem {}\fi
\ifx \bconflocation  \undefined \def \bconflocation#1{#1}\fi
\ifx \arxivurl  \undefined \def \arxivurl#1{\textsf{#1}}\fi
\csname PreBibitemsHook\endcsname

%%% 1
\bibitem[\protect\citeauthoryear{Mon-Williams et~al.}{2025}]{nmi2}
\begin{barticle}
\bauthor{\bsnm{Mon-Williams}, \binits{R.}},
\bauthor{\bsnm{Li}, \binits{G.}},
\bauthor{\bsnm{Long}, \binits{R.}},
\bauthor{\bsnm{Du}, \binits{W.}},
\bauthor{\bsnm{Lucas}, \binits{C.G.}}:
\batitle{Embodied large language models enable robots to complete complex tasks in unpredictable environments}.
\bjtitle{Nature Machine Intelligence}
\bvolume{7}(\bissue{4}),
\bfpage{592}--\blpage{601}
(\byear{2025})
\end{barticle}
\endbibitem

%%% 2
\bibitem[\protect\citeauthoryear{Yu et~al.}{2026}]{yu2025benchmarking}
\begin{bchapter}
\bauthor{\bsnm{Yu}, \binits{H.-T.}},
\bauthor{\bsnm{Peng}, \binits{Y.}},
\bauthor{\bsnm{Belongie}, \binits{S.}},
\bauthor{\bsnm{Wei}, \binits{X.-S.}}:
\bctitle{Benchmarking large vision-language models on fine-grained image tasks: A comprehensive evaluation}.
In: \bbtitle{International Conference on Learning Representations (ICLR)}
(\byear{2026})
\end{bchapter}
\endbibitem

%%% 3
\bibitem[\protect\citeauthoryear{Wei et~al.}{2022}]{wei2021fine}
\begin{barticle}
\bauthor{\bsnm{Wei}, \binits{X.-S.}},
\bauthor{\bsnm{Song}, \binits{Y.-Z.}},
\bauthor{\bsnm{Mac~Aodha}, \binits{O.}},
\bauthor{\bsnm{Wu}, \binits{J.}},
\bauthor{\bsnm{Peng}, \binits{Y.}},
\bauthor{\bsnm{Tang}, \binits{J.}},
\bauthor{\bsnm{Yang}, \binits{J.}},
\bauthor{\bsnm{Belongie}, \binits{S.}}:
\batitle{Fine-grained image analysis with deep learning: A survey}.
\bjtitle{IEEE Transactions on Pattern Analysis and Machine Intelligence}
\bvolume{44}(\bissue{12}),
\bfpage{8927}--\blpage{8948}
(\byear{2022})
\end{barticle}
\endbibitem

%%% 4
\bibitem[\protect\citeauthoryear{Fei et~al.}{2025}]{fei2025libero}
\begin{botherref}
\oauthor{\bsnm{Fei}, \binits{S.}},
\oauthor{\bsnm{Wang}, \binits{S.}},
\oauthor{\bsnm{Shi}, \binits{J.}},
\oauthor{\bsnm{Dai}, \binits{Z.}},
\oauthor{\bsnm{Cai}, \binits{J.}},
\oauthor{\bsnm{Qian}, \binits{P.}},
\oauthor{\bsnm{Ji}, \binits{L.}},
\oauthor{\bsnm{He}, \binits{X.}},
\oauthor{\bsnm{Zhang}, \binits{S.}},
\oauthor{\bsnm{Fei}, \binits{Z.}},
\oauthor{\bsnm{Fu}, \binits{J.}},
\oauthor{\bsnm{Gong}, \binits{J.}},
\oauthor{\bsnm{Qiu}, \binits{X.}}:
{LIBERO-Plus}: In-depth robustness analysis of vision-language-action models.
arXiv preprint arXiv:2510.13626
(2025)
\end{botherref}
\endbibitem

%%% 5
\bibitem[\protect\citeauthoryear{Billard and Kragic}{2019}]{billard2019trends}
\begin{barticle}
\bauthor{\bsnm{Billard}, \binits{A.}},
\bauthor{\bsnm{Kragic}, \binits{D.}}:
\batitle{Trends and challenges in robot manipulation}.
\bjtitle{Science}
\bvolume{364}(\bissue{6446}),
\bfpage{8414}
(\byear{2019})
\end{barticle}
\endbibitem

%%% 6
\bibitem[\protect\citeauthoryear{Li et~al.}{2026}]{li2026matters}
\begin{barticle}
\bauthor{\bsnm{Li}, \binits{X.}},
\bauthor{\bsnm{Li}, \binits{P.}},
\bauthor{\bsnm{Qian}, \binits{L.}},
\bauthor{\bsnm{Liu}, \binits{M.}},
\bauthor{\bsnm{Wang}, \binits{D.}},
\bauthor{\bsnm{Liu}, \binits{J.}},
\bauthor{\bsnm{Kang}, \binits{B.}},
\bauthor{\bsnm{Ma}, \binits{X.}},
\bauthor{\bsnm{Wang}, \binits{X.}},
\bauthor{\bsnm{Guo}, \binits{D.}},
\bauthor{\bsnm{Kong}, \binits{T.}},
\bauthor{\bsnm{Zhang}, \binits{H.}},
\bauthor{\bsnm{Liu}, \binits{H.}}:
\batitle{What matters in building vision--language--action models for generalist robots}.
\bjtitle{Nature Machine Intelligence}
\bvolume{8},
\bfpage{158}--\blpage{172}
(\byear{2026})
\end{barticle}
\endbibitem

%%% 7
\bibitem[\protect\citeauthoryear{Zeng et~al.}{2021}]{zeng2021transporter}
\begin{bchapter}
\bauthor{\bsnm{Zeng}, \binits{A.}},
\bauthor{\bsnm{Florence}, \binits{P.}},
\bauthor{\bsnm{Tompson}, \binits{J.}},
\bauthor{\bsnm{Welker}, \binits{S.}},
\bauthor{\bsnm{Chien}, \binits{J.}},
\bauthor{\bsnm{Attarian}, \binits{M.}},
\bauthor{\bsnm{Armstrong}, \binits{T.}},
\bauthor{\bsnm{Krasin}, \binits{I.}},
\bauthor{\bsnm{Duong}, \binits{D.}},
\bauthor{\bsnm{Sindhwani}, \binits{V.}},
\bauthor{\bsnm{Lee}, \binits{J.}}:
\bctitle{{Transporter Networks}: Rearranging the visual world for robotic manipulation}.
In: \bbtitle{Proceedings of the 2020 Conference on Robot Learning (CoRL)}.
\bsertitle{Proceedings of Machine Learning Research},
vol. \bseriesno{155},
pp. \bfpage{726}--\blpage{747}
(\byear{2021})
\end{bchapter}
\endbibitem

%%% 8
\bibitem[\protect\citeauthoryear{Kim et~al.}{2025}]{kim2024openvla}
\begin{bchapter}
\bauthor{\bsnm{Kim}, \binits{M.J.}},
\bauthor{\bsnm{Pertsch}, \binits{K.}},
\bauthor{\bsnm{Karamcheti}, \binits{S.}},
\bauthor{\bsnm{Xiao}, \binits{T.}},
\bauthor{\bsnm{Balakrishna}, \binits{A.}},
\bauthor{\bsnm{Nair}, \binits{S.}},
\bauthor{\bsnm{Rafailov}, \binits{R.}},
\bauthor{\bsnm{Foster}, \binits{E.P.}},
\bauthor{\bsnm{Sanketi}, \binits{P.R.}},
\bauthor{\bsnm{Vuong}, \binits{Q.}},
\bauthor{\bsnm{Kollar}, \binits{T.}},
\bauthor{\bsnm{Burchfiel}, \binits{B.}},
\bauthor{\bsnm{Tedrake}, \binits{R.}},
\bauthor{\bsnm{Sadigh}, \binits{D.}},
\bauthor{\bsnm{Levine}, \binits{S.}},
\bauthor{\bsnm{Liang}, \binits{P.}},
\bauthor{\bsnm{Finn}, \binits{C.}}:
\bctitle{{OpenVLA}: An open-source vision-language-action model}.
In: \bbtitle{Proceedings of The 8th Conference on Robot Learning (CoRL)}.
\bsertitle{Proceedings of Machine Learning Research},
vol. \bseriesno{270},
pp. \bfpage{2679}--\blpage{2713}
(\byear{2025})
\end{bchapter}
\endbibitem

%%% 9
\bibitem[\protect\citeauthoryear{Black et~al.}{2025}]{black2024pi0}
\begin{bchapter}
\bauthor{\bsnm{Black}, \binits{K.}},
\bauthor{\bsnm{Brown}, \binits{N.}},
\bauthor{\bsnm{Driess}, \binits{D.}},
\bauthor{\bsnm{Esmail}, \binits{A.}},
\bauthor{\bsnm{Equi}, \binits{M.R.}},
\bauthor{\bsnm{Finn}, \binits{C.}},
\bauthor{\bsnm{Fusai}, \binits{N.}},
\bauthor{\bsnm{Groom}, \binits{L.}},
\bauthor{\bsnm{Hausman}, \binits{K.}},
\bauthor{\bsnm{Ichter}, \binits{B.}},
\bauthor{\bsnm{Jakubczak}, \binits{S.}},
\bauthor{\bsnm{Jones}, \binits{T.}},
\bauthor{\bsnm{Ke}, \binits{L.}},
\bauthor{\bsnm{Levine}, \binits{S.}},
\bauthor{\bsnm{Li-Bell}, \binits{A.}},
\bauthor{\bsnm{Mothukuri}, \binits{M.}},
\bauthor{\bsnm{Nair}, \binits{S.}},
\bauthor{\bsnm{Pertsch}, \binits{K.}},
\bauthor{\bsnm{Shi}, \binits{L.X.}},
\bauthor{\bsnm{Smith}, \binits{L.}},
\bauthor{\bsnm{Tanner}, \binits{J.}},
\bauthor{\bsnm{Vuong}, \binits{Q.}},
\bauthor{\bsnm{Walling}, \binits{A.}},
\bauthor{\bsnm{Wang}, \binits{H.}},
\bauthor{\bsnm{Zhilinsky}, \binits{U.}}:
\bctitle{$\pi_0$: A vision-language-action flow model for general robot control}.
In: \bbtitle{Proceedings of Robotics: Science and Systems (RSS)}
(\byear{2025})
\end{bchapter}
\endbibitem

%%% 10
\bibitem[\protect\citeauthoryear{Ze et~al.}{2024}]{ze2024dp3}
\begin{bchapter}
\bauthor{\bsnm{Ze}, \binits{Y.}},
\bauthor{\bsnm{Zhang}, \binits{G.}},
\bauthor{\bsnm{Zhang}, \binits{K.}},
\bauthor{\bsnm{Hu}, \binits{C.}},
\bauthor{\bsnm{Wang}, \binits{M.}},
\bauthor{\bsnm{Xu}, \binits{H.}}:
\bctitle{{3D} diffusion policy: Generalizable visuomotor policy learning via simple {3D} representations}.
In: \bbtitle{Proceedings of Robotics: Science and Systems (RSS)}
(\byear{2024})
\end{bchapter}
\endbibitem

%%% 11
\bibitem[\protect\citeauthoryear{Zhu et~al.}{2025}]{geometry}
\begin{bchapter}
\bauthor{\bsnm{Zhu}, \binits{F.}},
\bauthor{\bsnm{Wu}, \binits{H.}},
\bauthor{\bsnm{Guo}, \binits{S.}},
\bauthor{\bsnm{Liu}, \binits{Y.}},
\bauthor{\bsnm{Cheang}, \binits{C.}},
\bauthor{\bsnm{Kong}, \binits{T.}}:
\bctitle{Irasim: A fine-grained world model for robot manipulation}.
In: \bbtitle{Proceedings of the IEEE/CVF International Conference on Computer Vision (ICCV)},
pp. \bfpage{9834}--\blpage{9844}
(\byear{2025})
\end{bchapter}
\endbibitem

%%% 12
\bibitem[\protect\citeauthoryear{Hendrycks and Dietterich}{2019}]{hendrycks2019benchmarking}
\begin{bchapter}
\bauthor{\bsnm{Hendrycks}, \binits{D.}},
\bauthor{\bsnm{Dietterich}, \binits{T.}}:
\bctitle{Benchmarking neural network robustness to common corruptions and perturbations}.
In: \bbtitle{International Conference on Learning Representations (ICLR)}
(\byear{2019})
\end{bchapter}
\endbibitem

%%% 13
\bibitem[\protect\citeauthoryear{Ma et~al.}{2025}]{ma2025comprehensive}
\begin{barticle}
\bauthor{\bsnm{Ma}, \binits{G.}},
\bauthor{\bsnm{Wang}, \binits{Z.}},
\bauthor{\bsnm{Yuan}, \binits{Z.}},
\bauthor{\bsnm{Wang}, \binits{X.}},
\bauthor{\bsnm{Yuan}, \binits{B.}},
\bauthor{\bsnm{Tao}, \binits{D.}}:
\batitle{A comprehensive survey of data augmentation in visual reinforcement learning}.
\bjtitle{International Journal of Computer Vision}
\bvolume{133}(\bissue{10}),
\bfpage{7368}--\blpage{7405}
(\byear{2025})
\end{barticle}
\endbibitem

%%% 14
\bibitem[\protect\citeauthoryear{Flash and Hogan}{1985}]{smoothness}
\begin{barticle}
\bauthor{\bsnm{Flash}, \binits{T.}},
\bauthor{\bsnm{Hogan}, \binits{N.}}:
\batitle{The coordination of arm movements: an experimentally confirmed mathematical model}.
\bjtitle{Journal of neuroscience}
\bvolume{5}(\bissue{7}),
\bfpage{1688}--\blpage{1703}
(\byear{1985})
\end{barticle}
\endbibitem

%%% 15
\bibitem[\protect\citeauthoryear{Mu et~al.}{2025}]{robotwin}
\begin{bchapter}
\bauthor{\bsnm{Mu}, \binits{Y.}},
\bauthor{\bsnm{Chen}, \binits{T.}},
\bauthor{\bsnm{Chen}, \binits{Z.}},
\bauthor{\bsnm{Peng}, \binits{S.}},
\bauthor{\bsnm{Lan}, \binits{Z.}},
\bauthor{\bsnm{Gao}, \binits{Z.}},
\bauthor{\bsnm{Liang}, \binits{Z.}},
\bauthor{\bsnm{Yu}, \binits{Q.}},
\bauthor{\bsnm{Zou}, \binits{Y.}},
\bauthor{\bsnm{Xu}, \binits{M.}},
\bauthor{\bsnm{Lin}, \binits{L.}},
\bauthor{\bsnm{Xie}, \binits{Z.}},
\bauthor{\bsnm{Ding}, \binits{M.}},
\bauthor{\bsnm{Luo}, \binits{P.}}:
\bctitle{{RoboTwin}: Dual-arm robot benchmark with generative digital twins}.
In: \bbtitle{Proceedings of the IEEE/CVF Conference on Computer Vision and Pattern Recognition (CVPR)},
pp. \bfpage{27649}--\blpage{27660}
(\byear{2025})
\end{bchapter}
\endbibitem

%%% 16
\bibitem[\protect\citeauthoryear{Mees et~al.}{2022}]{calvin}
\begin{barticle}
\bauthor{\bsnm{Mees}, \binits{O.}},
\bauthor{\bsnm{Hermann}, \binits{L.}},
\bauthor{\bsnm{Rosete-Beas}, \binits{E.}},
\bauthor{\bsnm{Burgard}, \binits{W.}}:
\batitle{Calvin: A benchmark for language-conditioned policy learning for long-horizon robot manipulation tasks}.
\bjtitle{IEEE Robotics and Automation Letters}
\bvolume{7}(\bissue{3}),
\bfpage{7327}--\blpage{7334}
(\byear{2022})
\end{barticle}
\endbibitem

%%% 17
\bibitem[\protect\citeauthoryear{Liu et~al.}{2023}]{libero}
\begin{barticle}
\bauthor{\bsnm{Liu}, \binits{B.}},
\bauthor{\bsnm{Zhu}, \binits{Y.}},
\bauthor{\bsnm{Gao}, \binits{C.}},
\bauthor{\bsnm{Feng}, \binits{Y.}},
\bauthor{\bsnm{Liu}, \binits{Q.}},
\bauthor{\bsnm{Zhu}, \binits{Y.}},
\bauthor{\bsnm{Stone}, \binits{P.}}:
\batitle{{LIBERO}: Benchmarking knowledge transfer for lifelong robot learning}.
\bjtitle{Advances in Neural Information Processing Systems}
\bvolume{36},
\bfpage{44776}--\blpage{44791}
(\byear{2023})
\end{barticle}
\endbibitem

%%% 18
\bibitem[\protect\citeauthoryear{Zhao et~al.}{2023}]{zhao2023act}
\begin{bchapter}
\bauthor{\bsnm{Zhao}, \binits{T.Z.}},
\bauthor{\bsnm{Kumar}, \binits{V.}},
\bauthor{\bsnm{Levine}, \binits{S.}},
\bauthor{\bsnm{Finn}, \binits{C.}}:
\bctitle{Learning fine-grained bimanual manipulation with low-cost hardware}.
In: \bbtitle{Robotics: Science and Systems (RSS)}
(\byear{2023})
\end{bchapter}
\endbibitem

%%% 19
\bibitem[\protect\citeauthoryear{Team et~al.}{2024}]{team2024octo}
\begin{botherref}
\oauthor{\bsnm{Team}, \binits{O.M.}},
\oauthor{\bsnm{Ghosh}, \binits{D.}},
\oauthor{\bsnm{Walke}, \binits{H.}},
\oauthor{\bsnm{Pertsch}, \binits{K.}},
\oauthor{\bsnm{Black}, \binits{K.}},
\oauthor{\bsnm{Mees}, \binits{O.}},
\oauthor{\bsnm{Dasari}, \binits{S.}},
\oauthor{\bsnm{Hejna}, \binits{J.}},
\oauthor{\bsnm{Kreiman}, \binits{T.}},
\oauthor{\bsnm{Xu}, \binits{C.}}, et al.:
Octo: An open-source generalist robot policy.
arXiv preprint arXiv:2405.12213
(2024)
\end{botherref}
\endbibitem

%%% 20
\bibitem[\protect\citeauthoryear{Kim et~al.}{2025}]{kim2025openVLAoft}
\begin{bchapter}
\bauthor{\bsnm{Kim}, \binits{M.J.}},
\bauthor{\bsnm{Finn}, \binits{C.}},
\bauthor{\bsnm{Liang}, \binits{P.}}:
\bctitle{Fine-tuning vision-language-action models: Optimizing speed and success}.
In: \bbtitle{Proceedings of Robotics: Science and Systems (RSS)}
(\byear{2025})
\end{bchapter}
\endbibitem

%%% 21
\bibitem[\protect\citeauthoryear{Black et~al.}{2025}]{black2025pi05}
\begin{bchapter}
\bauthor{\bsnm{Black}, \binits{K.}},
\bauthor{\bsnm{Brown}, \binits{N.}},
\bauthor{\bsnm{Darpinian}, \binits{J.}},
\bauthor{\bsnm{Dhabalia}, \binits{K.}},
\bauthor{\bsnm{Driess}, \binits{D.}},
\bauthor{\bsnm{Esmail}, \binits{A.}},
\bauthor{\bsnm{Equi}, \binits{M.R.}},
\bauthor{\bsnm{Finn}, \binits{C.}},
\bauthor{\bsnm{Fusai}, \binits{N.}},
\bauthor{\bsnm{Galliker}, \binits{M.Y.}},
\bauthor{\bsnm{Ghosh}, \binits{D.}},
\bauthor{\bsnm{Groom}, \binits{L.}},
\bauthor{\bsnm{Hausman}, \binits{K.}},
\bauthor{\bsnm{Ichter}, \binits{B.}},
\bauthor{\bsnm{Jakubczak}, \binits{S.}},
\bauthor{\bsnm{Jones}, \binits{T.}},
\bauthor{\bsnm{Ke}, \binits{L.}},
\bauthor{\bsnm{LeBlanc}, \binits{D.}},
\bauthor{\bsnm{Levine}, \binits{S.}},
\bauthor{\bsnm{Li-Bell}, \binits{A.}},
\bauthor{\bsnm{Mothukuri}, \binits{M.}},
\bauthor{\bsnm{Nair}, \binits{S.}},
\bauthor{\bsnm{Pertsch}, \binits{K.}},
\bauthor{\bsnm{Ren}, \binits{A.Z.}},
\bauthor{\bsnm{Shi}, \binits{L.X.}},
\bauthor{\bsnm{Smith}, \binits{L.}},
\bauthor{\bsnm{Springenberg}, \binits{J.T.}},
\bauthor{\bsnm{Stachowicz}, \binits{K.}},
\bauthor{\bsnm{Tanner}, \binits{J.}},
\bauthor{\bsnm{Vuong}, \binits{Q.}},
\bauthor{\bsnm{Walke}, \binits{H.}},
\bauthor{\bsnm{Walling}, \binits{A.}},
\bauthor{\bsnm{Wang}, \binits{H.}},
\bauthor{\bsnm{Yu}, \binits{L.}},
\bauthor{\bsnm{Zhilinsky}, \binits{U.}}:
\bctitle{$\pi_{0.5}$: a vision-language-action model with open-world generalization}.
In: \bbtitle{Proceedings of The 9th Conference on Robot Learning (CoRL)}.
\bsertitle{Proceedings of Machine Learning Research},
vol. \bseriesno{305},
pp. \bfpage{17}--\blpage{40}
(\byear{2025})
\end{bchapter}
\endbibitem

%%% 22
\bibitem[\protect\citeauthoryear{Angelopoulos et~al.}{2023}]{ppi}
\begin{barticle}
\bauthor{\bsnm{Angelopoulos}, \binits{A.N.}},
\bauthor{\bsnm{Bates}, \binits{S.}},
\bauthor{\bsnm{Cand{\`e}s}, \binits{E.J.}},
\bauthor{\bsnm{Jordan}, \binits{M.I.}},
\bauthor{\bsnm{Zrnic}, \binits{T.}}:
\batitle{Prediction-powered inference}.
\bjtitle{Science}
\bvolume{382}(\bissue{6671}),
\bfpage{669}--\blpage{674}
(\byear{2023})
\end{barticle}
\endbibitem

%%% 23
\bibitem[\protect\citeauthoryear{Gu et~al.}{2023}]{mu2023maniskill2}
\begin{bchapter}
\bauthor{\bsnm{Gu}, \binits{J.}},
\bauthor{\bsnm{Xiang}, \binits{F.}},
\bauthor{\bsnm{Li}, \binits{X.}},
\bauthor{\bsnm{Ling}, \binits{Z.}},
\bauthor{\bsnm{Liu}, \binits{X.}},
\bauthor{\bsnm{Mu}, \binits{T.}},
\bauthor{\bsnm{Tang}, \binits{Y.}},
\bauthor{\bsnm{Tao}, \binits{S.}},
\bauthor{\bsnm{Wei}, \binits{X.}},
\bauthor{\bsnm{Yao}, \binits{Y.}},
\bauthor{\bsnm{Yuan}, \binits{X.}},
\bauthor{\bsnm{Xie}, \binits{P.}},
\bauthor{\bsnm{Huang}, \binits{Z.}},
\bauthor{\bsnm{Chen}, \binits{R.}},
\bauthor{\bsnm{Su}, \binits{H.}}:
\bctitle{{ManiSkill2}: A unified benchmark for generalizable manipulation skills}.
In: \bbtitle{International Conference on Learning Representations (ICLR)}
(\byear{2023})
\end{bchapter}
\endbibitem

%%% 24
\bibitem[\protect\citeauthoryear{Song et~al.}{2025}]{song2023langpartgpd}
\begin{barticle}
\bauthor{\bsnm{Song}, \binits{Y.}},
\bauthor{\bsnm{Sun}, \binits{P.}},
\bauthor{\bsnm{Jin}, \binits{P.}},
\bauthor{\bsnm{Ren}, \binits{Y.}},
\bauthor{\bsnm{Zheng}, \binits{Y.}},
\bauthor{\bsnm{Li}, \binits{Z.}},
\bauthor{\bsnm{Chu}, \binits{X.}},
\bauthor{\bsnm{Zhang}, \binits{Y.}},
\bauthor{\bsnm{Li}, \binits{T.}},
\bauthor{\bsnm{Gu}, \binits{J.}}:
\batitle{Learning {6-DoF} fine-grained grasp detection based on part affordance grounding}.
\bjtitle{IEEE Transactions on Automation Science and Engineering}
\bvolume{22},
\bfpage{15200}--\blpage{15214}
(\byear{2025})
\end{barticle}
\endbibitem

%%% 25
\bibitem[\protect\citeauthoryear{Huang et~al.}{2024}]{huang2024copa}
\begin{bchapter}
\bauthor{\bsnm{Huang}, \binits{H.}},
\bauthor{\bsnm{Lin}, \binits{F.}},
\bauthor{\bsnm{Hu}, \binits{Y.}},
\bauthor{\bsnm{Wang}, \binits{S.}},
\bauthor{\bsnm{Gao}, \binits{Y.}}:
\bctitle{{CoPa}: General robotic manipulation through spatial constraints of parts with foundation models}.
In: \bbtitle{2024 IEEE/RSJ International Conference on Intelligent Robots and Systems (IROS)},
pp. \bfpage{9488}--\blpage{9495}
(\byear{2024})
\end{bchapter}
\endbibitem

%%% 26
\bibitem[\protect\citeauthoryear{Singh et~al.}{2024}]{tian2024cgdf}
\begin{bchapter}
\bauthor{\bsnm{Singh}, \binits{G.}},
\bauthor{\bsnm{Kalwar}, \binits{S.}},
\bauthor{\bsnm{Karim}, \binits{M.F.}},
\bauthor{\bsnm{Sen}, \binits{B.}},
\bauthor{\bsnm{Govindan}, \binits{N.}},
\bauthor{\bsnm{Sridhar}, \binits{S.}},
\bauthor{\bsnm{Krishna}, \binits{K.M.}}:
\bctitle{Constrained {6-DoF} grasp generation on complex shapes for improved dual-arm manipulation}.
In: \bbtitle{2024 IEEE/RSJ International Conference on Intelligent Robots and Systems (IROS)},
pp. \bfpage{7344}--\blpage{7350}
(\byear{2024})
\end{bchapter}
\endbibitem

%%% 27
\bibitem[\protect\citeauthoryear{Chi et~al.}{2023}]{chi2023diffusionpolicy}
\begin{bchapter}
\bauthor{\bsnm{Chi}, \binits{C.}}, \betal:
\bctitle{Diffusion policy: Visuomotor policy learning via action diffusion}.
In: \bbtitle{Robotics: Science and Systems (RSS)}
(\byear{2023})
\end{bchapter}
\endbibitem

%%% 28
\bibitem[\protect\citeauthoryear{Mo et~al.}{2019}]{mo2019partnet}
\begin{bchapter}
\bauthor{\bsnm{Mo}, \binits{K.}},
\bauthor{\bsnm{Zhu}, \binits{S.}},
\bauthor{\bsnm{Chang}, \binits{A.X.}},
\bauthor{\bsnm{Yi}, \binits{L.}},
\bauthor{\bsnm{Tripathi}, \binits{S.}},
\bauthor{\bsnm{Guibas}, \binits{L.J.}},
\bauthor{\bsnm{Su}, \binits{H.}}:
\bctitle{{PartNet}: A large-scale benchmark for fine-grained and hierarchical part-level {3D} object understanding}.
In: \bbtitle{IEEE/CVF Conference on Computer Vision and Pattern Recognition (CVPR)},
pp. \bfpage{909}--\blpage{918}
(\byear{2019})
\end{bchapter}
\endbibitem

%%% 29
\bibitem[\protect\citeauthoryear{Yin et~al.}{2025}]{geng2025partinstruct}
\begin{bchapter}
\bauthor{\bsnm{Yin}, \binits{Y.}},
\bauthor{\bsnm{Han}, \binits{Z.}},
\bauthor{\bsnm{Aarya}, \binits{S.}},
\bauthor{\bsnm{Xu}, \binits{S.}},
\bauthor{\bsnm{Wang}, \binits{J.}},
\bauthor{\bsnm{Peng}, \binits{J.}},
\bauthor{\bsnm{Wang}, \binits{A.}},
\bauthor{\bsnm{Yuille}, \binits{A.}},
\bauthor{\bsnm{Shu}, \binits{T.}}:
\bctitle{{PartInstruct}: Part-level instruction following for fine-grained robot manipulation}.
In: \bbtitle{Proceedings of Robotics: Science and Systems (RSS)}
(\byear{2025})
\end{bchapter}
\endbibitem

%%% 30
\bibitem[\protect\citeauthoryear{Radford et~al.}{2021}]{radford2021clip}
\begin{bchapter}
\bauthor{\bsnm{Radford}, \binits{A.}},
\bauthor{\bsnm{Kim}, \binits{J.W.}},
\bauthor{\bsnm{Hallacy}, \binits{C.}},
\bauthor{\bsnm{Ramesh}, \binits{A.}},
\bauthor{\bsnm{Goh}, \binits{G.}},
\bauthor{\bsnm{Agarwal}, \binits{S.}},
\bauthor{\bsnm{Sastry}, \binits{G.}},
\bauthor{\bsnm{Askell}, \binits{A.}},
\bauthor{\bsnm{Mishkin}, \binits{P.}},
\bauthor{\bsnm{Clark}, \binits{J.}},
\bauthor{\bsnm{Krueger}, \binits{G.}},
\bauthor{\bsnm{Sutskever}, \binits{I.}}:
\bctitle{Learning transferable visual models from natural language supervision}.
In: \bbtitle{Proceedings of the 38th International Conference on Machine Learning (ICML)}.
\bsertitle{Proceedings of Machine Learning Research},
vol. \bseriesno{139},
pp. \bfpage{8748}--\blpage{8763}
(\byear{2021})
\end{bchapter}
\endbibitem

%%% 31
\bibitem[\protect\citeauthoryear{Oquab et~al.}{2024}]{oquab2024dinov2}
\begin{botherref}
\oauthor{\bsnm{Oquab}, \binits{M.}},
\oauthor{\bsnm{Darcet}, \binits{T.}},
\oauthor{\bsnm{Moutakanni}, \binits{T.}},
\oauthor{\bsnm{Vo}, \binits{H.}},
\oauthor{\bsnm{Szafraniec}, \binits{M.}},
\oauthor{\bsnm{Khalidov}, \binits{V.}},
\oauthor{\bsnm{Fernandez}, \binits{P.}},
\oauthor{\bsnm{Haziza}, \binits{D.}},
\oauthor{\bsnm{Massa}, \binits{F.}},
\oauthor{\bsnm{El-Nouby}, \binits{A.}},
\oauthor{\bsnm{Assran}, \binits{M.}},
\oauthor{\bsnm{Ballas}, \binits{N.}},
\oauthor{\bsnm{Galuba}, \binits{W.}},
\oauthor{\bsnm{Howes}, \binits{R.}},
\oauthor{\bsnm{Huang}, \binits{P.-Y.}},
\oauthor{\bsnm{Li}, \binits{S.-W.}},
\oauthor{\bsnm{Misra}, \binits{I.}},
\oauthor{\bsnm{Rabbat}, \binits{M.}},
\oauthor{\bsnm{Sharma}, \binits{V.}},
\oauthor{\bsnm{Synnaeve}, \binits{G.}},
\oauthor{\bsnm{Xu}, \binits{H.}},
\oauthor{\bsnm{J{\'e}gou}, \binits{H.}},
\oauthor{\bsnm{Mairal}, \binits{J.}},
\oauthor{\bsnm{Labatut}, \binits{P.}},
\oauthor{\bsnm{Joulin}, \binits{A.}},
\oauthor{\bsnm{Bojanowski}, \binits{P.}}:
{DINOv2}: Learning robust visual features without supervision.
Transactions on Machine Learning Research
(2024)
\end{botherref}
\endbibitem

%%% 32
\bibitem[\protect\citeauthoryear{Zitkovich et~al.}{2023}]{brohan2023rt2}
\begin{bchapter}
\bauthor{\bsnm{Zitkovich}, \binits{B.}},
\bauthor{\bsnm{Yu}, \binits{T.}},
\bauthor{\bsnm{Xu}, \binits{S.}},
\bauthor{\bsnm{Xu}, \binits{P.}},
\bauthor{\bsnm{Xiao}, \binits{T.}},
\bauthor{\bsnm{Xia}, \binits{F.}},
\bauthor{\bsnm{Wu}, \binits{J.}},
\bauthor{\bsnm{Wohlhart}, \binits{P.}},
\bauthor{\bsnm{Welker}, \binits{S.}},
\bauthor{\bsnm{Wahid}, \binits{A.}},
\bauthor{\bsnm{Vuong}, \binits{Q.}},
\bauthor{\bsnm{Vanhoucke}, \binits{V.}},
\bauthor{\bsnm{Tran}, \binits{H.}},
\bauthor{\bsnm{Soricut}, \binits{R.}},
\bauthor{\bsnm{Singh}, \binits{A.}},
\bauthor{\bsnm{Singh}, \binits{J.}},
\bauthor{\bsnm{Sermanet}, \binits{P.}},
\bauthor{\bsnm{Sanketi}, \binits{P.R.}},
\bauthor{\bsnm{Salazar}, \binits{G.}},
\bauthor{\bsnm{Ryoo}, \binits{M.S.}},
\bauthor{\bsnm{Reymann}, \binits{K.}},
\bauthor{\bsnm{Rao}, \binits{K.}},
\bauthor{\bsnm{Pertsch}, \binits{K.}},
\bauthor{\bsnm{Mordatch}, \binits{I.}},
\bauthor{\bsnm{Michalewski}, \binits{H.}},
\bauthor{\bsnm{Lu}, \binits{Y.}},
\bauthor{\bsnm{Levine}, \binits{S.}},
\bauthor{\bsnm{Lee}, \binits{L.}},
\bauthor{\bsnm{Lee}, \binits{T.-W.E.}},
\bauthor{\bsnm{Leal}, \binits{I.}},
\bauthor{\bsnm{Kuang}, \binits{Y.}},
\bauthor{\bsnm{Kalashnikov}, \binits{D.}},
\bauthor{\bsnm{Julian}, \binits{R.}},
\bauthor{\bsnm{Joshi}, \binits{N.J.}},
\bauthor{\bsnm{Irpan}, \binits{A.}},
\bauthor{\bsnm{Ichter}, \binits{B.}},
\bauthor{\bsnm{Hsu}, \binits{J.}},
\bauthor{\bsnm{Herzog}, \binits{A.}},
\bauthor{\bsnm{Hausman}, \binits{K.}},
\bauthor{\bsnm{Gopalakrishnan}, \binits{K.}},
\bauthor{\bsnm{Fu}, \binits{C.}},
\bauthor{\bsnm{Florence}, \binits{P.}},
\bauthor{\bsnm{Finn}, \binits{C.}},
\bauthor{\bsnm{Dubey}, \binits{K.A.}},
\bauthor{\bsnm{Driess}, \binits{D.}},
\bauthor{\bsnm{Ding}, \binits{T.}},
\bauthor{\bsnm{Choromanski}, \binits{K.M.}},
\bauthor{\bsnm{Chen}, \binits{X.}},
\bauthor{\bsnm{Chebotar}, \binits{Y.}},
\bauthor{\bsnm{Carbajal}, \binits{J.}},
\bauthor{\bsnm{Brown}, \binits{N.}},
\bauthor{\bsnm{Brohan}, \binits{A.}},
\bauthor{\bsnm{Arenas}, \binits{M.G.}},
\bauthor{\bsnm{Han}, \binits{K.}}:
\bctitle{{RT-2}: Vision-language-action models transfer web knowledge to robotic control}.
In: \bbtitle{Proceedings of The 7th Conference on Robot Learning (CoRL)}.
\bsertitle{Proceedings of Machine Learning Research},
vol. \bseriesno{229},
pp. \bfpage{2165}--\blpage{2183}
(\byear{2023})
\end{bchapter}
\endbibitem

%%% 33
\bibitem[\protect\citeauthoryear{James et~al.}{2020}]{james2020rlbench}
\begin{barticle}
\bauthor{\bsnm{James}, \binits{S.}},
\bauthor{\bsnm{Ma}, \binits{Z.}},
\bauthor{\bsnm{Arrojo}, \binits{D.R.}},
\bauthor{\bsnm{Davison}, \binits{A.J.}}:
\batitle{{RLBench}: The robot learning benchmark \& learning environment}.
\bjtitle{IEEE Robotics and Automation Letters}
\bvolume{5}(\bissue{2}),
\bfpage{3019}--\blpage{3026}
(\byear{2020})
\end{barticle}
\endbibitem

%%% 34
\bibitem[\protect\citeauthoryear{Xie et~al.}{2024}]{xie2024decomposing}
\begin{bchapter}
\bauthor{\bsnm{Xie}, \binits{A.}},
\bauthor{\bsnm{Lee}, \binits{L.}},
\bauthor{\bsnm{Xiao}, \binits{T.}},
\bauthor{\bsnm{Finn}, \binits{C.}}:
\bctitle{Decomposing the generalization gap in imitation learning for visual robotic manipulation}.
In: \bbtitle{2024 IEEE International Conference on Robotics and Automation (ICRA)},
pp. \bfpage{3153}--\blpage{3160}
(\byear{2024})
\end{bchapter}
\endbibitem

%%% 35
\bibitem[\protect\citeauthoryear{Li et~al.}{2025}]{li2025simpler}
\begin{bchapter}
\bauthor{\bsnm{Li}, \binits{X.}},
\bauthor{\bsnm{Hsu}, \binits{K.}},
\bauthor{\bsnm{Gu}, \binits{J.}},
\bauthor{\bsnm{Mees}, \binits{O.}},
\bauthor{\bsnm{Pertsch}, \binits{K.}},
\bauthor{\bsnm{Walke}, \binits{H.R.}},
\bauthor{\bsnm{Fu}, \binits{C.}},
\bauthor{\bsnm{Lunawat}, \binits{I.}},
\bauthor{\bsnm{Sieh}, \binits{I.}},
\bauthor{\bsnm{Kirmani}, \binits{S.}},
\bauthor{\bsnm{Levine}, \binits{S.}},
\bauthor{\bsnm{Wu}, \binits{J.}},
\bauthor{\bsnm{Finn}, \binits{C.}},
\bauthor{\bsnm{Su}, \binits{H.}},
\bauthor{\bsnm{Vuong}, \binits{Q.}},
\bauthor{\bsnm{Xiao}, \binits{T.}}:
\bctitle{Evaluating real-world robot manipulation policies in simulation}.
In: \bbtitle{Proceedings of The 8th Conference on Robot Learning (CoRL)}.
\bsertitle{Proceedings of Machine Learning Research},
vol. \bseriesno{270},
pp. \bfpage{3705}--\blpage{3728}
(\byear{2025})
\end{bchapter}
\endbibitem

%%% 36
\bibitem[\protect\citeauthoryear{Torne et~al.}{2024}]{torne2024rialto}
\begin{bchapter}
\bauthor{\bsnm{Torne}, \binits{M.}},
\bauthor{\bsnm{Simeonov}, \binits{A.}},
\bauthor{\bsnm{Li}, \binits{Z.}},
\bauthor{\bsnm{Chan}, \binits{A.}},
\bauthor{\bsnm{Chen}, \binits{T.}},
\bauthor{\bsnm{Gupta}, \binits{A.}},
\bauthor{\bsnm{Agrawal}, \binits{P.}}:
\bctitle{Reconciling reality through simulation: A real-to-sim-to-real approach for robust manipulation}.
In: \bbtitle{Proceedings of Robotics: Science and Systems (RSS)}
(\byear{2024})
\end{bchapter}
\endbibitem

%%% 37
\bibitem[\protect\citeauthoryear{Kerbl et~al.}{2023}]{kerbl20233dgs}
\begin{botherref}
\oauthor{\bsnm{Kerbl}, \binits{B.}},
\oauthor{\bsnm{Kopanas}, \binits{G.}},
\oauthor{\bsnm{Leimk{\"u}hler}, \binits{T.}},
\oauthor{\bsnm{Drettakis}, \binits{G.}}:
3{D} {G}aussian {S}platting for real-time radiance field rendering.
ACM Transactions on Graphics
\textbf{42}(4)
(2023)
\end{botherref}
\endbibitem

%%% 38
\bibitem[\protect\citeauthoryear{Li et~al.}{2024}]{li2024robogsim}
\begin{botherref}
\oauthor{\bsnm{Li}, \binits{X.}},
\oauthor{\bsnm{Li}, \binits{J.}},
\oauthor{\bsnm{Zhang}, \binits{Z.}},
\oauthor{\bsnm{Zhang}, \binits{R.}},
\oauthor{\bsnm{Jia}, \binits{F.}},
\oauthor{\bsnm{Wang}, \binits{T.}},
\oauthor{\bsnm{Fan}, \binits{H.}},
\oauthor{\bsnm{Tseng}, \binits{K.-K.}},
\oauthor{\bsnm{Wang}, \binits{R.}}:
{RoboGSim}: A real2sim2real robotic {G}aussian {S}platting simulator.
arXiv preprint arXiv:2411.11839
(2024)
\end{botherref}
\endbibitem

%%% 39
\bibitem[\protect\citeauthoryear{Han et~al.}{2025}]{han2025re3sim}
\begin{botherref}
\oauthor{\bsnm{Han}, \binits{X.}},
\oauthor{\bsnm{Liu}, \binits{M.}},
\oauthor{\bsnm{Chen}, \binits{Y.}},
\oauthor{\bsnm{Yu}, \binits{J.}},
\oauthor{\bsnm{Lyu}, \binits{X.}},
\oauthor{\bsnm{Tian}, \binits{Y.}},
\oauthor{\bsnm{Wang}, \binits{B.}},
\oauthor{\bsnm{Zhang}, \binits{W.}},
\oauthor{\bsnm{Pang}, \binits{J.}}:
{Re3Sim}: Generating high-fidelity simulation data via {3D}-photorealistic real-to-sim for robotic manipulation.
arXiv preprint arXiv:2502.08645
(2025)
\end{botherref}
\endbibitem

%%% 40
\bibitem[\protect\citeauthoryear{Zhang et~al.}{2025}]{zhang2025real2sim_softbody}
\begin{botherref}
\oauthor{\bsnm{Zhang}, \binits{K.}},
\oauthor{\bsnm{Sha}, \binits{S.}},
\oauthor{\bsnm{Jiang}, \binits{H.}},
\oauthor{\bsnm{Loper}, \binits{M.}},
\oauthor{\bsnm{Song}, \binits{H.}},
\oauthor{\bsnm{Cai}, \binits{G.}},
\oauthor{\bsnm{Xu}, \binits{Z.}},
\oauthor{\bsnm{Hu}, \binits{X.}},
\oauthor{\bsnm{Zheng}, \binits{C.}},
\oauthor{\bsnm{Li}, \binits{Y.}}:
Real-to-sim robot policy evaluation with {G}aussian {S}platting simulation of soft-body interactions.
arXiv preprint arXiv:2511.04665
(2025)
\end{botherref}
\endbibitem

%%% 41
\bibitem[\protect\citeauthoryear{Sedlacek et~al.}{2025}]{sedlacek2025realm}
\begin{botherref}
\oauthor{\bsnm{Sedlacek}, \binits{M.}},
\oauthor{\bsnm{Yefanov}, \binits{P.}},
\oauthor{\bsnm{Ponimatkin}, \binits{G.}},
\oauthor{\bsnm{Bardhan}, \binits{J.}},
\oauthor{\bsnm{Pilc}, \binits{S.}},
\oauthor{\bsnm{Fourmy}, \binits{M.}},
\oauthor{\bsnm{Kazakos}, \binits{E.}},
\oauthor{\bsnm{Snoek}, \binits{C.G.M.}},
\oauthor{\bsnm{Sivic}, \binits{J.}},
\oauthor{\bsnm{Petrik}, \binits{V.}}:
{REALM}: A real-to-sim validated benchmark for generalization in robotic manipulation.
arXiv preprint arXiv:2512.19562
(2025)
\end{botherref}
\endbibitem

%%% 42
\bibitem[\protect\citeauthoryear{Nasiriany et~al.}{2024}]{nasiriany2024robocasa}
\begin{bchapter}
\bauthor{\bsnm{Nasiriany}, \binits{S.}},
\bauthor{\bsnm{Maddukuri}, \binits{A.}},
\bauthor{\bsnm{Zhang}, \binits{L.}},
\bauthor{\bsnm{Parikh}, \binits{A.}},
\bauthor{\bsnm{Lo}, \binits{A.}},
\bauthor{\bsnm{Joshi}, \binits{A.}},
\bauthor{\bsnm{Mandlekar}, \binits{A.}},
\bauthor{\bsnm{Zhu}, \binits{Y.}}:
\bctitle{{RoboCasa}: Large-scale simulation of everyday tasks for generalist robots}.
In: \bbtitle{Proceedings of Robotics: Science and Systems (RSS)}
(\byear{2024})
\end{bchapter}
\endbibitem

\end{thebibliography}

\newpage
\begin{appendices}
\section*{Appendix Overview}

This appendix provides additional details on MetaFine, extended experimental results, and architectural analyses omitted from the main body due to space limitations:
\begin{itemize}
    \item Section~\ref{supp:sec1} details MetaFine's framework design: the compositional task graph, the extensible asset library, the atomic skill vocabulary, the three-dimensional evaluation protocol (understanding, perception, behavior), perturbation definitions, and the seven evaluated policies.
    
    \item Section~\ref{supp:sec3} reports full numerical results across all tasks and policies, including perceptual robustness, instruction perturbation, stage-wise decomposition of long-horizon tasks, viewpoint adaptation ablations, and the coarse- versus fine-grained comparison.
    
    \item Section~\ref{appendix_method_sec} elaborates on the task graph's operational semantics, the benchmark adapter pipeline for absorbing RoboTwin, ManiSkill, and LIBERO, and the task generation stage.
    
    \item Section~\ref{app:hybrid-eval} formalizes the hybrid real--sim evaluation protocol with the PPI estimator, 3D Gaussian Splatting-based real2sim registration, and an empirical validation of the variance reduction over hardware-only estimation.
    
    \item Section~\ref{supp:sec4} describes the multi-scale cross-attention encoder that replaces $\pi_{0.5}$'s single-scale visual frontend.
    
    \item Section~\ref{supp:sec5} analyzes trajectories of different action generation paradigms, covering convergence versus drift, behavioral arrest versus sub-task transition, and the dissociation between trajectory quality and task success.
    
    \item Section~\ref{related_work} surveys related work on fine-grained manipulation, visual-motor policies, manipulation benchmarks, and real-to-sim evaluation.
\end{itemize}

\section{MetaFine Framework Design and Skill Definitions}
\label{supp:sec1}

\subsection{Framework Architecture}
\label{sec:appendix_framework}

MetaFine is structured as a meta-evaluation framework rather than a fixed task collection. Three independently extensible components form its core: an \emph{atomic skill vocabulary} that defines the primitive units of fine-grained manipulation, a \emph{compositional task graph} that governs how these primitives combine into evaluation scenarios, and an \emph{extensible asset library} that provides the physical substrate for task instantiation. The three components are deliberately decoupled, so that introducing a new skill, incorporating a new object, or composing a new task modifies only the relevant component. This separation enables MetaFine to function as a continuously expandable evaluation ecosystem.

\textbf{Compositional task graph.} The task graph is the generative core of MetaFine. Each node represents an atomic fine-grained skill specified by a tuple $(p, q, \mathcal{C})$, where $p$ denotes the preconditions required to initiate the skill, $q$ the postconditions that define successful completion, and $\mathcal{C}$ the physical constraints maintained throughout execution (the complete vocabulary appears in Section~\ref{sec:appendix_skills}). For example, \textsc{Grasp Part} requires that the target object is reachable and visible ($p$), terminates when stable contact is established at the specified part ($q$), and enforces that the contact region matches the instructed part rather than any graspable surface ($\mathcal{C}$).

Composability follows a \emph{postcondition–precondition compatibility} principle: skill $s_i$ may precede skill $s_j$ if and only if $q_i \Rightarrow p_j$. Valid compositions are therefore derived from the formal specifications themselves, without manual enumeration of permissible transitions. Edges encode three dependency types:

\begin{itemize}
    \item \textbf{Sequential} edges connect skills whose postconditions satisfy the subsequent skill's preconditions, forming linear chains. The peg-in-hole task, for instance, sequentially composes \textsc{Grasp Part} $\rightarrow$ \textsc{Align} $\rightarrow$ \textsc{Insert}.
    \item \textbf{Conditional} edges introduce runtime branching. In the sorting task, \textsc{Grasp Part} is followed by one of several \textsc{Move To} instances depending on the color attribute of the grasped object.
    \item \textbf{Parallel} edges specify constraints that must be simultaneously maintained, such as preserving orientation compliance while executing a constrained rotational motion.
\end{itemize}

\begin{figure}[!t]
    \centering
    \includegraphics[width=1\textwidth]{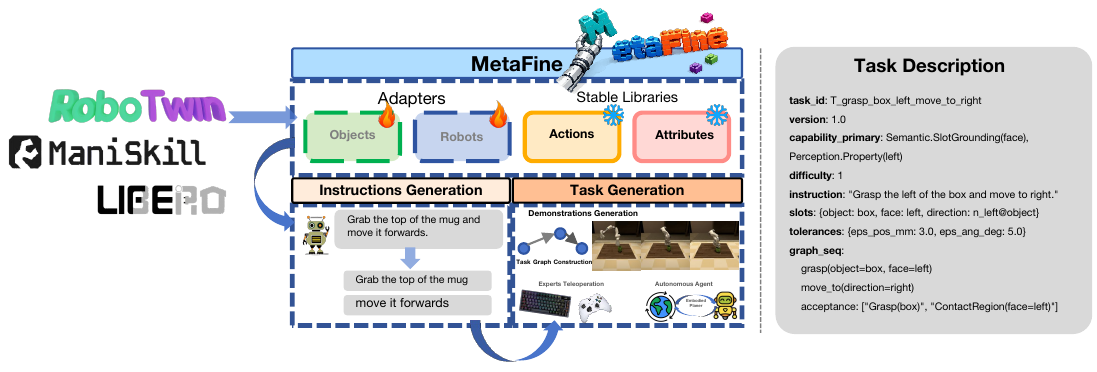}
    \caption{MetaFine converts heterogeneous benchmarks (RoboTwin, ManiSkill, and LIBERO) into a unified representation. Adapters standardize objects, robots, actions, and attributes into stable libraries, from which fine-grained skill graphs and evaluation criteria are automatically generated.}
    \label{fig4}
\end{figure}

Given a vocabulary of $N$ atomic skills, this formalism in principle generates a large open-ended task space through composition. A concrete evaluation task is instantiated by traversing a subgraph of the task graph and binding each skill node to specific objects, parts, and constraint parameters drawn from the asset library. The output is a complete specification comprising the natural-language instruction, the ordered skill sequence with edge types, per-stage acceptance criteria, and tolerance parameters (see the Task Description format in Fig.~\ref{fig4}).

Table~\ref{tab:task_specs} lists the complete set of tasks used throughout MetaFine's evaluation, organized by primary atomic skill. Each task instantiates one or more atomic skills on specific objects and parts drawn from the asset library, and several tasks span multiple object variants to enable within-skill generalization assessment. The multi-skill tasks at the bottom of the table are sequential compositions that traverse two or more atomic skills, and their stage-wise decompositions are reported in Section~\ref{sec:appendix_full_results}. All task specifications, demonstrations, and associated assets are publicly released to support reproducibility and to serve as a standardized data resource for the community.

\textbf{Extensibility at three levels.} Adding a new \emph{atomic skill} requires specifying its $(p, q, \mathcal{C})$ tuple along with acceptance criteria and tolerance parameters. Once registered, the new skill becomes immediately available for composition with all compatible existing skills. To reduce the formalization overhead, MetaFine integrates an automated agent that drafts a candidate specification from a natural-language description (e.g., ``slide the object along a surface edge until contact with the boundary''); the user reviews and refines the draft.

Adding a new \emph{asset} draws on the structured object annotation schema described in Section~\ref{sec:appendix_assets} and a lightweight annotation toolkit that completes the full pipeline in under 10 seconds per object. Once annotated, a new object automatically inherits evaluation coverage for all skills whose constraints are compatible with its annotated parts and poses.

Adding a new \emph{task} consists of traversing a subgraph of the task graph and binding skill nodes to specific objects and parameters. An automated agent can decompose a high-level task description (e.g., ``assemble the pen by inserting the refill and attaching the cap'') into a candidate skill sequence and generate a draft specification, which the user validates.

\textbf{Evaluation execution pipeline.} A single evaluation trial proceeds through six stages: (1)~a task specification is loaded, defining the skill sequence, acceptance criteria, and tolerance parameters; (2)~the simulation scene is initialized by placing the relevant objects from the asset library according to the task configuration; (3)~if the evaluation targets perceptual robustness, geometric or photometric perturbations are injected at the specified severity level prior to policy execution (Section~\ref{sec:appendix_perturbations}); (4)~the policy executes under the (possibly perturbed) conditions; (5)~acceptance criteria are evaluated independently at each skill stage, enabling stage-wise success assessment; (6)~understanding indicators, perceptual robustness scores, and behavioral quality measures are recorded in parallel. Metrics are aggregated by averaging over multiple trials per task configuration and then across task configurations for summary statistics.

\subsection{Extensible Asset Library}
\label{sec:appendix_assets}

\begin{figure}[ht]
    \centering
    \includegraphics[width=\textwidth]{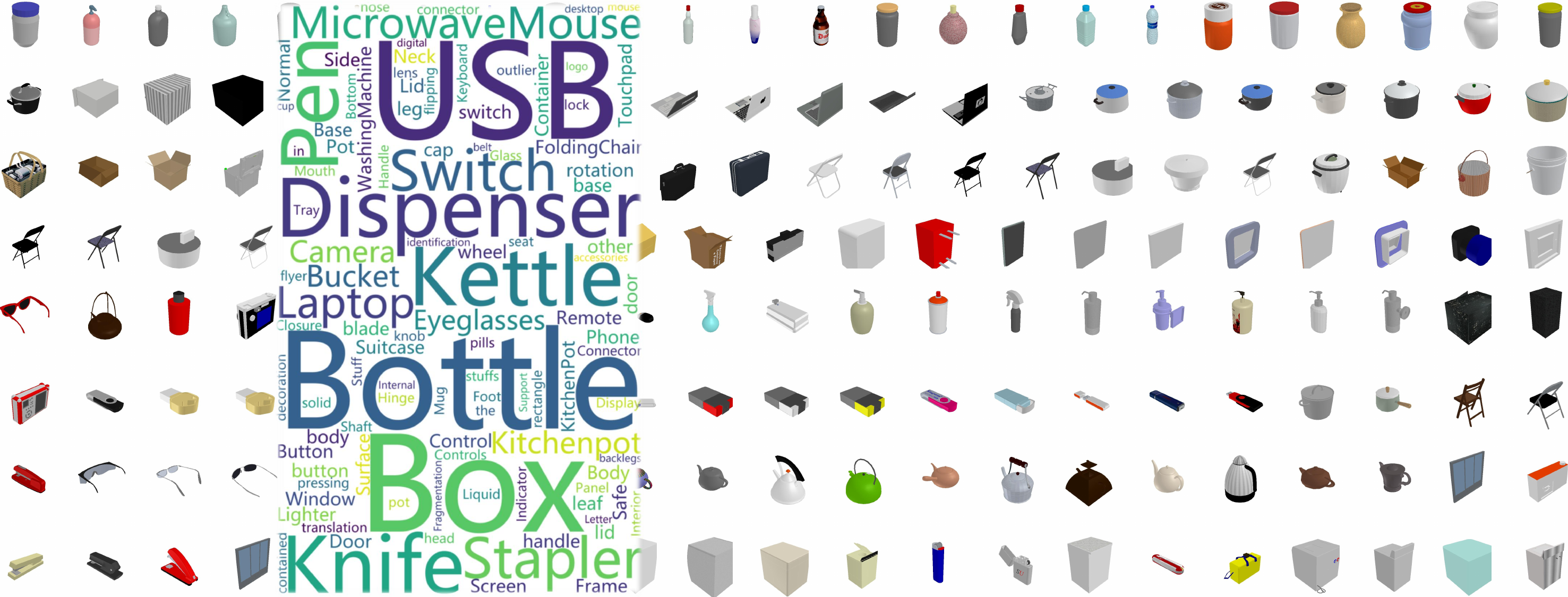}
    \caption{Overview of the extensible asset library. The library currently comprises 431 annotated objects spanning diverse everyday manipulation categories, with 1{,}078 labeled parts and 4{,}312 candidate grasp poses. Each object is annotated with part-level segmentation, per-part grasp poses, and manipulation constraints. A lightweight annotation toolkit completes the full annotation pipeline in under 10 seconds per object, enabling continuous community-driven expansion.}
    \label{fig:assets}
\end{figure}

The asset library provides the physical substrate over which the compositional task graph operates. Each abstract skill composition is instantiated into a concrete evaluation scenario by binding skill nodes to specific objects, parts, and constraint parameters drawn from this library. The library currently comprises 431 objects, 1{,}078 labeled parts, and 4{,}312 candidate grasp poses (Fig.~\ref{fig:assets}), spanning a diverse range of everyday manipulation targets including household items, tools, containers, articulated objects, and precision components.

\textbf{Object annotation schema.} Each object is annotated with three layers of information. First, \emph{part-level segmentation} identifies interactable regions on the object surface (e.g., handle, cap, body, lid, button), providing spatial targets for part-constrained skills such as \textsc{Grasp Part} and \textsc{Press Part}. Second, \emph{grasp pose candidates} specify feasible grasp configurations for each annotated part, with each pose defined by position, orientation, and approach direction in the object frame; objects carry approximately ten candidate poses on average. Third, \emph{manipulation constraints} encode directional or orientation requirements relevant to downstream skills, such as the rotation axis for \textsc{Rotate Along} or the hinge axis for \textsc{Open Hinge}. Together, these three layers ensure that a newly added object integrates automatically into any skill composition whose constraints are compatible with its annotated properties.

\textbf{Rapid annotation toolkit.} To minimize the barrier to community-driven expansion, MetaFine provides a lightweight annotation toolkit that streamlines the full pipeline. The toolkit leverages automated part segmentation to generate candidate part boundaries that the annotator refines; grasp pose specification is guided by part normals and surface accessibility; and constraint assignment (rotation axes, hinge axes, sliding directions) is assisted through geometric inference from the object mesh. This semi-automated pipeline reduces the time required to complete the full annotation of a new object---from part segmentation through grasp pose specification and constraint assignment---to under 10 seconds per object on average. The low overhead makes it practical for research groups to contribute objects from their own experimental domains, progressively broadening the ecological validity of the evaluation ecosystem without centralized curation.

\textbf{Diversity and coverage.} The library covers the principal object categories encountered in everyday fine-grained manipulation. Objects span a range of geometric complexities (from simple prismatic shapes to articulated multi-part assemblies), material properties (rigid, deformable, transparent), and functional categories (containers, tools, switches, connectors, stationery). Part annotations capture both macro-level distinctions (handle vs.\ body) and finer subdivisions (button vs.\ surrounding panel), enabling evaluation at multiple levels of constraint granularity. The grasp pose candidates include both power grasps and precision grasps. As the library grows through community contributions, the compositional task graph automatically gains access to new object--skill combinations without modifications to the task generation or evaluation infrastructure.

\begin{table}[h]
\centering
\caption{Task specifications and demonstration counts by primary atomic skill. ``Variants’’ indicates the number of object instances or instruction variants covered. All demonstrations and task configurations are publicly released.}
\label{tab:task_specs}
\vspace{0.2em}
\small
\setlength{\tabcolsep}{5pt}
\begin{tabular}{llrr}
\toprule
\textbf{Primary Skill} & \textbf{Task} & \textbf{Variants} & \textbf{\# Demos} \\
\midrule
\multirow{5}{*}{Grasp Part}
 & Grasp mug handle           & 1 & 1{,}000 \\
 & Grasp bottle cap           & 1 & 1{,}000 \\
 & Grasp cabinet handle       & 1 & 1{,}000 \\
 & Grasp box lid              & 1 & 1{,}000 \\
 & Grasp box handle           & 1 & 1{,}000 \\
\midrule
Slide Along    & Slide along                & 5  & 5{,}000  \\
\midrule
\multirow{3}{*}{Press Part}
 & Press switch               & 5 & 5{,}000 \\
 & Press stapler head         & 3 & 3{,}000 \\
 & Press stapler top          & 3 & 3{,}000 \\
\midrule
Toggle Part    & Toggle switch              & 6  & 6{,}000  \\
\midrule
Rotate Along   & Rotate button clockwise    & 5  & 5{,}000  \\
\midrule
Open Hinge     & Open box with lid          & 1  & 1{,}000  \\
\midrule
Flip           & Flip                       & 3  & 3{,}000  \\
\midrule
Move To        & Grasp part and move        & 12 & 12{,}000 \\
\midrule
\multirow{5}{*}{Multi-Skill}
 & Peg in hole                                & 1                 & 1{,}000  \\
 & Plug in charger                            & 1                 & 1{,}000  \\
 & Stack pyramid                              & 1                 & 1{,}000 \\
 & Sort colored cubes into boxes\textsuperscript{$\dagger$}  & 3 (green/red/blue) & 3{,}000  \\
 & Grasp character to assemble\textsuperscript{$\ddagger$}   & 3 (I/L/T)          & 3{,}000  \\
\bottomrule
\end{tabular}
\raggedright
\footnotesize
\textsuperscript{$\dagger$}\,``Place the \{color\} cube in the left box and the others in the right box.'' \quad
\textsuperscript{$\ddagger$}\,``Grasp character \{I/L/T\} for assembly.''
\end{table}

\subsection{Atomic Fine-Grained Skill Vocabulary}
\label{sec:appendix_skills}

The atomic skill vocabulary defines the primitive manipulation units from which all MetaFine evaluation tasks are composed. Each skill encodes a physically distinct fine-grained constraint that cannot be reduced to coarse object-level interaction: executing the skill correctly requires satisfying specific geometric, semantic, or dynamic conditions throughout the manipulation process. The vocabulary is organized around three design principles.

\textbf{Constraint-centered definition.} Each atomic skill foregrounds a primary fine-grained constraint that characterizes its core manipulation demand. \textsc{Grasp Part} centers on a part-level contact region; \textsc{Rotate Along} centers on directional compliance along a specified axis. Every skill involves multiple interacting physical requirements in practice, but the vocabulary is structured so that each foregrounds a distinct constraint, making it straightforward to associate performance variation with specific manipulation demands.

\textbf{Compositional completeness.} The vocabulary is chosen so that common real-world fine-grained tasks---tool use, precision assembly, articulated object operation---can be expressed as compositions of atomic skills. The task graph compositions in Section~\ref{sec:appendix_framework} demonstrate this coverage.

\textbf{Open extensibility.} The vocabulary is not closed. New atomic skills can be added by specifying their formal definition and acceptance criteria; once registered, they become available for composition through the postcondition--precondition compatibility mechanism.

\vspace{0.5em}
The current vocabulary comprises the following ten atomic skills, each illustrated in Supplementary Fig.~\ref{fig:skill_demos}.

\textbf{Grasp Part}
The end-effector must establish stable contact with a \emph{specified part} of the target object, rather than any graspable surface.
\emph{Core constraint:} the contact region must fall within the annotated boundary of the instructed part (e.g., handle, cap, body); grasps that achieve stable hold but contact an incorrect part are counted as failures.
\emph{Representative instruction:} ``Grasp the cap of the bottle.''
\emph{Acceptance criterion:} stable grasp established, with the contact region verified against the annotated part segmentation.

\textbf{Press Part}
The end-effector must apply a downward pressing force on a specific interactable region such as a button, key, or switch surface.
\emph{Core constraint:} contact must lie within the spatial extent of the target pressable element, with force applied predominantly along the element's actuation axis; off-axis contact or contact outside the target region constitutes failure.
\emph{Representative instruction:} ``Press the power button on the device.''
\emph{Acceptance criterion:} the target part is actuated (e.g., button depressed) through contact at the correct region.

\textbf{Toggle Part}
The end-effector must actuate a movable part through its full range of motion, such as flipping a switch or rotating a lever.
\emph{Core constraint:} the actuation trajectory must conform to the part's kinematic degrees of freedom (revolute joint axis for a lever, prismatic axis for a slider) and traverse sufficient displacement to transition the part between its discrete states.
\emph{Representative instruction:} ``Toggle the switch to the on position.''
\emph{Acceptance criterion:} the target part transitions to the specified state, verified through joint-angle or contact-state monitoring.

\textbf{Rotate Along}
The end-effector must rotate a grasped object or articulated part along a specified axis while maintaining directional compliance throughout.
\emph{Core constraint:} the instantaneous rotation axis must remain within an angular tolerance $\epsilon_{\text{axis}}$ of the designated axis, and the cumulative signed rotation must match the instructed direction (clockwise vs.\ counterclockwise) and target angle.
\emph{Representative instruction:} ``Rotate the knob clockwise by 90 degrees.''
\emph{Acceptance criterion:} cumulative rotation along the correct axis reaches the target angle within tolerance, with angular deviation from the prescribed axis remaining below $\epsilon_{\text{axis}}$ throughout execution.

\textbf{Slide Along}
The end-effector must move a grasped object or execute a contact-rich sliding motion along a constrained surface or edge.
\emph{Core constraint:} the trajectory must remain within a spatial envelope defined by the guiding geometry (rail, edge, surface), with the normal distance to the guiding surface staying below a clearance threshold $\epsilon_{\text{clear}}$.
\emph{Representative instruction:} ``Slide the drawer open along its rail.''
\emph{Acceptance criterion:} the object displaces along the constrained path by the target distance while maintaining contact with the guiding surface.

\textbf{Open Hinge}
The end-effector must actuate a hinged part (door, lid, cabinet panel) by applying force that produces rotation about the hinge axis.
\emph{Core constraint:} the applied force must produce a net torque about the hinge axis; the contact point and force direction must be compatible with hinge kinematics, since forces directed through or parallel to the hinge axis produce no useful actuation.
\emph{Representative instruction:} ``Open the cabinet door.''
\emph{Acceptance criterion:} the hinged part rotates beyond a minimum angular threshold about its hinge axis.

\textbf{Align}
The end-effector must adjust the pose of a grasped object to match a target configuration defined by a receptacle, slot, or reference frame.
\emph{Core constraint:} the 6-DoF relative pose error must simultaneously satisfy translational tolerance $\epsilon_{\text{pos}}$ and rotational tolerance $\epsilon_{\text{ang}}$; satisfying one without the other is insufficient, since downstream insertion or mating operations require both.
\emph{Representative instruction:} ``Align the peg with the hole.''
\emph{Acceptance criterion:} positional error below $\epsilon_{\text{pos}}$ and angular error below $\epsilon_{\text{ang}}$, evaluated in the target frame.

\textbf{Insert}
The end-effector must guide a grasped object into a receptacle or slot through a constrained entry path.
\emph{Core constraint:} throughout the insertion trajectory, the lateral clearance between the object and receptacle walls must remain positive (no collision), while the object's orientation stays within a narrow angular corridor dictated by the receptacle geometry; the required positional precision is typically on the order of millimeters.
\emph{Representative instruction:} ``Insert the peg into the hole.''
\emph{Acceptance criterion:} the object reaches the target insertion depth within the receptacle, verified through positional monitoring.

\textbf{Move To}
The end-effector must transport a grasped object to a target location specified in the instruction, potentially subject to directional or path constraints.
\emph{Core constraint:} the displacement vector from initial to final object position must align with the instructed direction within an angular tolerance; intermediate path constraints (e.g., maintaining upright orientation during transport to prevent spilling) must be satisfied at every time step.
\emph{Representative instruction:} ``Move the mug to the left side of the table.''
\emph{Acceptance criterion:} the final position falls within the target region and directional compliance is verified throughout the trajectory.

\textbf{Flip}
The end-effector must invert or reorient an object by a specified angular displacement, typically $180^{\circ}$, about an axis parallel to the support surface.
\emph{Core constraint:} the object must undergo a controlled rotation of the target magnitude about the specified axis while the grasp remains stable; the end-effector must coordinate translational compensation with the rotation to prevent the object's center of mass from departing the workspace or colliding with the support surface.
\emph{Representative instruction:} ``Flip the box upside down.''
\emph{Acceptance criterion:} the object's orientation changes by the specified angle about the correct axis, and the object remains stably grasped or placed upon completion.

\vspace{1em}
\noindent These ten atomic skills span the principal fine-grained constraints encountered in everyday manipulation: part-level contact selection (\textsc{Grasp Part}, \textsc{Press Part}, \textsc{Toggle Part}), constrained motion along prescribed paths or axes (\textsc{Rotate Along}, \textsc{Slide Along}, \textsc{Open Hinge}), precision pose adjustment (\textsc{Align}, \textsc{Insert}), and constrained transport and reorientation (\textsc{Move To}, \textsc{Flip}). Through the compositional task graph, these primitives combine to cover a wide and extensible space of fine-grained manipulation tasks.

\clearpage

\begin{figure}[p]
    \centering
    \includegraphics[width=\textwidth]{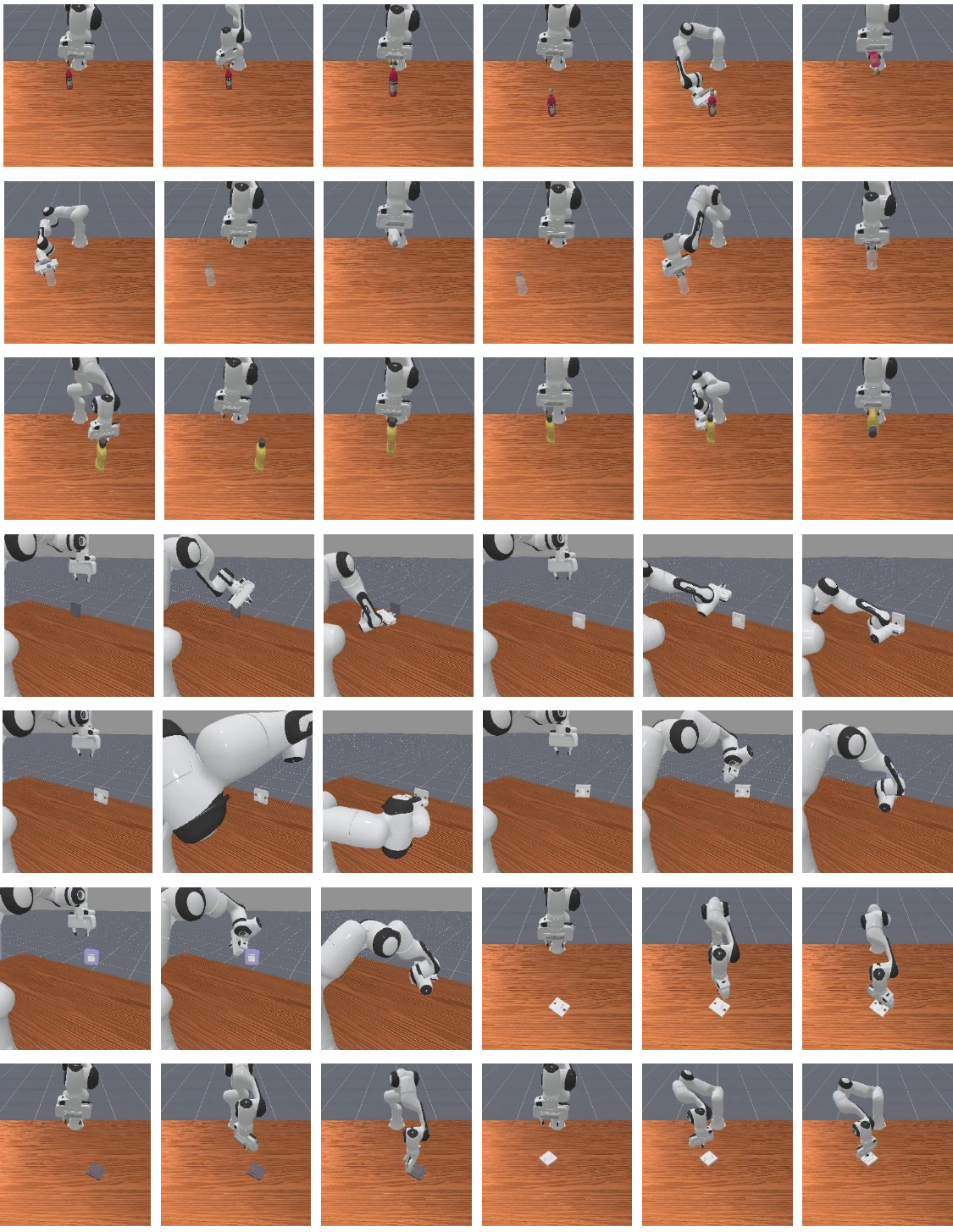}
    \caption{Visualization of atomic fine-grained skills (Part 1). Each panel shows a representative execution sequence for one atomic skill, illustrating the manipulation trajectory from approach through task completion. Multiple object instances are displayed per skill to demonstrate the diversity of the asset library.}
    \label{fig:skill_demos}
\end{figure}

\begin{figure}[p]
    \centering
    \includegraphics[width=\textwidth]{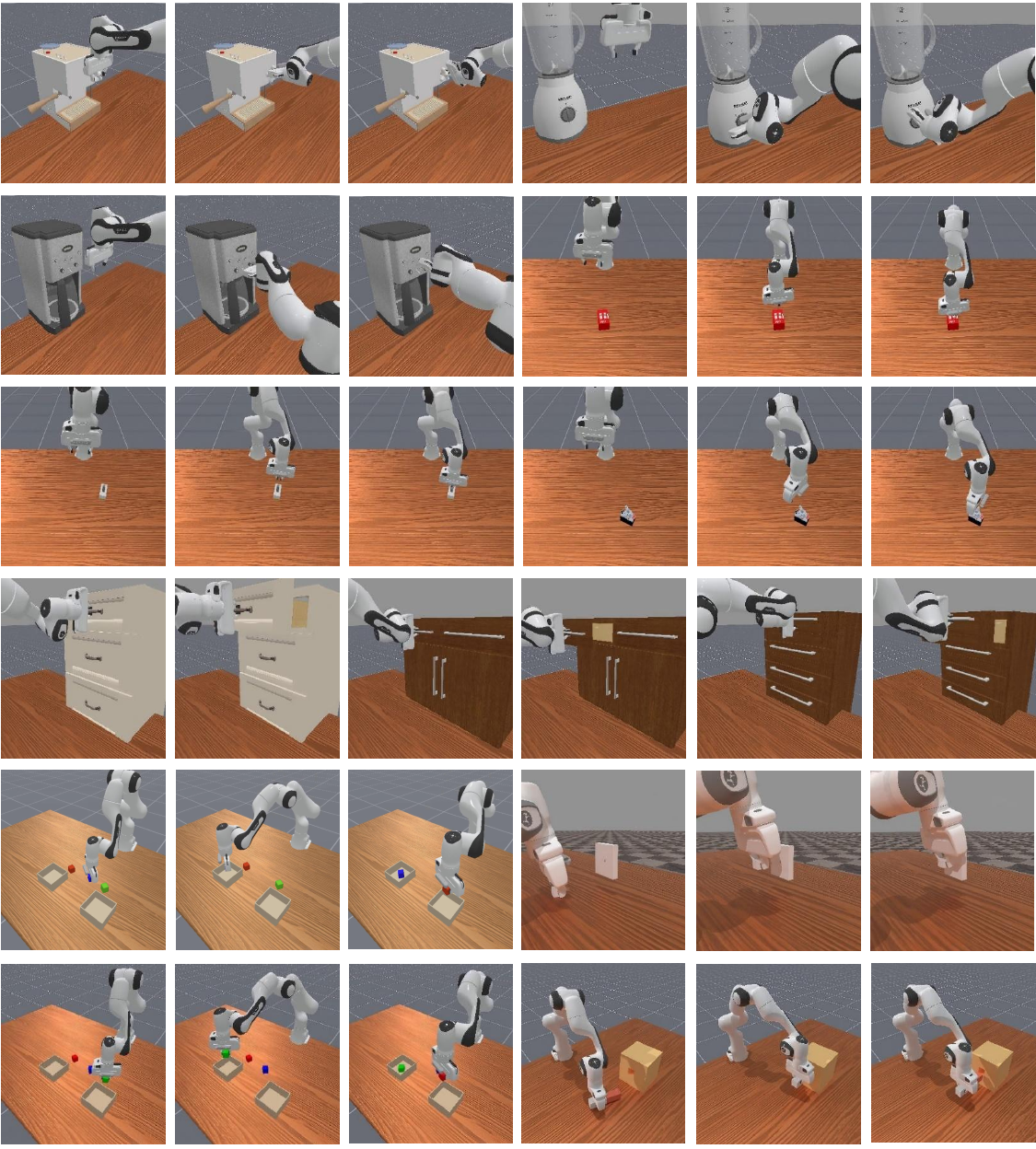}
    \caption{Visualization of atomic fine-grained skills (Part 2, continued).}
    \label{fig:skill_demos_p2}
\end{figure}

\clearpage

\subsubsection{Understanding evaluation}
\label{sec:appendix_understanding}

Fine-grained manipulation instructions encode attribute-level specifications that constrain not only which object to manipulate, but which part to grasp, which direction to move, and which orientation to maintain. The understanding axis tests whether a policy grounds its behavior in these specifications or replays a memorized visuomotor routine triggered by the visual scene.

\textbf{Semantic intervention design.} At test time, MetaFine applies controlled semantic interventions to the natural-language instruction while holding the visual scene fixed. Interventions modify one attribute-level constraint at a time:
\begin{itemize}
    \item \textbf{Part substitution.} A part target is replaced with a different annotated part of the same object (e.g., ``grasp the cap of the bottle'' $\rightarrow$ ``grasp the body of the bottle'').
    \item \textbf{Directional reversal.} A directional cue is reversed (e.g., ``move left'' $\rightarrow$ ``move right'').
    \item \textbf{Property alteration.} An object attribute referenced in the instruction is altered (e.g., color, size).
\end{itemize}
The visual scene, initial robot state, and all non-instruction inputs are held identical to the unperturbed trial, isolating the intervention to language alone. A genuinely grounded policy should redirect its behavior to satisfy the modified specification; unchanged behavior under modified instructions indicates that the policy treats language as a coarse task selector rather than a compositional constraint.

\textbf{Metrics.} Two complementary metrics quantify the response:
\begin{equation}
    \Delta_{\text{drop}} \;=\; \text{SR}_{\text{orig}} - \text{SR}_{\text{pert}},
\end{equation}
where $\text{SR}_{\text{orig}}$ is the success rate under the training instruction and $\text{SR}_{\text{pert}}$ the success rate when the policy is queried with the perturbed instruction but evaluated against the \emph{original} acceptance criteria. A large $\Delta_{\text{drop}}$ indicates that the policy's behavior is disrupted by the surface-level change, but does not by itself reveal whether the policy has redirected to the new target.
The second metric, $\text{SR}_{\text{mod}}$, evaluates the same rollouts against the acceptance criteria of the \emph{modified} instruction. A grounded policy is expected to exhibit non-trivial $\text{SR}_{\text{mod}}$ proportional to its baseline competence; a policy that merely replays the original routine will show $\text{SR}_{\text{mod}} \approx 0$.

\textbf{Compound instructions.} For instructions that specify sequential sub-tasks (e.g., ``put the green cube in the left box, and put the other cubes in the right box''), each sub-task is evaluated independently using stage-wise acceptance criteria (Section~\ref{sec:appendix_behavior}). This reveals whether a policy can parse and transition between sub-instructions or arrests after executing only the first, a distinction invisible to overall task success.

\subsubsection{Perception evaluation}
\label{sec:appendix_perception_protocol}

Fine-grained manipulation demands perceptual fidelity at the level of parts, poses, and spatial relationships---a resolution substantially finer than that required for coarse object detection. The perception axis stress-tests this fidelity through two categories of perturbation that probe distinct aspects of perceptual robustness. Perturbation definitions and severity parameters are specified in Section~\ref{sec:appendix_perturbations}; here we define how robustness is quantified.

\textbf{Area under the success curve.} Robustness is summarized by the area under the success curve (AUSC), computed over success rates at levels $L_0$ (nominal) through $L_3$ (most severe):
\begin{equation}
    \text{AUSC} = \frac{1}{L_{\max}} \int_0^{L_{\max}} \text{SR}(\ell)\, d\ell \approx \frac{1}{2(K-1)} \sum_{k=1}^{K-1} \left[\text{SR}(\ell_k) + \text{SR}(\ell_{k+1})\right] \cdot (\ell_{k+1} - \ell_k),
    \label{eq:ausc}
\end{equation}
where $\text{SR}(\ell)$ denotes the success rate at perturbation level $\ell$ and $K$ is the number of evaluated levels including $L_0$. AUSC captures both the magnitude and the profile of degradation: a policy that degrades gradually under moderate perturbation but collapses at $L_3$ scores differently from one that drops sharply at $L_1$ and then stabilizes.

\textbf{Separated reporting.} Geometric and photometric AUSC are reported separately throughout the evaluation. The two perturbation types probe distinct aspects of perceptual competency and are governed by different architectural and training factors; a single aggregate robustness score would obscure these dissociations.

\subsubsection{Behavior evaluation}
\label{sec:appendix_behavior}

Even when understanding and perception are adequate, fine-grained manipulation can fail at execution if generated actions lack the precision or stability that tight physical constraints demand. The behavior axis characterizes controlled execution through two complementary lenses.

\textbf{Stage-wise success.} Multi-stage tasks are segmented into their constituent atomic skills, and acceptance criteria are evaluated independently at each stage. For the peg-in-hole task, \textsc{Grasp Part}, \textsc{Align}, and \textsc{Insert} are assessed separately, revealing at which point in the manipulation pipeline a policy fails rather than reporting only the end-to-end outcome. Stage-wise evaluation exposes competency differences that binary success rate actively conceals: two policies may both score near zero on the overall task while differing substantially in how far they progress through the skill sequence.

The per-stage criteria are inherited directly from the atomic skill definitions in Section~\ref{sec:appendix_skills} (e.g., positional and angular tolerances for \textsc{Align}; insertion depth for \textsc{Insert}). For compound long-horizon tasks such as compositional sorting, stages correspond to sub-instructions parsed from the compound command, and each sub-stage is evaluated as an independent atomic-skill instance.

\textbf{Trajectory stability.}
Trajectories are recorded at each policy's native control frequency and characterized through a smoothness metric:
\begin{equation}
    \text{Stability} = \exp\!\left( -\frac{1}{T-1} \sum_{t=1}^{T} \| a_t - a_{t-1} \|_2 \right),
    \label{eq:smoothness}
\end{equation}
where $a_t$ is the action at time step $t$ and $T$ the trajectory length. Stability $\in (0, 1]$, with values near $1$ indicating smooth, consistent action sequences. The metric dissociates execution quality from task outcome: a policy may achieve high stability with low success rate, indicating well-controlled but misdirected motion, or low stability with moderate success, indicating that brute-force corrections compensate for imprecise planning. Together, stage-wise decomposition and trajectory stability provide a behavioral profile that complements the upstream diagnostics of understanding and perception in a structured way.
\subsection{Perturbation Definitions}
\label{sec:appendix_perturbations}

Fine-grained manipulation places stringent demands on perceptual fidelity: localizing a specific part, resolving the relative pose between a grasped object and a target receptacle, and tracking a constrained trajectory all require spatial and appearance information at a resolution far beyond that needed for coarse object-level interaction. Real-world deployment, however, rarely affords static perceptual conditions: camera viewpoints differ across installations or shift due to mounting imprecision, and lighting varies with time of day, room configuration, and ambient light sources. A policy whose performance degrades substantially under such variations possesses brittle perception that will limit its reliability in practice.

MetaFine's perception protocol therefore measures how perceptual perturbations affect end-to-end manipulation performance, capturing the functional impact of perceptual degradation on the downstream task rather than testing perception in isolation. Perturbations fall into two categories defined by the nature of the visual change they introduce.

\textbf{Geometric perturbations.} Geometric perturbations alter the spatial structure of the observation. The primary instantiation is \emph{camera viewpoint shift}: the camera is displaced from its nominal pose by controlled amounts in position and orientation, changing the perspective from which the scene is observed. Geometric perturbations test whether the policy's spatial representations---object localization, part-level pose estimation, relative spatial reasoning---generalize across viewing angles. As demonstrated in the main text, geometric robustness is closely tied to the visual encoder's capacity to maintain spatial fidelity under viewpoint variation.

\textbf{Photometric perturbations.} Photometric perturbations modify low-level visual appearance without altering spatial structure. The primary instantiation is \emph{lighting variation}: the intensity, direction, and color temperature of scene illumination are modified, changing pixel-level appearance while preserving the 3D geometry of the scene. Photometric perturbations test whether the policy's visual representations are invariant to appearance changes that do not carry task-relevant spatial information. Photometric robustness is governed by different factors than geometric robustness, with the nature of the perceptual representation and pretraining data diversity playing a more prominent role than encoder architecture alone.

\textbf{Severity levels.} Each perturbation type is applied at three severity levels ($L_1$, $L_2$, $L_3$) representing progressively larger deviations from the nominal condition $L_0$. $L_1$ corresponds to mild perturbations representative of minor environmental variation; $L_2$ approaches the boundary of conditions encountered in typical deployment settings; $L_3$ stress-tests the limits of perceptual robustness. The specific parameters used in our experiments are summarized in Table~\ref{tab:perturbation_params}.

\begin{table}[ht]
\centering
\caption{Perturbation parameters at each severity level. Geometric parameters specify the maximum camera displacement from its nominal pose, including joint position and orientation offsets. Photometric parameters specify the lighting intensity scale, principal light direction offset, and color temperature shift relative to the nominal condition.}
\label{tab:perturbation_params}
\vspace{0.4em}
\small
\setlength{\tabcolsep}{6pt}
\begin{tabular}{l ccc}
\toprule
\textbf{Parameter} & \textbf{$L_1$ (mild)} & \textbf{$L_2$ (moderate)} & \textbf{$L_3$ (severe)} \\
\midrule
\multicolumn{4}{l}{\emph{Geometric: viewpoint shift}} \\
\quad Camera position offset (cm)              & 3 & 6 & 12 \\
\quad Camera orientation offset ($^\circ$)     & 3 & 6 & 12 \\
\midrule
\multicolumn{4}{l}{\emph{Photometric: lighting variation}} \\
\quad Ambient intensity offset ($\pm$, nominal $=1$) & 0.10 & 0.25 & 0.40 \\
\bottomrule
\end{tabular}
\end{table}

\textbf{Injection protocol.} All perturbations are applied after scene initialization and before policy execution begins. For geometric perturbations, the camera is repositioned to the perturbed pose prior to the first observation. For photometric perturbations, the lighting configuration is modified before the first frame is rendered. The perturbation remains fixed throughout the trial, ensuring that any performance change is attributable to the perceptual shift rather than to dynamic variation during execution.

\subsection{Evaluated Policies}
\label{sec:appendix_policies}

We evaluate seven representative policies spanning four architectural families. Together, these policies cover the principal design axes of current manipulation architectures, including visual encoder design (single vs.\ dual encoder, 2D vs.\ 3D), action generation paradigm (deterministic regression, diffusion, flow matching, autoregressive token prediction), and depth of language conditioning. Table~\ref{tab:policy_summary} summarizes their key architectural properties.

\begin{table}[ht]
\centering
\caption{Summary of the evaluated policies organized along three principal design axes: visual encoder, action generation paradigm, and language conditioning.}
\label{tab:policy_summary}
\vspace{0.4em}
\small
\setlength{\tabcolsep}{3pt}
\begin{tabular}{l l l}
\toprule
\textbf{Policy} & \textbf{Visual Encoder} & \textbf{Action Generation}  \\
\midrule
ACT          & ResNet (2D)              & Deterministic regression          \\
DP3          & Point cloud encoder (3D) & Diffusion                         \\
Octo         & Lightweight encoder (2D) & Diffusion                        \\
OpenVLA      & DINOv2+SigLIP (2D, dual) & Autoregressive (discrete)         \\
OpenVLA-OFT  & DINOv2+SigLIP (2D, dual) & $\ell_1$ regression (continuous) \\
$\pi_0$      & SigLIP (2D, single)      & Flow matching                     \\
$\pi_{0.5}$  & SigLIP (2D, single)      & Flow matching                    \\
\bottomrule
\end{tabular}
\end{table}

\begin{itemize}

\item \textbf{ACT}~\cite{zhao2023act}. Action Chunking with Transformers predicts chunks of future actions through a conditional variational autoencoder (CVAE) with a transformer backbone. A CVAE encoder compresses sequences of joint positions into a style variable, and a decoder generates action chunks conditioned on this latent and current observations. ACT processes raw images through a ResNet backbone and outputs deterministic continuous actions via $\ell_1$ regression. It uses no pretrained visual encoder and no language conditioning, representing the baseline of lightweight, non-VLA imitation learning.

\item \textbf{DP3}~\cite{ze2024dp3}. 3D Diffusion Policy extends diffusion-based action generation to 3D point cloud observations. A compact point cloud encoder produces a scene representation that conditions an iterative denoising diffusion process over action chunks. DP3 uses no language conditioning. Its 3D perception pipeline directly captures geometric structure and is inherently decoupled from 2D appearance properties such as lighting and texture, making it a natural reference point for separating geometric from photometric robustness.

\item \textbf{Octo}~\cite{team2024octo}. Octo is a 93M-parameter generalist robot policy built on a transformer backbone with a diffusion action head. A lightweight visual encoder processes image observations and a T5 language encoder processes instructions; the transformer integrates visual, language, and proprioceptive tokens, and a diffusion head generates action chunks through iterative denoising. Octo is pretrained on approximately 800K robot trajectories from the Open X-Embodiment dataset and supports flexible fine-tuning to new observation and action spaces. It represents the lightweight, diffusion-based generalist paradigm with language conditioning.

\item \textbf{OpenVLA}~\cite{kim2024openvla}. OpenVLA is a 7B-parameter vision-language-action model built on the Prismatic VLM backbone. Its visual encoder fuses features from DINOv2 and SigLIP through a dual-encoder design, combining self-supervised geometric features with language-grounded semantics. Fused visual tokens are projected into the embedding space of a Llama~2 7B language model, which processes visual and language tokens jointly and predicts discretized robot actions via autoregressive next-token prediction. Actions are discretized into 256 bins per dimension and decoded into continuous values for execution. OpenVLA is pretrained on approximately 970K robot manipulation trajectories from Open X-Embodiment.

\item \textbf{OpenVLA-OFT}~\cite{kim2025openVLAoft}. OpenVLA-OFT applies an Optimized Fine-Tuning recipe to the OpenVLA architecture. Autoregressive decoding is replaced with parallel decoding under bidirectional attention, the language model's output layer is substituted with a 4-layer MLP action head that generates continuous actions via $\ell_1$ regression, and the model outputs chunks of $K$ actions per forward pass. The visual encoder (DINOv2+SigLIP) and language backbone (Llama~2 7B) are inherited from OpenVLA but fine-tuned with LoRA. The MLP head provides a weaker gradient pathway between linguistic input and motor output compared to architectures that jointly optimize language and action losses.

\item $\boldsymbol{\pi_0}$~\cite{black2024pi0}. $\pi_0$ is a vision-language-action flow model built on a pretrained PaliGemma VLM (approximately 3B parameters) with a SigLIP visual encoder and a Gemma language backbone. It introduces a separate flow-matching action expert (approximately 300M parameters) that generates continuous actions by sampling from a learned conditional distribution. The VLM processes visual and language inputs into contextualized representations that condition the action expert’s flow-matching generation. $\pi_0$ is pretrained on a large cross-embodiment robot dataset.

\item $\boldsymbol{\pi_{0.5}}$~\cite{black2025pi05}. $\pi_{0.5}$ extends $\pi_0$ by jointly optimizing a next-token prediction objective over language tokens and a flow-matching loss over action tokens within the same VLM backbone. This joint training maintains gradient flow through the language representations during action learning, providing tighter coupling between linguistic input and motor output compared to architectures that route actions through a separate head. $\pi_{0.5}$ uses a single SigLIP encoder without DINOv2, retaining semantic capacity but lacking the spatial fidelity provided by a dual-encoder design. It represents the current state of the art in general-purpose VLA models.

\end{itemize}

\textbf{Fine-tuning protocol.} All policies are fine-tuned on MetaFine's demonstration data under matched conditions. Each policy is fine-tuned on the demonstrations specified in Table~\ref{tab:task_specs}, starting from the original authors' released checkpoints where available, and using the official fine-tuning hyperparameters (learning rate, batch size, optimizer, training schedule) released by each policy's authors without modification. Fixing hyperparameters to the official recipe isolates the comparison to architectural and training-paradigm differences across policies rather than tuning effort. Evaluation rollout counts and random seeds are recorded in the released codebase to support reproducibility.

\section{Extended Evaluation Results}
\label{supp:sec3}

This section provides the complete numerical results underlying the figures and analyses in the main text. All values are averaged over multiple trials per task configuration; evaluation scripts and random seeds are included in the released codebase.

\subsection{Full Numerical Results across All Tasks and Policies}
\label{sec:appendix_full_results}

We evaluate seven representative policies spanning distinct architectural families: ACT, DP3, Octo, OpenVLA, OpenVLA-OFT, $\pi_0$, and $\pi_{0.5}$. All policies are trained on the demonstrations specified in Table~\ref{tab:task_specs} and evaluated under identical conditions across all three competency dimensions.

\textbf{Success rates and trajectory quality.} Table~\ref{tab:nominal_sr} reports the success rate and trajectory stability score for each policy on all evaluated tasks under standard (unperturbed) conditions. These results correspond to Fig.~\ref{fig2}A in the main text.

\begin{table}[ht]
\centering
\caption{Success rates (\%) and trajectory stability scores under standard conditions, reported as ``SR / Stab.''. Stability is computed via Eq.~\eqref{eq:smoothness}; values near 1.0 indicate smooth execution. ``--'' indicates that no successful trajectory was available for stability computation. Best SR per task is \textbf{bolded}. All policies score 0\% on Plug Charger and Stack Pyramid under binary evaluation; on Peg in Hole only OpenVLA-OFT records a non-zero overall SR (3\% / 0.71), with stage-wise decomposition provided in Table~\ref{tab:stagewise_peg}.}
\label{tab:nominal_sr}
\vspace{0.8em}
\tiny
\setlength{\tabcolsep}{4.2pt}
\begin{tabular}{l cccccc}
\toprule
\textbf{Policy} & \textbf{Grasp Part} & \textbf{Toggle Part} & \textbf{Press Part} & \textbf{Slide Along} & \textbf{Rotate Along} & \textbf{Open Hinge} \\
\midrule
ACT          & 49 / 0.95          & 63 / 0.98          & 63 / 0.98          & 8 / 0.93           &  0 / --   &  0 / --            \\
DP3          & 75 / 0.95          & \textbf{85} / 0.95 & 63 / 0.95          & 5 / 0.91           &  0 / --   &  3 / 0.93          \\
Octo         & 35 / 0.93          & 49 / 0.92          & 44 / 0.94          & --                 &  0 / --   & --                 \\
OpenVLA      & 35 / 0.88          & 45 / 0.89          & 53 / 0.86          & --                 &  5 / 0.77 & --                 \\
OpenVLA-OFT  & 37 / 0.87          & 50 / 0.90          & 59 / 0.90          & 18 / 0.91          & 12 / 0.83 & 10 / 0.89          \\
$\pi_0$      & 75 / 0.92          & 81 / 0.87          & 65 / 0.93          & --                 & 13 / 0.87 & --                 \\
$\pi_{0.5}$  & \textbf{80} / 0.91 & 79 / 0.98          & \textbf{68} / 0.93 & \textbf{22} / 0.95 & 10 / 0.90 & \textbf{13} / 0.93 \\
\bottomrule
\end{tabular}
\end{table}

\textbf{Perceptual robustness.} Table~\ref{tab:perception_full} reports success rates under lighting and viewpoint perturbations at three severity levels ($L_1$--$L_3$), together with the corresponding AUSC (Eq.~\eqref{eq:ausc}) for each perturbation type. These results correspond to Fig.~\ref{fig2}A and Fig.~\ref{fig2}B in the main text. For atomic tasks where all policies achieve 0\% under all perturbation levels (Peg in Hole, Plug Charger, Stack Pyramid), entries are omitted; their complete results are zero across all conditions.

\begin{table}[ht]
\centering
\caption{Perceptual robustness: success rates (\%) under lighting and viewpoint perturbation at levels $L_1$--$L_3$, with corresponding AUSC (\%) computed via Eq.~\eqref{eq:ausc}. Best AUSC per task is \textbf{bolded}. Tasks where all policies score 0\% under all perturbation conditions are omitted.}
\label{tab:perception_full}
\vspace{0.5em}
\scriptsize
\setlength{\tabcolsep}{2.5pt}
\begin{tabular}{ll cccc cccc}
\toprule
& & \multicolumn{4}{c}{\textbf{Lighting Perturbation}} & \multicolumn{4}{c}{\textbf{Viewpoint Perturbation}} \\
\cmidrule(lr){3-6} \cmidrule(lr){7-10}
\textbf{Task} & \textbf{Policy} & $L_1$ & $L_2$ & $L_3$ & AUSC & $L_1$ & $L_2$ & $L_3$ & AUSC \\
\midrule
\multirow{7}{*}{Grasp Part}
 & ACT          & 41 & 34 & 12 & 35.17          & 47 & 39 & 30 & 41.83 \\
 & DP3          & 75 & 76 & 75 & \textbf{75.25} & 49 & 21 & 10 & 38.75 \\
 & Octo         & 25 & 20 &  9 & 22.33          & 19 & 10 &  0 & 15.50 \\
 & OpenVLA      & 35 & 20 & 11 & 26.00          & 16 &  9 &  0 & 14.17 \\
 & OpenVLA-OFT  & 31 & 21 &  7 & 25.50          & 14 &  7 &  0 & 14.50 \\
 & $\pi_0$      & 71 & 40 & 12 & 51.50          & 68 & 57 & 54 & \textbf{63.17} \\
 & $\pi_{0.5}$  & 74 & 45 & 15 & 53.50          & 63 & 55 & 55 & 61.83 \\
\midrule
\multirow{7}{*}{Toggle Part}
 & ACT          & 41 & 27 &  7 & 34.33          & 60 & 49 & 11 & 48.67 \\
 & DP3          & 85 & 85 & 83 & \textbf{84.50} & 63 & 41 & 25 & 53.50 \\
 & Octo         & 35 & 12 &  2 & 24.17          & 35 & 16 &  9 & 26.67 \\
 & OpenVLA      & 39 &  3 &  0 & 21.50          & 24 & 19 &  7 & 23.00 \\
 & OpenVLA-OFT  & 43 &  0 &  0 & 23.25          & 37 & 25 & 19 & 32.75 \\
 & $\pi_0$      & 65 & 47 & 13 & 53.00          & 69 & 51 & 33 & \textbf{59.00} \\
 & $\pi_{0.5}$  & 63 & 35 & 11 & 47.67          & 71 & 49 & 38 & 59.50 \\
\midrule
\multirow{7}{*}{Press Part}
 & ACT          & 47 & 21 & 10 & 34.83          & 43 & 39 & 23 & 41.67 \\
 & DP3          & 62 & 61 & 62 & \textbf{62.00} & 41 & 30 & 22 & 39.00 \\
 & Octo         & 26 & 18 &  7 & 23.17          & 38 & 25 & 14 & 30.67 \\
 & OpenVLA      & 33 & 20 & 14 & 28.83          & 41 & 28 & 19 & 35.00 \\
 & OpenVLA-OFT  & 43 & 27 & 13 & 35.50          & 43 & 34 & 26 & 40.50 \\
 & $\pi_0$      & 57 & 44 & 20 & 47.83          & 47 & 41 & 33 & 45.67 \\
 & $\pi_{0.5}$  & 49 & 37 & 15 & 42.50          & 47 & 49 & 31 & \textbf{48.50} \\
\midrule
\multirow{4}{*}{Slide Along}
 & ACT          &  5 &  3 &  1 &  4.17          &  6 &  3 &  1 &  4.50          \\
 & DP3          &  4 &  3 &  0 &  3.17          &  3 &  2 &  1 &  2.67          \\
 & OpenVLA-OFT  & 12 &  5 &  1 &  8.83          & 14 &  7 &  1 & 10.17          \\
 & $\pi_{0.5}$  & 15 &  7 &  2 & \textbf{11.33} & 17 & 10 &  3 & \textbf{13.17} \\
\midrule
\multirow{7}{*}{Rotate Along}
 & ACT          &  0 &  0 &  0 &  0.00         &  0 &  0 &  0 &  0.00         \\
 & DP3          &  0 &  0 &  0 &  0.00         &  0 &  0 &  0 &  0.00         \\
 & Octo         &  0 &  0 &  0 &  0.00         &  0 &  0 &  0 &  0.00         \\
 & OpenVLA      &  3 &  0 &  0 &  1.83         &  0 &  0 &  0 &  0.83         \\
 & OpenVLA-OFT  &  9 &  0 &  0 & \textbf{5.00} &  8 &  3 &  0 &  5.67         \\
 & $\pi_0$      & 11 &  2 &  0 &  6.50         & 10 &  5 &  2 & \textbf{7.50} \\
 & $\pi_{0.5}$  &  5 &  0 &  0 &  3.33         &  8 &  1 &  0 &  4.67         \\
\midrule
\multirow{4}{*}{Open Hinge}
 & ACT          &  0 &  0 &  0 &  0.00         &  0 &  0 &  0 &  0.00         \\
 & DP3          &  1 &  1 &  1 &  1.33         &  1 &  1 &  1 &  1.33         \\
 & OpenVLA-OFT  &  5 &  2 &  0 &  4.00         &  7 &  2 &  1 &  4.83         \\
 & $\pi_{0.5}$  &  6 &  2 &  1 & \textbf{5.00} &  8 &  3 &  0 & \textbf{5.83} \\
\bottomrule
\end{tabular}
\end{table}

\textbf{Understanding: instruction perturbation.} Table~\ref{tab:understanding} reports the impact of controlled semantic interventions on the five VLA policies. For the Grasp Part task, the instruction is modified from ``grasp the cap of the bottle'' to ``grasp the body of the bottle'' while the visual scene remains unchanged. These results correspond to Fig.~\ref{fig2}C\,ii in the main text.

\begin{table}[ht]
\centering
\caption{Instruction perturbation results for VLA policies on Grasp Part. ``Original SR'' is the success rate under the training instruction. ``Perturbed SR'' is the success rate when the attribute-level constraint is modified but the original acceptance criteria are applied. ``SR Drop'' quantifies the disruption to the learned visuomotor routine. ``Modified Task SR’’ measures whether the policy successfully executes the \emph{modified} instruction under the perturbed condition.}
\label{tab:understanding}
\small
\begin{tabular}{l cccc}
\toprule
\textbf{Policy} & \textbf{Original SR} & \textbf{Perturbed SR} & \textbf{SR Drop} & \textbf{Modified Task SR} \\
\midrule
Octo         & 35\% & 27\% &  8.0\% & 0\% \\
OpenVLA      & 35\% & 29\% &  6.0\% & 0\% \\
OpenVLA-OFT  & 37\% & 27\% & 10.0\% & 0\% \\
$\pi_0$      & 75\% & 41\% & 34.0\% & 0\% \\
$\pi_{0.5}$  & 80\% & 55\% & 31.2\% & 0\% \\
\bottomrule
\end{tabular}
\end{table}

\textbf{Stage-wise decomposition of long-horizon tasks.} Tables~\ref{tab:stagewise_peg} and~\ref{tab:stagewise_sort} provide stage-wise success rates for two representative long-horizon tasks. Binary evaluation reports near-zero overall success for all policies on both tasks, which would suggest equivalent failure. Stage-wise decomposition reveals qualitatively different failure patterns, with policies diverging at distinct points along the manipulation pipeline.

\begin{table}[ht]
\centering
\caption{Stage-wise success rates (\%) for the peg-in-hole task (\textsc{Grasp Part} $\to$ \textsc{Align} $\to$ \textsc{Insert}). The critical divergence occurs at the \textsc{Align} stage: OpenVLA-OFT reaches 19\%, roughly double the next best policy, while $\pi_{0.5}$ achieves 0\% despite a competitive grasp rate. These distinctions are entirely invisible to binary evaluation.}
\label{tab:stagewise_peg}
\small
\begin{tabular}{l cccc}
\toprule
\textbf{Policy} & \textbf{Overall SR} & \textbf{Grasp} & \textbf{Align} & \textbf{Insert} \\
\midrule
Octo         &  0\% & 15\% &  4\%          &  0\% \\
OpenVLA      &  0\% & 21\% & 10\%          &  0\% \\
OpenVLA-OFT  &  3\% & 47\% & \textbf{19\%} &  3\% \\
$\pi_0$      &  0\% & 33\% &  7\%          &  0\% \\
$\pi_{0.5}$  &  0\% & 39\% &  0\%          &  0\% \\
\bottomrule
\end{tabular}
\end{table}

\begin{table}[ht]
\centering
\caption{Stage-wise success rates (\%) for the compound sorting task. Each variant requires grasping a specified cube, placing it in the left box, then grasping the remaining cubes and placing them in the right box. Stages 3--4 test whether the policy can parse and transition to the second sub-instruction. The flow-matching family ($\pi_0$ and $\pi_{0.5}$) maintains non-zero success across all four stages; OpenVLA-OFT, OpenVLA, and Octo collapse entirely at stages 3--4 (with the sole exception of Octo's 3\% on stage three of the green variant), indicating behavioral arrest after the first sub-task.}
\label{tab:stagewise_sort}
\vspace{0.5em}
\scriptsize
\setlength{\tabcolsep}{3pt}
\renewcommand{\arraystretch}{1.2}
\begin{tabular}{ll cccc}
\toprule
\textbf{Variant} & \textbf{Policy} & \textbf{Stage 1} & \textbf{Stage 2} & \textbf{Stage 3} & \textbf{Stage 4} \\
& & \tiny{Grasp target} & \tiny{Place left} & \tiny{Grasp others} & \tiny{Place right} \\
\midrule
\multirow{5}{*}{\parbox{3.2cm}{\raggedright Put the \textbf{green} cube in left box, others in right box}}
 & Octo         & 16 &  7 &  3 &  0 \\
 & OpenVLA      & 14 &  9 &  0 &  0 \\
 & OpenVLA-OFT  & 26 & 14 &  0 &  0 \\
 & $\pi_0$      & 29 & 17 & 25 & 21 \\
 & $\pi_{0.5}$  & 22 &  4 & 24 & 12 \\
\midrule
\multirow{5}{*}{\parbox{3.2cm}{\raggedright Put the \textbf{red} cube in left box, others in right box}}
 & Octo         & 19 & 11 &  0 &  0 \\
 & OpenVLA      & 27 & 13 &  0 &  0 \\
 & OpenVLA-OFT  & 32 & 17 &  0 &  0 \\
 & $\pi_0$      & 33 & 20 & 28 & 16 \\
 & $\pi_{0.5}$  & 25 &  9 & 28 & 13 \\
\midrule
\multirow{5}{*}{\parbox{3.2cm}{\raggedright Put the \textbf{blue} cube in left box, others in right box}}
 & Octo         & 23 & 17 &  0 &  0 \\
 & OpenVLA      & 26 & 18 &  0 &  0 \\
 & OpenVLA-OFT  & 30 & 19 &  0 &  0 \\
 & $\pi_0$      & 30 & 21 & 23 & 11 \\
 & $\pi_{0.5}$  & 33 & 15 & 25 & 11 \\
\bottomrule
\end{tabular}
\end{table}

\textbf{Viewpoint adaptation ablation.} Table~\ref{tab:adaptation} compares two one-shot adaptation strategies for recovering viewpoint robustness: LoRA fine-tuning applied to the VLM backbone, and a lightweight learnable mapping applied solely to the visual encoder output. Both use a single demonstration from the perturbed viewpoint. All configurations achieve identical performance under standard conditions, confirming that the adaptations do not degrade baseline capability. Under progressive viewpoint perturbation, encoder-only adaptation matches or exceeds backbone-level adaptation across all tasks, providing direct evidence that geometric robustness degradation originates in the encoder's spatial representations. These results correspond to Fig.~\ref{fig5}B in the main text.

\begin{table}[ht]
\centering
\caption{Viewpoint adaptation ablation: success rates (\%) under viewpoint perturbation. Encoder-only lightweight adaptation matches or exceeds VLM backbone LoRA across all conditions, confirming the encoder as the geometric robustness bottleneck. Best perturbed result per condition is \textbf{bolded}.}
\label{tab:adaptation}
\vspace{0.5em}
\small
\begin{tabular}{ll cccc}
\toprule
\textbf{Task} & \textbf{Configuration} & \textbf{Standard} & $L_1$ & $L_2$ & $L_3$ \\
\midrule
\multirow{3}{*}{Grasp Part}
 & $\pi_{0.5}$ (zero-shot)              & 80 & 63          & 55          & 55          \\
 & $\pi_{0.5}$ + LoRA SFT (one-shot)    & 80 & 73          & 66          & 59          \\
 & $\pi_{0.5}$ + lightweight (one-shot) & 80 & \textbf{72} & \textbf{68} & \textbf{64} \\
\midrule
\multirow{3}{*}{Toggle Part}
 & $\pi_{0.5}$ (zero-shot)              & 79 & 71          & 49          & 38          \\
 & $\pi_{0.5}$ + LoRA SFT (one-shot)    & 79 & 73          & 58          & 47          \\
 & $\pi_{0.5}$ + lightweight (one-shot) & 79 & \textbf{77} & \textbf{61} & \textbf{55} \\
\midrule
\multirow{3}{*}{Press Part}
 & $\pi_{0.5}$ (zero-shot)              & 68 & 47          & 49          & 31          \\
 & $\pi_{0.5}$ + LoRA SFT (one-shot)    & 68 & 56          & 58          & 40          \\
 & $\pi_{0.5}$ + lightweight (one-shot) & 68 & \textbf{60} & \textbf{61} & \textbf{51} \\
\bottomrule
\end{tabular}
\end{table}

\subsection{Coarse-Grained vs.\ Fine-Grained Task Comparison}
\label{sec:appendix_coarse_vs_fine}

A central claim of MetaFine is that existing benchmarks overestimate manipulation competency by evaluating at the object level without enforcing fine-grained constraints. To quantify this directly, Table~\ref{tab:coarse_vs_fine} compares success rates under coarse-grained evaluation (any successful interaction, regardless of which part is contacted or which direction is followed) with fine-grained evaluation (the attribute-level constraint in the instruction must be satisfied). Both evaluations use identical scenes and policy rollouts; only the acceptance criterion differs.

\noindent The fine-grained success rates are reproduced from Table~\ref{tab:nominal_sr}; the coarse-grained rates represent evaluation under relaxed acceptance criteria applied to the same policy rollouts. The consistent gap across all policies and tasks confirms that object-level evaluation systematically inflates reported manipulation performance. The magnitude of the gap scales with constraint stringency: \textsc{Rotate Along}, which demands directional compliance, exhibits the largest discrepancy, while tasks with comparatively looser constraints show smaller gaps.

\section{Supplementary Methods: Task Graph Mechanics and Benchmark Adapters}
\label{appendix_method_sec}

Section~\ref{supp:sec1} introduces the components of MetaFine and the protocol it applies. This section provides supplementary methodological detail on two aspects that warrant deeper treatment: the operational semantics of the compositional task graph (how skills compose, how tasks are generated, how an evaluation trial is dispatched), and the design of the benchmark adapter mechanism that allows MetaFine to absorb tasks from external suites such as RoboTwin, ManiSkill, and LIBERO. Together, these mechanisms underpin MetaFine's positioning as a meta-evaluation framework rather than a fixed task collection.

\subsection{Task Graph Operational Semantics}
\label{sec:appendix_task_graph_semantics}

The compositional task graph is more than a notational convenience: it is the operational substrate that turns the atomic skill vocabulary (Section~\ref{sec:appendix_skills}) into executable evaluation trials. This subsection makes the underlying mechanics explicit.

\textbf{Skill composability through specification compatibility.} As stated in Section~\ref{sec:appendix_framework}, each atomic skill is specified by a tuple $(p, q, \mathcal{C})$. The compositional logic of the task graph is fully determined by these specifications: a directed edge from skill $s_i$ to skill $s_j$ exists if and only if $q_i \Rightarrow p_j$, that is, the postcondition of $s_i$ implies the precondition of $s_j$. Implication is checked symbolically over the predicate vocabulary used to express conditions---predicates such as ``grasp-stable($o$)'', ``aligned($o$, $r$, $\epsilon_{\text{pos}}$, $\epsilon_{\text{ang}}$)'', or ``in-hand($o$)''. When $q_i$ entails the conjunction of predicates required by $p_j$, the transition is admissible.

This design choice has two consequences. First, the set of admissible skill compositions is derived from the specifications themselves rather than from a manually curated transition table, which would have to be re-enumerated whenever a new skill is introduced. Second, the same specifications drive the per-stage acceptance criteria that the protocol uses during evaluation: when an edge $s_i \rightarrow s_j$ is traversed, the criteria that verify completion of $s_i$ are exactly the predicates appearing in $q_i$. The specification therefore plays a dual role---both as a compositional constraint and as an evaluation contract.

\begin{table}[t]
\centering
\caption{Coarse-grained vs.\ fine-grained success rates (\%). Coarse evaluation applies the criteria used by conventional benchmarks, where any stable grasp or successful actuation is accepted regardless of contact region. Fine-grained evaluation enforces the part-level or directional constraint specified in the MetaFine task, requiring contact at the instructed part or motion along the instructed axis. The gap $\Delta$ quantifies the overestimation introduced by conventional evaluation: many rollouts succeed at the object level but fail to satisfy the tighter physical constraints that fine-grained tasks impose. Tasks without a natural coarse--fine distinction (e.g., Toggle Part) are excluded.}
\label{tab:coarse_vs_fine}
\vspace{0.5em}
\small
\renewcommand{\arraystretch}{1.3}
\setlength{\tabcolsep}{3pt}
\begin{tabular}{l ccc ccc ccc ccc}
\toprule
& \multicolumn{3}{c}{\textbf{ACT}} & \multicolumn{3}{c}{\textbf{DP3}} & \multicolumn{3}{c}{$\pi_{0.5}$} & \multicolumn{3}{c}{\textbf{OpenVLA-OFT}} \\
\cmidrule(lr){2-4} \cmidrule(lr){5-7} \cmidrule(lr){8-10} \cmidrule(lr){11-13}
\textbf{Task} & Coarse & Fine & $\Delta$ & Coarse & Fine & $\Delta$ & Coarse & Fine & $\Delta$ & Coarse & Fine & $\Delta$ \\
\midrule
Grasp Part    & 87 & 49 & 38 & 95 & 75 & 20 & 92 & 80 & 12 & 73 & 37 & 36 \\
Press Part    & 79 & 63 & 16 & 82 & 63 & 19 & 84 & 68 & 16 & 76 & 59 & 17 \\
Slide Along   & 32 &  8 & 24 & 27 &  5 & 22 & 49 & 22 & 27 & 43 & 18 & 25 \\
Rotate Along  & 14 &  0 & 14 & 11 &  0 & 11 & 40 & 10 & 30 & 37 & 12 & 25 \\
\bottomrule
\end{tabular}
\renewcommand{\arraystretch}{1.0}
\end{table}

\textbf{Edge types and execution semantics.}
The three edge types introduced in Section~\ref{sec:appendix_framework} (sequential, conditional, parallel) impose distinct execution semantics that the runtime respects during evaluation. A sequential edge is satisfied when $q_i$ has been verified at runtime and the policy then receives an observation reflecting the post-$s_i$ scene state; control flow is single-threaded. A conditional edge maps a runtime scene predicate (e.g., the color attribute of the held object) to a branch selector, with each branch corresponding to a distinct successor skill; the predicate is evaluated at the moment the parent skill terminates. A parallel edge is satisfied throughout the execution window of its host skill rather than at a single transition point: the runtime continuously monitors the associated constraint (e.g., orientation compliance) and a violation at any time step counts as failure regardless of the eventual end-state.

These execution semantics ensure that a multi-stage task is not merely a sequence of skills but a structured execution graph in which control, branching, and parallel monitoring are first-class operations. The runtime exposes per-skill, per-edge, and per-constraint logging, which is what enables the stage-wise and constraint-respecting metrics reported throughout Section~\ref{supp:sec3}.

\textbf{From abstract subgraph to bound task.}
Converting an abstract skill composition into an executable task involves selecting a subgraph $\mathcal{G}_{\text{task}}$ from the global task graph, binding each skill node to specific objects, parts, and constraint parameters drawn from the asset library, and filling the associated natural-language instruction template with the bound entities. The output is a fully self-contained task specification: instruction, skill sequence with edge types, per-stage acceptance criteria, and scene initialization configuration. The downstream conversion of this specification into a simulation scene and demonstration trajectories is handled by the task-generation stage described in Section~\ref{sec:appendix_task_gen}. Because each step operates over typed structured data, instantiation requires no modification to the evaluation infrastructure---only a new $\mathcal{G}_{\text{task}}$ or a new binding.

\textbf{Agent-assisted skill and task construction.}
The task graph is designed for human curation but admits agent-assisted authoring at two levels. For \emph{new skills}, an LLM agent is provided with a natural-language description of the desired manipulation behavior and drafts a candidate $(p, q, \mathcal{C})$ tuple along with suggested tolerance parameters and an acceptance test. For \emph{new tasks}, the agent receives a high-level task description (e.g., ``assemble the pen by inserting the refill and attaching the cap'') and proposes a candidate subgraph by selecting compatible skill nodes from the vocabulary, ordering them under valid edge types, and binding them to objects from the asset library. In both cases the agent's output is a draft that a human reviewer validates and refines; the protocol design ensures that the validation surface is the small set of predicates and tolerance parameters rather than the entire task implementation. This division of labor lowers the overhead of expanding the evaluation vocabulary without ceding ground on the formal rigor that the compositional design requires.

\subsection{Benchmark Adapter Design}
\label{sec:appendix_adapters}

Existing manipulation benchmarks define tasks under heterogeneous conventions: each suite has its own object set, its own robot embodiment, its own action interface, and its own success criteria. Comparing policies across benchmarks under these conventions is unreliable, since a performance difference may reflect the underlying competency of the policy or merely a difference in how success is defined. MetaFine's adapter mechanism resolves this by re-expressing tasks from external benchmarks within the task graph formalism, so that they can be evaluated under the three-dimensional protocol on common ground.

The system separates concerns between two structural layers (Fig.~\ref{fig4}). The \emph{stable libraries}---actions and attributes---encode MetaFine's normalized action interface and the attribute vocabulary used throughout the framework; these are held fixed across all benchmarks and never modified by an adapter. The \emph{adapters}, by contrast, are benchmark-specific translators: each external benchmark has its own adapter that ingests the benchmark's objects and robots, maps them onto MetaFine's representations, and routes them into the downstream task graph construction stage. New benchmarks are absorbed by writing a new adapter, leaving the stable libraries---and therefore the semantics of MetaFine's evaluation contract---untouched.

\textbf{Adapter pipeline.}
Each adapter follows a four-stage processing pipeline that is systematically applied to every task in its source benchmark.

\textbf{Stage 1: Asset and robot ingestion.}
The adapter ingests the object meshes, robot embodiment, and scene configuration of the source task and maps each entity onto MetaFine's representations. Objects are normalized into the asset library schema (Section~\ref{sec:appendix_assets}), reusing existing library entries where the source object matches and registering new entries otherwise. Robots are mapped onto MetaFine's standardized end-effector interface; embodiment-specific parameters (workspace bounds, gripper geometry, control frequency) are recorded in the task specification but do not enter the abstract skill sequence.

\textbf{Stage 2: Atomic skill identification.}
Given the normalized asset and robot, the adapter identifies which atomic skills from MetaFine's vocabulary the configuration can support. This identification is structural rather than semantic: a part with an annotated grasp pose and a contact region admits \textsc{Grasp Part}; an articulated joint with a hinge axis admits \textsc{Open Hinge}; a part with an actuation axis admits \textsc{Press Part} or \textsc{Toggle Part}; and so on. The atomic-skill specifications themselves (Section~\ref{sec:appendix_skills}) define which annotations are required for support. The output of this stage is the set of skills that can be validly exercised on the asset, conditioned on the available annotations.

\textbf{Stage 3: Fine-grained annotation backfilling.}
When an asset lacks the fine-grained annotations (part segmentation, grasp poses, manipulation constraints) that a required skill demands, the adapter invokes the automated annotation toolkit described in Section~\ref{sec:appendix_assets} to backfill the missing annotations. The toolkit's part segmentation, grasp pose proposal, and constraint inference run automatically; a human reviewer validates the output, with the toolkit completing the full pipeline in under 10 seconds per object on average. After backfilling, the asset carries the same annotation schema as natively curated entries and can support any compatible skill.

\textbf{Stage 4: Task graph construction.}
The adapter then constructs the task graph for the source task by selecting atomic skills from the identified set, ordering them under sequential, conditional, or parallel edges (Section~\ref{sec:appendix_task_graph_semantics}), and binding each skill node to the specific objects, parts, and constraint parameters required. The decomposition operates at the level of physical constraints rather than narrative description: a source task that requires opening a drawer is decomposed into \textsc{Grasp Part} (handle) followed by \textsc{Slide Along} (drawer rail), regardless of how the source benchmark describes the task. Source-task attribute-level conditions (part designators, directional cues, color conditions) are extracted as explicit constraints attached to the corresponding nodes.

The constructed task graph is then handed off to MetaFine's task generation stage (Section~\ref{sec:appendix_task_gen}), which instantiates the simulation scene and produces demonstration trajectories under the unified runtime. From the perspective of the evaluation runtime, an absorbed task is indistinguishable from a natively defined task: the same protocol, the same metrics, and the same three-dimensional decomposition apply. Instruction text is produced in parallel via the instruction-generation pipeline associated with the task graph, which fills the templated phrasing with the bound entities.

\textbf{Example adapters: RoboTwin, ManiSkill, LIBERO.}
We have implemented adapters for three external benchmarks that span complementary task profiles. \emph{RoboTwin}~\cite{robotwin} provides dual-arm scenarios with generative digital twins; its tasks frequently combine articulated object manipulation with bimanual coordination, which the adapter expresses as sequential compositions of \textsc{Open Hinge}, \textsc{Grasp Part}, and \textsc{Insert} skills, with parallel constraints enforcing inter-arm coordination. \emph{ManiSkill}~\cite{mu2023maniskill2} contributes a wide variety of single-arm tasks ranging from precision assembly to articulated object operation; many of its tasks correspond to direct instantiations of MetaFine's atomic skills, with the adapter primarily handling object and action normalization. \emph{LIBERO}~\cite{libero} targets lifelong learning over compositional task distributions; its long-horizon language-conditioned tasks decompose naturally into multi-stage skill sequences, with the adapter extracting attribute-level constraints from instruction templates and binding them to the appropriate skill nodes.

In each case, the adapter expresses source tasks within MetaFine's formalism without modifying the source benchmark's underlying simulation or assets. Tasks that contain operations outside MetaFine's current atomic vocabulary are flagged for vocabulary extension rather than forced into approximate mappings; this conservatism preserves the diagnostic integrity of the three-dimensional protocol, since reporting a result under MetaFine's metrics presupposes that the constituent skills carry MetaFine's acceptance contracts.

\subsection{Task Generation}
\label{sec:appendix_task_gen}

Once a task graph has been constructed---whether natively or through an adapter---it is converted into a concrete simulation task and an associated demonstration set by the task generation stage. The stage proceeds in three steps. (1)~The simulation scene is initialized: the asset library entries bound to the task graph are placed into the simulation environment according to the scene configuration recorded by the task specification (or, for adapter-absorbed tasks, inherited from the source benchmark). (2)~Per-stage acceptance criteria are emitted in machine-readable form, with tolerance parameters ($\epsilon_{\text{pos}}$, $\epsilon_{\text{ang}}$, $\epsilon_{\text{clear}}$, $\epsilon_{\text{axis}}$) inherited from the skill specifications and instantiated to numerical values appropriate for the bound objects and scene geometry. (3)~Demonstration trajectories are produced through one of two channels: expert teleoperation for high-precision contact-rich tasks, or an embodied planning agent for tasks whose skill compositions can be solved by sampling and verification against the per-stage criteria. In both cases, the resulting trajectories are filtered against the task graph's acceptance criteria, ensuring that only trajectories satisfying every stage constraint enter the released demonstration set.

An instruction-generation step runs alongside task generation: the natural-language template associated with the skill sequence is filled with the bound entities to produce the instruction shown to the policy (e.g., ``Grasp the left of the box and move to right'' from a \textsc{Grasp Part} $\to$ \textsc{Move To} subgraph bound to a box face and the right direction). Both the instruction and the structured task description (including \texttt{task\_id}, \texttt{slots}, \texttt{tolerances}, \texttt{graph\_seq}, and \texttt{acceptance}, as illustrated in the right panel of Fig.~\ref{fig4}) are recorded in the released task specification.

\textbf{Implications for cross-benchmark comparison.}
The adapter and task-generation stages together reposition external benchmarks from competing evaluation regimes into compatible task instances within a shared formalism. A policy evaluated on RoboTwin and on LIBERO through their respective native protocols cannot be directly compared, because the two suites neither define success identically nor decompose competency into the same axes. The same policy evaluated on adapter-translated versions of those tasks is assessed under one acceptance contract and one diagnostic decomposition, making cross-benchmark comparison meaningful at the level of competency rather than at the level of conventions. This is the operational form of MetaFine's claim that benchmark fragmentation is not an inherent property of the field but a consequence of evaluation infrastructure choices.

\section{Hybrid Real--Sim Evaluation Protocol}
\label{app:hybrid-eval}

This section details the hybrid real--sim evaluation protocol introduced in Section~\ref{methodology}. We first formalize the statistical problem and the prediction-powered inference (PPI) estimator (Section~\ref{app:ppi-formalism}); then describe how MetaFine instantiates the real2sim function via 3D Gaussian Splatting (Section~\ref{app:ppi-real2sim}); specify the end-to-end evaluation pipeline (Section~\ref{app:ppi-pipeline}); and finally report an empirical validation that quantifies the gain in estimator precision (Section~\ref{app:ppi-implementation}).

\subsection{Problem Setup and Statistical Formalism}
\label{app:ppi-formalism}

Let $\mathcal{D}_{\text{eval}}$ denote a distribution over physical manipulation environments $\mathcal{X}$ on which we wish to assess a policy $\pi$. We seek to estimate the mean policy performance
\begin{equation}
    A^{*} \;=\; \mathbb{E}_{X \sim \mathcal{D}_{\text{eval}}}\!\left[Y(X)\right],
\end{equation}
where $Y(X) \in [0,1]$ is the real-world evaluation outcome of $\pi$ on environment $X$ under a bounded metric $M$. Direct estimation from real-world rollouts alone is statistically unreliable: hardware experimentation affords only small sample sizes and the resulting empirical mean exhibits high variance.

\textbf{Real2sim function.}
We define a real2sim function $g: \mathcal{X} \rightarrow \mathcal{X}_{\text{sim}}$ that maps a real environment $X$ to its simulation counterpart $\tilde{X} = g(X)$. Simulation evaluations are then given by $f(\tilde{X}) = M_{\text{sim}}(\tilde{X}, \pi)$, where $f$ serves as a scalable, low-variance predictor of the real outcome.

\textbf{Two-stage PPI estimator.}
Following the prediction-powered inference framework~\cite{ppi} as adapted to robot policy evaluation, we sample $X_1, \ldots, X_{n+N} \overset{\text{iid}}{\sim} \mathcal{D}_{\text{eval}}$ and select $n$ of them uniformly at random for paired real-world evaluation, leaving the remaining $N$ for simulation-only evaluation. The paired set yields $\mathcal{D}_{\text{paired}} = \{(Y_i, f(\tilde{X}_i))\}_{i=1}^{n}$, and the unpaired simulations yield $\mathcal{D}_{\text{sim}} = \{f(\tilde{X}_i)\}_{i=n+1}^{n+N}$. The two-stage PPI estimator for $A^{*}$ is
\begin{equation}
    \label{eq:ppi}
    \hat{A}_{\text{PPI}}
    \;=\;
    \underbrace{\frac{1}{n}\sum_{i=1}^{n}\bigl(Y_i - f(\tilde{X}_i)\bigr)}_{\text{rectifier}}
    \;+\;
    \underbrace{\frac{1}{N}\sum_{i=n+1}^{n+N} f(\tilde{X}_i)}_{\text{simulation evaluations}}.
\end{equation}
The rectifier corrects for any systematic bias between simulation and real outcomes, while the simulation term provides a low-variance estimate from the large unlabeled set. Crucially, $\hat{A}_{\text{PPI}}$ is unbiased regardless of simulator fidelity: when $f$ closely tracks $Y$, the rectifier vanishes and the estimator inherits the variance reduction of $N \gg n$; when $f$ is biased, the rectifier absorbs the discrepancy.

\textbf{Finite-sample confidence intervals.}
For a significance level $\alpha$, a finite-sample valid confidence interval on $A^{*}$ is obtained by separately computing intervals on the rectifier (at level $\delta \approx 0.9\alpha$) and on the simulation mean (at level $\alpha - \delta$) via the Waudby-Smith--Ramdas (WSR) algorithm, then taking their Minkowski sum. The resulting interval is valid for any finite $n$ and $N$ without distributional assumptions beyond boundedness, and its width contracts as $N$ grows, up to a floor determined by the rectifier variance~\cite{ppi}.

\subsection{Real2Sim Registration via 3D Gaussian Splatting}
\label{app:ppi-real2sim}

\begin{figure}[h]
    \centering
    \includegraphics[width=0.95\textwidth]{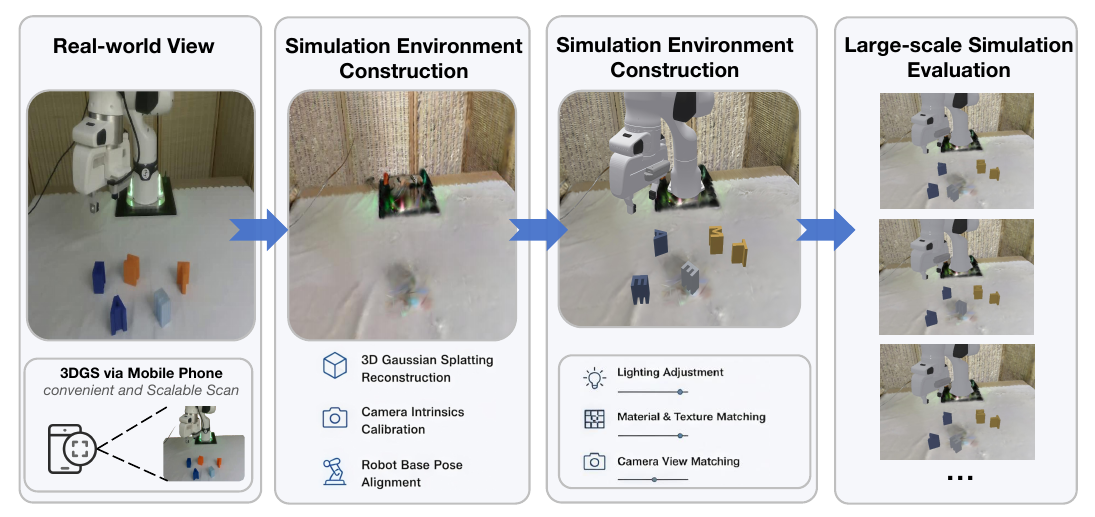}
    \caption{Real2sim registration pipeline. Researchers scan their physical workspace with off-the-shelf 3D Gaussian Splatting tools, export the reconstruction as a PLY file, and remove regions unrelated to the manipulation workspace. MetaFine then aligns the cleaned reconstruction into its simulation backbone with calibrated camera intrinsics and robot base pose. The resulting digital twin supports the platform's standard evaluation interface and the hybrid PPI protocol.}
    \label{fig:real2sim-pipeline}
\end{figure}

In MetaFine, the real2sim function $g$ is instantiated through 3D Gaussian Splatting (3DGS)-based scene reconstruction. Researchers scan their physical workspace using off-the-shelf tools (e.g., Scaniverse) and export the result as a Gaussian Splatting or mesh-compatible PLY file. The PLY is post-processed to remove regions unrelated to the manipulation workspace (background, occluding bodies, scan artifacts) before MetaFine automatically registers the cleaned reconstruction into its simulation backbone, aligning camera intrinsics, robot base pose, and task-relevant object placements. Once registered, the digital twin supports the same evaluation interface as MetaFine's native scenes, enabling prediction-powered inference (Section~\ref{app:ppi-formalism}) to operate without further user-side adjustment. The end-to-end pipeline is illustrated in Fig.~\ref{fig:real2sim-pipeline}.

\subsection{Evaluation Pipeline}
\label{app:ppi-pipeline}

The hybrid protocol unfolds in five stages.

\textbf{(1) Distribution definition.} The platform fixes a public evaluation distribution $\mathcal{D}_{\text{eval}}$ over object placements, initial poses, and other task-relevant factors for each task in MetaFine's compositional task graph. The distribution is specified as a bounded region in configuration space.

\textbf{(2) Configuration sampling.} The platform pre-samples $n + N$ configurations $\{X_1, \ldots, X_{n+N}\}$ iid from $\mathcal{D}_{\text{eval}}$ and randomly designates $n$ of them as the paired set, which is mandatory for any valid evaluation submission. When the protocol is run multiple times for variance characterization, independent paired sets are produced by sampling additional batches of $n$ points uniformly within the same configuration region; each sampled point then initializes the object position for the corresponding paired trial.

\textbf{(3) Real2sim registration.} The user scans their physical workspace via 3D Gaussian Splatting (Section~\ref{app:ppi-real2sim}) and uploads the reconstruction; MetaFine automatically aligns the scene to its simulation infrastructure.

\textbf{(4) Paired execution.} The user runs all $n + N$ configurations in simulation and the prescribed $n$ configurations on real hardware, yielding $\{Y_i\}_{i=1}^{n}$ and $\{f(\tilde{X}_i)\}_{i=1}^{n+N}$.

\textbf{(5) Calibrated reporting.} The platform computes $\hat{A}_{\text{PPI}}$ together with its WSR confidence interval (Eq.~\ref{eq:ppi}) and reports them as the official policy performance for the task.

Because $\mathcal{D}_{\text{eval}}$ and the paired-set sampling procedure are platform-prescribed rather than user-selected, evaluations conducted at different laboratories assess the same policy under the same statistical contract and become directly comparable.

\subsection{Empirical Validation}
\label{app:ppi-implementation}

We instantiate the protocol on a representative fine-grained task, \emph{Grasp the character E}, where the policy must grasp a custom-designed character-shaped target whose geometry imposes part-level contact constraints. The character models are open-sourced as part of the MetaFine release. The real-world setup uses a Franka Panda arm with a wrist-mounted RealSense D435; the simulated counterpart is reconstructed via 3D Gaussian Splatting through Scaniverse and registered into MetaFine's simulation backbone. We evaluate two policies under the protocol: $\pi_0$ and $\pi_{0.5}$.

For each policy, we collect $N = 1000$ unpaired simulation rollouts and a large-sample real-world reference of 300 hardware rollouts to serve as ground truth. To characterize estimator variance under realistic hardware budgets, we further collect three independent paired sets of $n = 20$ rollouts each (60 paired real--sim rollouts per policy in total). The three paired sets are produced by sampling three independent batches of object placements within the configuration region of $\mathcal{D}_{\text{eval}}$, each batch then initializing the corresponding paired trials on both hardware and simulation.

\textbf{Per-paired-set estimates.}
Table~\ref{tab:ppi_results} reports, for each of the three paired sets, the hardware-only estimate (using only the 20 real rollouts on the paired configurations) and the PPI estimate that combines the paired real--sim outcomes with the unpaired simulation mean.

\begin{table}[ht]
\centering
\caption{PPI-calibrated estimates across three independent paired sets ($n=20$ each, $N=1000$ unpaired simulation rollouts per policy). ``Hardware-only'' is the per-set real-only estimate. ``PPI estimate'' combines the paired real--sim outcomes with the unpaired simulation mean via Eq.~\ref{eq:ppi}. The gold-standard real-world reference $A$ is computed from 300 independent real rollouts.}
\label{tab:ppi_results}
\vspace{0.4em}
\small
\setlength{\tabcolsep}{6pt}
\begin{tabular}{ll cc}
\toprule
\textbf{Policy} & \textbf{Paired set} & \textbf{Hardware-only} & \textbf{PPI estimate} \\
\midrule
\multirow{4}{*}{$\pi_{0.5}$}
 & $B_1$ & 75\% & 70\% \\
 & $B_2$ & 55\% & 66\% \\
 & $B_3$ & 75\% & 71\% \\
 & \emph{Reference $A$} & \multicolumn{2}{c}{\emph{65.0\% (195/300)}} \\
\midrule
\multirow{4}{*}{$\pi_0$}
 & $B_1$ & 65\% & 57\% \\
 & $B_2$ & 65\% & 57\% \\
 & $B_3$ & 55\% & 62\% \\
 & \emph{Reference $A$} & \multicolumn{2}{c}{\emph{60.3\% (181/300)}} \\
\bottomrule
\end{tabular}
\end{table}

\textbf{Variance reduction across paired sets.}
The dispersion and accuracy of the two estimators across the three paired sets expose the central benefit of the PPI calibration. Table~\ref{tab:ppi_aggregate} summarizes their behavior relative to the gold-standard reference.

\begin{table}[ht]
\centering
\caption{Aggregate behavior of the hardware-only and PPI-calibrated estimators across the three paired sets. Dispersion is the sample standard deviation of the three per-set estimates; ``Max $|\cdot - A|$'' is the largest absolute deviation from the 300-rollout reference among the three sets.}
\label{tab:ppi_aggregate}
\vspace{0.4em}
\small
\setlength{\tabcolsep}{6pt}
\begin{tabular}{l ccc ccc}
\toprule
& \multicolumn{3}{c}{\textbf{Hardware-only}} & \multicolumn{3}{c}{\textbf{PPI-calibrated}} \\
\cmidrule(lr){2-4} \cmidrule(lr){5-7}
\textbf{Policy} & Mean & Dispersion & Max $|\cdot - A|$ & Mean & Dispersion & Max $|\cdot - A|$ \\
\midrule
$\pi_{0.5}$ & 68.3\% & 11.5\% & 10.0 pp & 69.0\% & \textbf{2.6\%} & \textbf{6.0 pp} \\
$\pi_0$     & 61.7\% &  5.8\% &  5.3 pp & 58.7\% & \textbf{2.9\%} & \textbf{3.3 pp} \\
\bottomrule
\end{tabular}
\end{table}

For both policies, PPI calibration substantially reduces cross-set dispersion ($\pi_{0.5}$: $11.5\% \to 2.6\%$; $\pi_0$: $5.8\% \to 2.9\%$) and improves worst-case deviation from the 300-rollout reference ($\pi_{0.5}$: $10.0 \to 6.0$ pp; $\pi_0$: $5.3 \to 3.3$ pp). The hardware-only estimator at $n=20$ remains highly sensitive to the particular configurations sampled---for $\pi_{0.5}$, the three paired sets yield 55\%, 75\%, and 75\%, a 20-point spread around the true 65.0\% reference. PPI contracts these estimates to a much narrower 66\%--71\% range. This variance reduction appears in both regimes: when the rectifier is small, the unpaired simulation mean dominates the estimate; when the rectifier is larger, the paired correction term still contracts cross-set variability relative to hardware-only evaluation.

\textbf{Connection to the main-text comparison.}
Fig.~\ref{fig3} of the main text visualizes this contrast at the level of individual paired-set estimates: the red markers correspond to the three hardware-only estimates of size $n=20$, the blue markers to the corresponding PPI-calibrated estimates, and the horizontal dashed line to the 300-rollout reference $A$. The figure makes visible what the tables quantify: PPI calibration pulls the three estimates substantially closer to $A$, removing the wide swings of the hardware-only estimator at small $n$.

\textbf{Practical implications.}
Two observations carry beyond this case study. First, the calibration provides a meaningful reduction in estimator variance even at the modest hardware budget of $n=20$ paired rollouts, which is realistic for most laboratory settings. Second, the worst-case hardware-only error at $n=20$ (10.0 pp on $\pi_{0.5}$, 5.3 pp on $\pi_0$) is large enough that hardware-only point estimates should be treated with caution in cross-policy comparisons; PPI calibration reduces the worst-case error to 6.0 pp and 3.3 pp, respectively. These properties motivate the hybrid protocol as MetaFine's default for real-world policy comparison.

\section{Multi-Scale Encoder Design}
\label{supp:sec4}

The main text demonstrates that the visual encoder's spatial fidelity establishes an upper bound on downstream fine-grained manipulation precision (Fig.~\ref{fig5}A).
Class activation map analysis reveals that $\pi_{0.5}$'s single-scale SigLIP encoder produces diffuse activation spread broadly across the scene, lacking the spatial concentration needed to resolve the relative pose between a grasped object and a target receptacle.
This section describes the multi-scale cross-attention encoder that replaces $\pi_{0.5}$'s original visual frontend, achieving the results reported in Fig.~\ref{fig5} of the main text (grasp success from 39\% to 67\%; alignment from 0\% to 32\%) without modifying the VLM backbone or action head.

Fine-grained manipulation demands spatial information at multiple scales simultaneously: coarse features for object-level localization and scene understanding, and fine features for part-level contact selection and pose-level alignment.
$\pi_{0.5}$'s original pipeline feeds a single-scale feature representation from SigLIP into the VLM backbone, a design inherited from vision-language models where global semantic understanding suffices.
Under this single-scale regime, the encoder must trade off between field of view and spatial detail within a fixed resolution budget, and the resulting representation lacks the fine-grained spatial fidelity that precision manipulation requires.
Our design goal is to enhance the spatial quality of the visual representation while preserving full compatibility with the pretrained VLM backbone and flow-matching action head, modifying only the perception frontend.

For each of the $M$ input cameras, we generate two views at different spatial scales: the original image and a center crop at 0.5$\times$ the spatial extent, bilinearly resized back to the original resolution.
The center crop effectively doubles the spatial sampling density over the central workspace region where manipulation occurs, surfacing fine-grained geometric details that are underresolved in the original view.
This yields $M \times 2$ images in an interleaved arrangement: $[\text{orig}_{\text{cam}_0}, \text{crop}_{\text{cam}_0}, \text{orig}_{\text{cam}_1}, \text{crop}_{\text{cam}_1}, \ldots]$.

All $M \times 2$ images are encoded by the same pretrained SigLIP encoder followed by the PaliGemma multi-modal projector, producing $N_{\text{patch}} = 256$ tokens of dimension $D = 2048$ per image.
Using a shared encoder rather than separate encoders for each scale ensures that low-level feature extraction is consistent across scales and avoids parameter duplication.
To enable downstream modules to distinguish the spatial scale from which each token originates, we inject a learned scale embedding:
\begin{equation}
    \mathbf{z}_{i}^{(s)} = \mathbf{z}_{i}^{(s)} + \mathbf{e}_s, \quad s \in \{0, 1\},
\end{equation}
where $\mathbf{e}_s \in \mathbb{R}^{D}$ is a learned embedding for scale $s$ (original or crop), broadcast to all patch tokens of that scale.
The resulting tokens from all cameras and scales are concatenated into a single sequence of length $M \times 2 \times N_{\text{patch}}$ (\eg, 1,536 tokens for $M = 3$ cameras).

Directly feeding the concatenated multi-scale tokens into the VLM backbone would substantially increase the sequence length and exceed practical context limits.
We therefore compress the variable-length visual token sequence into a fixed-length representation using a Perceiver resampler.
The resampler maintains a set of 128 learned query vectors $\mathbf{Q} \in \mathbb{R}^{128 \times D}$ that attend to the full concatenated visual token sequence through cross-attention:
\begin{equation}
    \mathbf{Q} \leftarrow \mathbf{Q} + \text{MHA}\!\left(\text{LN}(\mathbf{Q}),\; \text{LN}(\mathbf{Z}_{\text{cat}}),\; \text{LN}(\mathbf{Z}_{\text{cat}})\right),
\end{equation}
followed by a feed-forward network with GELU activation and 4$\times$ expansion:
\begin{equation}
    \mathbf{Q} \leftarrow \mathbf{Q} + \text{FFN}\!\left(\text{LN}(\mathbf{Q})\right).
\end{equation}
This cross-attention and feed-forward block is repeated for 2 layers with 8 attention heads.
The output is a fixed set of 128 compressed visual tokens regardless of the number of input cameras, achieving a compression ratio of approximately 12$\times$ (1,536 $\to$ 128 for $M = 3$).
These compressed tokens replace the original visual prefix in $\pi_{0.5}$'s pipeline: they are concatenated with language tokens and passed to the PaliGemma backbone and flow-matching action head, both of which remain entirely unchanged.

The multi-scale encoder introduces approximately 25M new trainable parameters (Perceiver resampler weights and scale embeddings).
The PaliGemma backbone and action expert weights are initialized from the pretrained $\pi_{0.5}$ checkpoint; only the newly introduced parameters are trained from scratch.
This modular design ensures that the enhancement to spatial perception does not disrupt the pretrained language understanding and action generation capabilities.

Several design choices merit brief justification.
Center cropping, as opposed to random cropping, provides a deterministic and reproducible augmentation that focuses spatial enhancement on the central workspace region where the robot's end-effector and manipulation targets are typically located.
The shared encoder across scales ensures that features from different spatial resolutions are extracted by the same learned filters, with scale embeddings providing the disambiguation signal; this is more parameter-efficient than maintaining separate encoder branches.
Most importantly, restricting modifications to the visual frontend is motivated directly by the empirical finding in the main text that encoder-only adaptation suffices to recover manipulation precision (Fig.~\ref{fig5}B): the VLM backbone and action head remain functionally competent once the spatial representation is corrected, and retraining them risks disrupting capabilities that are not bottlenecked by perception.

\section{Action Head Trajectory Analysis}
\label{supp:sec5}

This section provides a detailed analysis of how different action generation paradigms shape fine-grained execution behavior. The main text presents the key findings; here we describe the analysis methodology and expand on the behavioral patterns observed in the trajectory data.

\subsection{Analysis Methodology}

Three complementary lenses are applied to the raw action sequences recorded during evaluation rollouts.

\textbf{End-effector trajectory visualization.}
For each rollout, the 3D position of the end-effector is recorded at every time step and rendered in the task workspace. Color-coding along the trajectory encodes temporal progression, enabling visual inspection of whether successive steps exhibit directional consistency (convergence toward the target) or directional incoherence (spatial drift). Representative visualizations are shown in Fig.~\ref{fig3}A of the main text.

\textbf{Action magnitude.}
The $\ell_2$ norm $\|a_t\|$ of the action vector is computed at each time step and its evolution plotted over the course of a rollout. Under normal execution, action magnitudes exhibit task-dependent temporal structure: typically large during the approach phase and decreasing as the end-effector nears the target. A sudden and sustained collapse in action magnitude to near-zero values, without task completion, indicates that the policy has ceased generating meaningful motor commands---a phenomenon we term \emph{action-space collapse}.

\textbf{Directional consistency.}
To quantify the coherence of successive actions, the cosine similarity between consecutive action vectors is computed:
\begin{equation}
    \text{cos\_sim}(a_t, a_{t+1}) = \frac{a_t \cdot a_{t+1}}{\|a_t\| \, \|a_{t+1}\|}.
\end{equation}
Values near 1.0 indicate that successive steps move in a consistent direction; values near 0 or negative indicate directional incoherence. Averaging this measure over the trajectory yields a scalar summary of directional consistency. Deterministic regression tends to produce high average directional consistency even under perceptual uncertainty, because the same noisy observation deterministically maps to the same action. Stochastic flow matching, by contrast, can produce low directional consistency when conditioned on spatially ambiguous observations, since independent samples at successive steps may point in different directions.

\subsection{Convergence versus Drift on Peg-in-Hole}

The peg-in-hole task provides a revealing testbed for comparing action generation paradigms, as it requires sustained multi-step precision under imperfect perception. Both OpenVLA-OFT (deterministic regression) and $\pi_{0.5}$ (flow matching) receive imprecise spatial input from their respective encoders, yet their behavioral responses to this shared perceptual limitation diverge sharply.

OpenVLA-OFT's trajectory (Fig.~\ref{fig3}A) exhibits a characteristic convergence pattern. The end-effector approaches the target region through a series of steps that maintain directional consistency: each step moves in approximately the same direction as the previous one, and the cumulative effect is steady progress toward the target. This follows directly from the deterministic mapping $a_t = f(\hat{o}_t)$: because the encoder's systematic bias $b$ shifts the perceived target position in a consistent direction, the policy produces actions consistently biased in the same way. Over multiple steps, this correlated bias accumulates into coherent motion rather than canceling out, enabling the end-effector to reach the vicinity of the target despite never receiving a perfectly accurate observation. As formalized in Eq.~1 of the main text, the expected positional error after $T$ steps grows linearly with the bias magnitude $\|b\|$ but does not include a variance term from the action generation process itself. The policy's limitation becomes apparent at the alignment and insertion stages, where it converges to a narrow behavioral pattern and repeatedly retries the same approach trajectory without exploring alternative adjustment strategies.

$\pi_{0.5}$'s trajectory (Fig.~\ref{fig3}A) shows a qualitatively different pattern. Successive steps lack directional consistency: the end-effector changes direction frequently and, over multiple steps, fails to maintain sustained progress toward the target. This arises because flow matching draws a fresh stochastic sample $a_t \sim p_\theta(\cdot|\hat{o}_t)$ at each step. When the conditioning observation $\hat{o}_t$ is spatially ambiguous---as occurs when the encoder provides diffuse rather than concentrated activation over the task-relevant region---the learned conditional distribution is broad, and successive samples may point in inconsistent directions. As captured by the $T\sigma_\xi^2$ term in Eq.~2 of the main text, this per-step sampling variance accumulates linearly with the number of steps, producing the spatial drift visible in the trajectory data. Under precise perception ($\|b\| \approx 0$), this term would be negligible and flow matching's distributional expressiveness would be advantageous for expressing diverse corrective strategies. Under the imprecise perception characteristic of current fine-grained manipulation, however, the accumulated sampling variance dominates, and the end-effector fails to converge.

\subsection{Behavioral Arrest versus Sub-Task Transition in Long-Horizon Sorting}

The compound sorting task reveals a second dimension along which deterministic and stochastic action generation differ: the capacity to transition between sub-tasks in long-horizon manipulation.

After successfully completing the first sub-task (grasping and placing the specified cube), OpenVLA-OFT's action magnitudes drop to near-zero and remain there for the rest of the rollout (Fig.~\ref{fig3}A). The policy does not attempt to seek, approach, or grasp the remaining cubes. This behavioral arrest is consistent across all three color variants (Table~\ref{tab:stagewise_sort}): stages 3--4 are uniformly 0\% for OpenVLA-OFT. The pattern indicates that the deterministic mapping from observation to action has converged to a fixed point in action space once the first sub-task is completed, and the policy lacks the mechanism to break out of this attractor and initiate a qualitatively different behavioral mode for the second sub-instruction.

$\pi_{0.5}$, by contrast, continues to produce non-trivial actions after completing the first sub-task (Fig.~\ref{fig3}A). Inspection of successful trials at stages 3--4 confirms that these actions are purposeful: the end-effector actively seeks and approaches the remaining cubes rather than persisting with the completed sub-task. The stochasticity of flow matching appears to facilitate this transition by preventing the policy from collapsing to a single behavioral mode. Even when the observation provides weak guidance toward the second sub-task, the sampling process can explore actions outside the attractor basin of the completed first sub-task, enabling the policy to discover and execute the transition. This advantage manifests consistently across all three color variants, with stages 3--4 success rates ranging from 11--28\% (Table~\ref{tab:stagewise_sort}).

\subsection{Dissociation between Trajectory Quality and Task Success}

MetaFine’s joint evaluation of trajectory smoothness and task success reveals cases in which the two metrics diverge, exposing failure modes that neither metric alone can fully capture.

On Rotate Along, $\pi_{0.5}$ achieves a stability score of 0.90 with only 10\% success rate (Table~\ref{tab:nominal_sr}). The vast majority of its rollouts produce smooth, controlled rotational motions that nonetheless fail to satisfy the directional constraint---the rotation is fluid but along the wrong axis or in the wrong direction. Binary success rate would report this as simple failure, while trajectory smoothness alone would suggest competent execution. Only the combination of both metrics, as provided by MetaFine's behavioral evaluation, reveals the true failure mode: the policy has learned smooth rotational primitives but has not grounded the directional specification in the instruction.

On the peg-in-hole task, OpenVLA-OFT achieves 47\% grasp success and 19\% alignment but only 3\% insertion (Table~\ref{tab:stagewise_peg}). Its trajectory stability score of 0.71---the lowest among all successful rollouts---reflects the repeated retry behavior at the insertion stage: the policy converges to the vicinity of the hole and then oscillates as it attempts the same insertion trajectory multiple times without adaptation. The low stability does not indicate loss of control but rather the absence of corrective diversity: the deterministic action head cannot express the range of fine adjustments that the tight insertion constraint demands, resulting in repetitive, ultimately unsuccessful approach attempts.

Taken together, these observations point to a structural trade-off in fine-grained manipulation under imperfect perception. Deterministic regression provides noise-filtering consistency that enables stable approach trajectories but lacks the expressiveness needed for fine corrections and sub-task transitions. Flow matching provides the distributional capacity for diverse corrective strategies and behavioral flexibility but is vulnerable to directional drift when conditioned on uncertain perceptual input. These complementary failure modes locate a concrete design target: action generation architectures for fine-grained manipulation should modulate the balance between stability and expressiveness as task demands shift across manipulation stages, rather than commit to a single paradigm across the full execution horizon.

\section{Related Works}
\label{related_work}
\subsection{Fine-Grained Manipulation}
Robotic manipulation has progressed from object-level pick-and-place toward tasks that impose constraints at finer granularities. Part-level grasping methods leverage affordance reasoning and language grounding to generate grasp poses on specific object parts~\cite{song2023langpartgpd,huang2024copa,tian2024cgdf}, while dexterous manipulation and precision assembly address the execution side, requiring contact-rich control and compliance with tight geometric tolerances~\cite{chi2023diffusionpolicy,ze2024dp3}. Part-level 3D understanding has also advanced: PartNet~\cite{mo2019partnet} established large-scale hierarchical part annotations, and PartInstruct~\cite{geng2025partinstruct} recently introduced part-level instruction following as a benchmark category. However, these efforts each address isolated aspects of fine-grained manipulation, whether part-level grasp selection, constrained motion, or attribute-aware instruction following, and universally rely on binary success rate for evaluation. No existing framework formalizes what fine-grained manipulation demands from an agent across these aspects simultaneously, nor provides evaluation that can diagnose whether failure originates from misunderstood semantics, imprecise perception, or unstable execution. MetaFine addresses this gap with a compositional skill vocabulary that captures the full scope of fine-grained constraints and a three-dimensional protocol that decomposes competency along understanding, perception, and controlled behavior.

\subsection{Visual Motor Policy}

Visual motor policies map raw observations directly to robot actions, and their architectures have diversified rapidly along several axes. Action generation has evolved from deterministic regression~\cite{zhao2023act} through diffusion-based generation~\cite{chi2023diffusionpolicy,ze2024dp3} to flow-matching formulations~\cite{black2024pi0}, each offering different tradeoffs between expressiveness and inference stability. Visual representations span 2D image encoders~\cite{radford2021clip,oquab2024dinov2}, 3D point-cloud pipelines~\cite{ze2024dp3}, and dual-encoder fusions that combine self-supervised geometric features with language-grounded semantics~\cite{kim2024openvla,kim2025openVLAoft}. Language conditioning has progressed from absent or coarse task-level selection to vision-language-action (VLA) models that jointly process language instructions and visual observations within a unified transformer backbone, with recent systems such as RT-2~\cite{brohan2023rt2}, OpenVLA~\cite{kim2024openvla}, $\pi_0$~\cite{black2024pi0}, and $\pi_{0.5}$~\cite{black2025pi05} demonstrating increasingly general manipulation capabilities. These architectural choices, including encoder design, action head formulation, and the depth of language--action coupling, each influence distinct aspects of manipulation competency. Yet existing evaluation protocols assess all policies through the same binary success metric, providing no means to determine whether an architectural improvement benefits semantic grounding, spatial precision, or execution stability. MetaFine's three-dimensional protocol enables precisely this diagnosis, linking specific design choices to their effects on understanding, perception, and controlled behavior.

\subsection{Manipulation Benchmarks and Evaluation}
Manipulation benchmarks have scaled from single-task grasping evaluations to multi-task suites that cover diverse skills and objects. RLBench~\cite{james2020rlbench} and ManiSkill~\cite{mu2023maniskill2} provide large-scale simulated task collections with standardized interfaces, CALVIN~\cite{calvin} targets long-horizon language-conditioned manipulation, LIBERO~\cite{libero} benchmarks lifelong learning across task distributions, and RoboTwin~\cite{robotwin} introduces dual-arm scenarios with generative digital twins. Despite their growing breadth, these benchmarks share two structural limitations. First, they universally rely on binary task success as the primary evaluation metric, collapsing all sources of failure into a single scalar that cannot distinguish perceptual degradation from semantic misunderstanding or execution instability. Some recent efforts introduce auxiliary metrics such as subtask progress or robustness tests under domain shift~\cite{xie2024decomposing}, but these remain benchmark-specific additions rather than components of a unified evaluation framework. Second, each benchmark defines its own task specifications, success criteria, and environmental conditions, making cross-benchmark comparison unreliable: a policy that excels on one suite and struggles on another cannot be diagnosed without understanding which competency dimension accounts for the discrepancy. MetaFine addresses both limitations simultaneously. Its three-dimensional protocol replaces binary evaluation with structured competency diagnosis, and its benchmark absorption mechanism re-expresses tasks from heterogeneous benchmarks within a shared task-graph formalism, enabling fair comparison under a unified fine-grained evaluation standard.

\subsection{Real-to-Sim Evaluation}
\label{sec:related_real2sim}

A complementary line of work seeks to narrow the real-world evaluation gap by constructing simulators that closely mirror physical setups. SIMPLER~\cite{li2025simpler} showed that simulated success rates can correlate strongly with real-world performance once control and visual discrepancies are reduced, while RialTo~\cite{torne2024rialto} extends this idea into the learning loop through on-the-fly digital twins. The maturation of 3D Gaussian Splatting~\cite{kerbl20233dgs} has further accelerated photorealistic real2sim pipelines for both rigid~\cite{li2024robogsim,han2025re3sim} and deformable~\cite{zhang2025real2sim_softbody} manipulation, and recent benchmarks~\cite{sedlacek2025realm,nasiriany2024robocasa} increasingly adopt real2sim fidelity as a design principle. A shared premise across these efforts is that simulation can \emph{replace} real-world rollouts, with credibility depending on how faithfully the digital twin captures reality, an assumption that is difficult to verify and may silently fail as the sim-real gap widens. MetaFine adopts a different stance: rather than substituting for reality, simulation serves as a variance-reduction mechanism within a statistically valid estimator. By combining many simulated rollouts with a small paired set of real rollouts through prediction-powered inference~\cite{ppi}, MetaFine produces a calibrated estimate of true real-world performance whose uncertainty decreases as simulation scales, even when the twin is imperfect. Sim-real fidelity therefore shifts from a prerequisite for trustworthy evaluation to a factor governing estimator efficiency, recasting real2sim not as a substitution problem but as one of statistical efficiency, in line with MetaFine's broader view of evaluation-as-diagnosis rather than evaluation-as-ranking.

\end{appendices}
\end{document}